\documentclass[journal]{IEEEtran}

\usepackage[]{todonotes}
\usepackage{graphicx}
\graphicspath{{../graphics/}{./figure/}}
\usepackage{xcolor} % colored text
\usepackage{booktabs} % for fancy table
\usepackage{multirow} % similar as multicolumn for rows
\usepackage{tabularx}
\newcolumntype{Y}{>{\centering\arraybackslash}X}
\usepackage{soul} % highlighting text
\usepackage{amsmath,mathtools,amsfonts} % for \prescript
\usepackage[most]{tcolorbox}
\usepackage{color, colortbl}
\usepackage[hidelinks]{hyperref}
\usepackage{BOONDOX-cal} % for the screw notations
\usepackage[caption=false,font=footnotesize]{subfig}
\usepackage{cite}

\newcommand{\norm}[1]{\left\lVert#1\right\rVert}
\newcommand{\refframe}[1]{\{#1\}}
\newcommand{\T}[2]
{\prescript{#1}{#2}{\boldsymbol{T}}}
\newcommand{\R}[2]
{\prescript{#1}{#2}{\boldsymbol{R}}}
\newcommand{\pvector}[2]
{\prescript{#2}{#1}{\boldsymbol{p}}}
\newcommand{\velvector}[4]
{\vphantom{v}_{#3}^{#4}\boldsymbol{v}_{#2}^{#1}\vphantom{v}}

\newcommand{\omegavector}[3]
{\vphantom{\omega}_{#3}^{#1}\boldsymbol{\omega}_{#2}\vphantom{\omega}}

\newcommand{\Tdot}[2]
{\prescript{#1}{#2}{\vphantom{T}}\dot{\boldsymbol{T}}}
\newcommand{\zerom}[1]{\boldsymbol{\mathit{0}}_{\hspace*{-0.0ex}#1}}

\newcommand{\vect}[1]{\boldsymbol{#1}} % vector in reference frame
\newcommand{\vectmf}[1]{\boldsymbol{\widetilde{#1}}} % vector in moving frame
\newcommand{\mat}[1]{\boldsymbol{#1}} % matrix in reference frame
\newcommand{\matmf}[1]{\boldsymbol{\widetilde{#1}}} % matrix in moving frame
\newcommand{\screw}[1]{\boldsymbol{\mathcal{#1}}} % screw in reference frame
\newcommand{\screwmf}[1]{\boldsymbol{\widetilde{\mathcal{#1}}}} % screw in moving frame

\newcommand\scalemath[2]{\scalebox{#1}{\mbox{\ensuremath{\displaystyle #2}}}}

\begin{document}

\title{Invariant Descriptors of Motion and Force Trajectories for Interpreting Object Manipulation Tasks in Contact}

\author{Maxim~Vochten${}^{1,3,4}$, Ali~Mousavi~Mohammadi${}^{1,3}$, Arno~Verduyn${}^{1,3}$, Tinne~De~Laet${}^2$, Erwin~Aertbeli\"en${}^1$, and Joris~De~Schutter${}^1$% <-this % stops a space
	\thanks{This result is part of a project that has received funding from the European Research Council (ERC) under the European Union's Horizon 2020 research and innovation programme (Grant agreement No. 788298).}%
	\thanks{${}^{1}$ Department of Mechanical Engineering and Flanders Make at KU Leuven, 3001 Leuven, Belgium.}%
	\thanks{${}^{2}$ Faculty of Engineering Science, KU Leuven, 3001 Leuven, Belgium.}%
	\thanks{${}^{3}$ These authors contributed equally to this work.}%
	\thanks{${}^{4}$ Corresponding author (maxim.vochten@kuleuven.be).}%
}%

\maketitle

\begin{abstract}
	Invariant descriptors of point and rigid-body motion trajectories have been proposed in the past as representative task models for motion recognition and generalization. Currently, no invariant descriptor exists for representing force trajectories, which appear in contact tasks. This paper introduces invariant descriptors for force trajectories by exploiting the duality between motion and force. Two types of invariant descriptors are presented depending on whether the trajectories consist of screw or vector coordinates. Methods and software are provided for robustly calculating the invariant descriptors from noisy measurements using optimal control. Using experimental human demonstrations of 3D contour following and peg-on-hole alignment tasks, invariant descriptors are shown to result in task representations that do not depend on the calibration of reference frames or sensor locations. The tuning process for the optimal control problems is shown to be fast and intuitive. Similar to motions in free space, the proposed invariant descriptors for motion and force trajectories may prove useful for the recognition and generalization of constrained motions, such as during object manipulation in contact.
\end{abstract}


\section{Introduction}

\IEEEPARstart{I}{n} \emph{human-robot interaction} (HRI) there is a need for generalizable task models that allow robots to reactively adapt the robot's execution and to robustly recognize human actions in continuously changing environments. For execution purposes, these models should be easily adaptable to deal with obstacles, different starting/ending/via points, robot platform constraints, and human-robot interaction. For recognition purposes, these models should be sufficiently general to recognize the human's action at different locations in space, from different viewpoints, and with different execution styles (e.g. varying scale and motion profile). The focus of this paper is on extracting task representations that involve the manipulation of objects in contact, where not only the relative motion between objects but also interaction forces are relevant as input for the task modeling process. Typical applications include surface following tasks and assembly tasks.

Task models in HRI are commonly learned from one or multiple human demonstrations of the task using \emph{Learning by Demonstration} \cite{billard2016learning}. The challenge is to achieve a representative task model without the need for a high number of demonstrations.

One approach to achieve representative task models from few demonstrations is to exploit \text{invariance} in the task. Here, \textit{invariance} refers to properties of the task that remain unchanged under certain transformations such as changes in reference frame. Invariance has proven useful in various fields of robotics. Typically, the objectives were: (1) to ensure that exactly the same task execution or behavior of a robot system was obtained within its workspace, and (2) to ensure that the quantitative characterization of such robot action using a metric resulted in exactly the same number(s) regardless of where and how the robot action was recorded or measured. Below, some  examples of the importance and use of invariant solutions in different robotics contexts are briefly reported.

For the \textit{control} of free-space robot motion such as mobile robots and UAVs, invariance was achieved by representing the trajectory tracking error  in such a way that the same closed-loop behavior for trajectory tracking and disturbance rejection was obtained regardless of the direction in which the robot was moving \cite{5499132}. For constrained motion, i.e., motion in contact, which is the subject of this paper, the risk of obtaining non-invariant control solutions is more pronounced since both motion (translation and rotation) and interaction wrenches (i.e. forces and moments) have to be controlled simultaneously. For example, as pointed out in \cite{duffy1990fallacy}, it does not make sense to define \textit{orthogonality} between two twists (i.e. translational and rotational velocities), or between two contact wrenches, because of (1) dimensional inconsistency, (2) dependence on the choice of units, and (3) dependence on the choice of the origin of the coordinates. In contrast, \emph{reciprocity} between a twist and a wrench is an invariant concept. In \cite{lipkin1985,lipkinduffy1988} a firm geometric foundation for kinestatics and invariant hybrid force/position control was laid based on screw theory. Taking this into consideration, invariant solutions were proposed, for example to obtain dynamic models of redundant and constrained robot manipulators \cite{846414}, or in applications of hybrid force/position control \cite{DeSchutter1997}.

In \textit{estimation}, invariance refers to obtaining the same values for the estimated quantities regardless of the reference frame in which the estimator's states are represented. For example, the Invariant Extended Kalman Filter \cite{barrau2016invariant} formulated the state correction term in an invariant way using Lie group theory and showed a better convergence of the estimation compared to a standard Extended Kalman filter with a linear correction term. The Invariant Extended Kalman Filter was applied, for instance, to attitude estimation for UAVs \cite{4434662}.

In \textit{motion generation}, invariance refers to obtaining exactly the same physical desired motion trajectory regardless of the reference frame in which the trajectory coordinates are expressed. Human reaching motions, represented as point trajectories, were found to correspond well to a minimum-jerk profile, an invariant property that was shown to be equivalent to having a constant equi-affine curvature \cite{Bennequin2009}. Based on this property, new trajectories can be generated or adapted from reference trajectories  in an invariant way \cite{meirovitch2016geometrical}. This and other invariant properties such as shape preservation were considered for the deformation of trajectories using global transformations \cite{pham2015}. For rigid-body motion, approaches were proposed where the trajectories are either generated in a left-invariant way, meaning independent of the world reference frame, or in a right-invariant way, meaning independent of the body reference frame \cite{Belta2002}. This was also taken into account for the adaptation of reference or demonstrated rigid-body trajectories \cite{Laha2022,vochten2019generalizing}.

Invariance in \textit{action recognition} refers to recognizing actions regardless of where and how they are recorded. This can be done by comparing the recorded action instance and its model using invariant distance metrics. For example, local invariant signatures were proposed for describing and recognizing object outlines \cite{Calabi1998} and trajectories \cite{wu2008signature}. These invariant representations were shown to be invariant to occlusion, rotation, translation, and scaling.

These previous works showed that invariant representations offer many advantages that are relevant for different robotics applications. In the remainder of the paper we focus on invariant descriptors for trajectories. These descriptors can be used to build representative invariant models of a demonstrated task in a human-robot interaction context with fewer demonstrations compared to conventional methods using coordinate-based descriptors. In addition, thanks to their invariant properties, these models enable robust action recognition in different recording contexts and exhibit good generalization properties in motion generation.

\subsection{Invariant trajectory descriptors for motion}

Invariant trajectory descriptors are features of the trajectory that remain unchanged under certain transformations of the original trajectory coordinates, such as changes in reference frame, scale, and time profile. Invariant descriptors can be categorized in many types \cite{Weiss1993} depending on the features that are extracted from the curve. Many have been proposed, both for point trajectories \cite{Calabi1998,piao2006,WuLi2009,shao2015integral,rao2002view} and for rigid-body trajectories \cite{guo2018dsrf,guo2017rrv,DeSchutter2010,lee2017bidirectional,vochten2015}.

This paper focuses on a particular type of invariant descriptor, referred to as \textit{differential invariants} \cite{Calabi1998,Weiss1993}. These invariants correspond to differential-geometric properties of the trajectory and provide a local invariant representation that is robust to occlusion and segmentation errors. Furthermore, they allow to reconstruct and generate new trajectories from the invariant descriptor by integrating corresponding differential equations.
The definition of such local differential-geometric invariants can be seen as an application of \textit{Cartan's method of moving frames} \cite{guggenheimer1977differential}, which consists of defining a \textit{natural moving frame} in the considered manifold, after which the derivative of the frame is shown to correspond to a set of geometric invariants.

For point motion in Euclidean space, the differential invariants correspond to curvature and torsion defined in a moving frame referred to as the \textit{Frenet-Serret frame}. Originally introduced in vision for characterizing and recognizing objects by their outlines \cite{Calabi1998,moons1995foundations}, the same concepts of curvature and torsion were applied for describing and recognizing point motion trajectories \cite{wu2008signature,WuLi2009}.

For rigid-body motion, differential invariants were introduced based on the concept of the screw axis in screw theory \cite{DeSchutter2010}. A moving frame was defined on the screw axis of which the motion was described by a set of invariants referred to as \textit{screw invariants}. These invariants are \textit{bi-invariant}, meaning independent of both the world reference frame and the reference frame attached to the body. An alternative invariant descriptor was proposed by defining two separate Frenet-Serret frames for the position and orientation of the rigid body \cite{lee2017bidirectional,vochten2015}, resulting in a simpler set of formulas. However, invariance for the choice of reference point for translation (the body frame's origin) was lost.

Invariant trajectory descriptors for motion have, for the large majority, been applied in motion recognition \cite{WuLi2009,lee2017bidirectional,vochten2015}. Trajectory generation using invariant descriptors has also been explored. The invariant descriptors are capable of both reconstructing the original trajectory \cite{DeSchutter2010}, as well as reproducing the trajectory at new locations and in different directions by applying the correct initial values when integrating the differential equations \cite{WU2010204}. Building further on that, trajectory adaptation methods based on optimal control and invariant descriptors were proposed \cite{vochten2019generalizing}. The goal was to adapt a reference trajectory to comply with new constraints while minimizing deviation from the invariant descriptor of the reference trajectory. This was shown to result in excellent extrapolation capabilities \cite{vochten2019generalizing}.

Calculating differential invariants from measurement data is challenging due to noise-sensitivity and singularities. \textit{Singularities} are defined as instances along the trajectory where components of the descriptor are arbitrarily defined. Near these singularities, small variations in the trajectory (e.g., due to measurement noise or irrelevant human variations) will result in large variations in the ill-defined components of the descriptor. To remedy this, efforts have been spent to develop robust approaches to calculate invariant descriptors, ranging from discrete approximations of the descriptor \cite{Calabi1998,WuLi2009,lee2017bidirectional,wu2008signature} to averaging the descriptor over segments of the curve \cite{arn2018motion}. More recently, approaches based on optimal control were introduced for calculating point trajectory descriptors \cite{perantoni2015optimal} and rigid-body trajectory descriptors \cite{vochten2018}. Using optimal control, the local descriptor is calculated in a window of measurements while regularization terms deal with singularities and noise by smoothing the invariants in a local neighborhood. A disadvantage of current optimal control-based approaches is their dependency on many tuning parameters for weighting the trajectory accuracy cost and regularization costs, and their long calculation times.

\subsection{Paper objective and contributions}

Up till now, invariant trajectory descriptors for interaction forces with similar benefits as the invariant descriptors for motion have not been proposed. This paper's main objective is to extend the existing concepts and methodology for motion invariants towards interaction forces by exploiting duality between motion and force. The resulting invariant descriptors can be used to model motion and force trajectories that occur in motion-in-contact tasks. Table~\ref{tab:table_trajs} summarizes this duality for the two cases where motion and force trajectories are either represented with vector coordinates or with screw coordinates.

\begin{table}[!t]
	\caption{Overview of trajectory types of which invariant descriptors can be calculated.}
	\label{tab:table_trajs}
	\resizebox{\linewidth}{!}{
		\begin{tabular}{@{}llll@{}}
			\toprule
			\textbf{Trajectory}                          &                                                                       & \textbf{Applied to motion}                                                             & \textbf{Applied to force}                                                        \\ \midrule
			\multirow{3}{*}{\textbf{vectors} $\vect{c}$} & \multirow{4}{*}{\includegraphics[width=2cm]{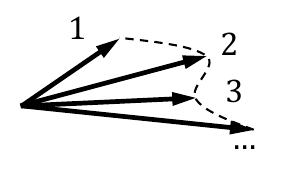}} & \multirow{2}{*}{rot. velocity $\vect{\omega}$}                                         & \multirow{2}{*}{force $\vect{f}$}                                                \\
			&                                                                       &                                                                                        &                                                                                  \\
			\multirow{1}{*}{$~~~\in \mathbb{R}^3$}      &                                                                       & \multirow{2}{*}{transl. velocity $\vect{v}$}                                           & \multirow{2}{*}{moment $\vect{m}$}                                               \\
			&                                                                       &                                                                                        &                                                                                  \\ \midrule
			\multirow{3}{*}{\textbf{screws} $\screw{s}$} & \multirow{4}{*}{\includegraphics[width=2cm]{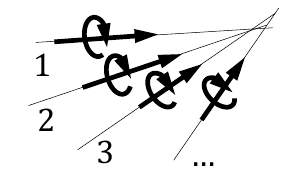}}  & \textit{twist}:                                                                        & \textit{wrench}:                                                                 \\
			&                                                                       & \multirow{3}{*}{ $\screw{t}= \begin{pmatrix} \vect{\omega} \\ \vect{v} \end{pmatrix}$} & \multirow{3}{*}{ $\screw{w}=\begin{pmatrix} \vect{f} \\ \vect{m} \end{pmatrix}$} \\
			\multirow{1}{*}{$~~~\in \mathbb{R}^6$}      &                                                                       &                                                                                        &                                                                                  \\ & & & \\
			\bottomrule
		\end{tabular}
	}
\end{table}

The main contribution of this paper is the introduction of two types of local differential invariant descriptors for force and moment trajectories. The first type is referred to as the \textit{vector invariant descriptor} for vector trajectories and is applied to force and moment vectors separately. The second type is referred to as the \textit{screw axis invariant descriptor} for screw trajectories and is applied to a screw wrench, combining force and moment in one entity. The duality is shown between the new invariant descriptors for force and moment and the existing invariant descriptors for motion: the screw invariants for motion \cite{DeSchutter2010} and the vector invariants for translation and rotation \cite{lee2017bidirectional,vochten2015}.
Furthermore, we explain how to calculate the invariant descriptors for motion and force using numerical optimal control schemes, which filter out the effects of measurement noise and improve the behavior of the descriptor near singularities. Triggered by the higher noise levels of measured force and moment trajectories compared to measured motion trajectories, we improved the formulation of these numerical optimal control schemes compared to \cite{vochten2018}, resulting in much more intuitive and faster tuning as well as increased robustness of the calculation. The invariant properties of the proposed descriptors were experimentally verified using a human-demonstrated 3D contour following task and a peg-on-hole alignment task. All software used in the validation experiment, including data, has been made publicly available \cite{software}.

The remainder of the paper is structured as follows. Section~\ref{sec:preliminaries} provides the necessary background on rigid-body kinematics and statics. Sections~\ref{sec:invars_vec} and \ref{sec:invars_screw} introduce local invariant descriptors in a general form, resulting in existing descriptors when applied to motion trajectories and new descriptors when applied to force and moment trajectories. Sections~\ref{sec:invars_vec} and \ref{sec:invars_screw} deal with vector and screw invariant descriptors, respectively. Section~\ref{sec:experiments} highlights a practical 3D contour following task, while Section~\ref{sec:experiments2} examines a peg-on-hole alignment task.  Section~\ref{sec:conclusion} provides a discussion about the benefits and limitations of the proposed concepts and methodology, it points at future work and presents a conclusion.

\section{Preliminaries}
\label{sec:preliminaries}

This section reviews essential background and introduces notation as summarized in Table~\ref{tab:notation}.

\begin{table}
	\centering
	\caption{Overview of notation used throughout the paper. (In this table $a$ and $b$ are placeholders for two frames of interest.)}
	\label{tab:notation}
	\renewcommand{\arraystretch}{1.2}
	\resizebox{\linewidth}{!}{
		\begin{tabular}{cp{0.85\columnwidth}}
			\toprule
			\textbf{Variable} & \multicolumn{1}{c}{\textbf{Description}}                                                                                                                                                                   \\
			\toprule
			$\refframe{a}$
			                  & reference frame named \textit{``a''} attached to a rigid body or the world                                                                                                                                 \\
			$\R{b}{a}$
			                  & 3$\times$3 rotation matrix representing the orientation of frame $\refframe{b}$ with respect to frame $\refframe{a}$                                                                                       \\
			$\T{b}{a}$
			                  & 4$\times$4 homogeneous transformation matrix representing the orientation and position of frame $\{b\}$ with respect to frame $\{a\}$                                                                      \\
			$_{a}\vect{\omega}$
			                  & 3$\times$1 rotational velocity vector of the rigid body expressed in $\refframe{a}$                                                                                                                        \\
			$_{a}\vect{v}^{b}$
			                  & 3$\times$1 translational velocity vector of the rigid body at the origin of $\refframe{b}$ and expressed in $\refframe{a}$                                                                                 \\
			$_{a}\vect{f}$
			                  & 3$\times$1 force vector acting on the rigid body expressed in $\refframe{a}$                                                                                                                               \\
			$_{a}\vect{m}^{b}$
			                  & 3$\times$1 moment vector acting on the rigid body at the origin of $\refframe{b}$ and expressed in $\refframe{a}$                                                                                          \\
			$_{a}\screw{t}^{a}$
			                  & 6$\times$1 screw twist of rigid body with the first three elements the rotational velocity vector $_{a}\vect{\omega}^{}$ and the last three elements the translational velocity vector  $_{a}\vect{v}^{a}$ \\
			$_{a}\screw{w}^{a}$
			                  & 6$\times$1 screw wrench acting on the rigid body with the first three elements the force vector $_{a}\vect{f}^{}$ and the last three elements the moment vector  $_{a}\vect{m}^{a}$                        \\
			$ {}_a^b\mat{S}$
			                  & 6$\times$6 screw transformation matrix that, when multiplied with a screw twist or wrench, transforms the screw from $\refframe{b}$ to $\refframe{a}$                                                        \\
			\bottomrule
		\end{tabular}%
	}
\end{table}

\subsection{Rigid-body kinematics}

The motion trajectory of a rigid body is commonly represented by attaching a reference frame $\refframe{b}$ to the \textit{body} and expressing the position and orientation of this frame as a function of time $t$ with respect to a fixed reference frame $\refframe{w}$, also referred to as the \textit{world} reference frame. For all variables introduced below, the explicit dependency on time $t$ is omitted to simplify the expressions.

The position is represented by the 3-dimensional \textit{displacement vector} $\pvector{w}{b}$ from the origin of $\refframe{w}$ to the origin of $\refframe{b}$. The orientation is represented by the \textit{rotation matrix} $\R{b}{w}$, which expresses the coordinates of the unit vectors of the moving body frame $\refframe{b}$ with respect to the world frame $\refframe{w}$:
\begin{equation}
	\R{b}{w} =
	\begin{bmatrix}
		\prescript{}{}{\vect{e}}_x & \prescript{}{}{\vect{e}}_y & \prescript{}{}{\vect{e}}_z
	\end{bmatrix}.
\end{equation}

Position and orientation can be combined into the \textit{homogeneous transformation matrix} or \textit{pose matrix} $\T{b}{w}$:
\begin{equation}
	\T{b}{w} =
	\begin{bmatrix}
		\R{b}{w}         & \pvector{w}{b} \\
		\zerom{1\times3} & 1
	\end{bmatrix},
	\label{eq:T}
\end{equation}
which is part of the $SE(3)$ group according to Lie theory. The coordinates of the pose matrix $\T{b}{w}$ depend on the choice of the fixed reference frame $\refframe{w}$ in the world and the reference frame $\refframe{b}$ on the rigid body.

The velocity of a rigid body can be fully characterized by two 3-dimensional vectors: the rotational velocity vector $\omegavector{}{}{}$ and the translational velocity vector $\velvector{}{}{}{}$ of the origin of the considered reference frame. These vectors are summarized with a 6-dimensional screw \cite{ball1900treatise}, the \textit{twist} $\screw{t} = \begin{pmatrix} \omegavector{}{}{} \\ \velvector{}{}{}{} \end{pmatrix}$.

To define the time-derivative of the homogeneous transformation matrix, $\dot{\T{}{}} = \frac{\textup{d}\mat{T}}{\textup{d}t}$, we first introduce the operator $\left[\cdot\right]_\times$. When applied to the vector $\vect{\omega}=[\omega_x ~ \omega_y ~ \omega_z]^T$, it results in a $3\times3$ skew-symmetric matrix:
\begin{equation}
	\left[\omegavector{}{}{}\right]_{\times} =
	\begin{bmatrix}
		0         & -\omega_z & \omega_y  \\
		\omega_z  & 0         & -\omega_x \\
		-\omega_y & \omega_x  & 0
	\end{bmatrix},
\label{eq:skew}
\end{equation}
and when applied to a screw $\screw{t}$, it results in the following $4\times4$ \textit{twist matrix} $\left[\screw{t}\right]_{\times}$, which is an element of the Lie algebra $se(3)$:
\begin{equation}
	\left[\screw{t}\right]_{\times} = \begin{bmatrix}
		\left[\omegavector{}{}{}\right]_{\times} & \velvector{}{}{}{} \\ \zerom{1\times3} & 0
	\end{bmatrix}.
\label{eq:skew2}
\end{equation}
\begin{figure}[t]
	\centering
	\subfloat[Screw twists]{\includegraphics[width=0.475\linewidth]{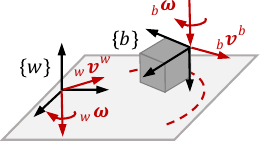}%
		\label{fig:screw_twist}}
	\hfill
	\subfloat[Screw wrenches]{\includegraphics[width=0.475\linewidth]{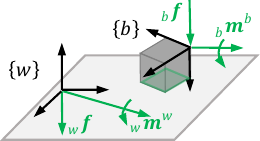}%
		\label{fig:screw_wrench}}
	\caption{Visualization of screw twist $\screw{t}=(\vect{\omega}^T ~ \vect{v}^T)^T$ and screw wrench $\screw{w}=(\vect{f}^T ~ \vect{m}^T)^T$  in the world frame $\refframe{w}$ and in the body frame $\refframe{b}$.}
	\label{fig:screws}
\end{figure}
The definition of the time-derivative of the homogeneous transformation matrix $\dot{\T{}{}}$ depends on the choice of reference frame in which the twist $\screw{t}$ is expressed.
Choosing the body frame $\refframe{b}$, the \textit{body twist} ${}_b\screw{t}^b{}$ contains the rotational velocity vector $\omegavector{}{}{b}$ expressed in $\refframe{b}$ and the translational velocity vector $\velvector{b}{}{b}{}$ of the origin of $\refframe{b}$ and expressed in $\refframe{b}$ (see Fig.~\ref{fig:screw_twist}). The derivative of the pose matrix $\Tdot{b}{w}$ is then defined as:
\begin{align}
	\Tdot{b}{w}
	= \T{b}{w} \left[{}_b\screw{t}^b{}\right]_{\times},
	~\text{with}~~
	\left[{}_b\screw{t}^b{}\right]_{\times} =
	\begin{bmatrix}
		\left[\omegavector{}{}{b}\right]_{\times} & \velvector{b}{}{b}{}
		\\ \zerom{1\times3} & 0
	\end{bmatrix}.
	\label{eq:Tdot}
\end{align}
The twist ${}_b\screw{t}^b{}$ has the property of being invariant for a change of $\refframe{w}$ (\textit{left-invariance}).

Alternatively, choosing the world frame $\refframe{w}$, the spatial twist $_w\screw{t}^w$ consists of the rotational velocity vector $\omegavector{}{}{w}$ expressed in $\refframe{w}$, and the translational velocity vector $\velvector{w}{}{w}{}$ of the point on the rigid body that instantaneously coincides with the origin of $\refframe{w}$ expressed in $\refframe{w}$. The pose derivative $\Tdot{b}{w}$ is then defined as:
\begin{align}
	\Tdot{b}{w}  = \left[_w\screw{t}^w\right]_{\times} \T{b}{w},
	~\text{with}~~
	\left[_w\screw{t}^w\right]_{\times} = \begin{bmatrix}
		                                      \left[\omegavector{}{}{w}\right]_{\times} & \velvector{w}{}{w}{} \\ \zerom{1\times3} & 0
	                                      \end{bmatrix}.
	\label{eq:Tdotleft}
\end{align}
where $\left[_w\screw{t}^w\right]_{\times}$ is now multiplied on the left of $\T{b}{w}$ since the twist is expressed in $\refframe{w}$. The twist ${}_w\screw{t}^w{}$ has the property of being invariant for a change of $\refframe{b}$ (\textit{right-invariance}).

The relation between twists $_w\screw{t}^w$ and $_b\screw{t}^b$ is provided by the $6\times6$ \textit{screw transformation matrix} $\mat{S}(\T{b}{w})$ (also referred to as the \textit{adjoint matrix} \cite{murray1994}), which is constructed from elements of the pose matrix $\T{b}{w}$:
\begin{align}
	\label{eq:screwtransformation}
	_w\screw{t}^w & = \mat{S}(\T{b}{w}) \ _b\screw{t}^b  = \begin{bmatrix}
		                                                       \R{b}{w}                          & \zerom{3\times3} \\
		                                                       [\pvector{w}{b}]_{\times}\R{b}{w} & \R{b}{w}
	                                                       \end{bmatrix} \ _b\screw{t}^b.
\end{align}
Twists $_b\screw{t}^b$ and $_w\screw{t}^w$ represent first-order kinematics of the trajectory and are invariant to the choice of world frame and body frame, respectively. But neither $_w\screw{t}^w$ nor $_b\screw{t}^b$ is invariant for the choice of \textit{both} reference frames $\refframe{w}$ and $\refframe{b}$. Trajectory descriptors that are invariant to \textit{both} reference frames (also referred to as coordinate-invariance) will be discussed in Sections~\ref{sec:invars_vec} and \ref{sec:invars_screw}.

\subsection{Duality between twist and wrench as general screws}

All forces and moments acting on a rigid body can be reduced to a single resultant force vector and a single resultant moment vector. One option is to express these vectors in the body frame: ${}_{b}\vect{f}$ represents the resultant force expressed in $\refframe{b}$, and ${}_{b}\vect{m}^b$ represents the moment with respect to the origin of $\refframe{b}$ and with coordinates expressed in $\refframe{b}$ (see Fig.~\ref{fig:screw_wrench}). Similar to the \textit{twist}, a compact notation for the resulting force and moment is the 6-dimensional \textit{screw wrench}:
\begin{equation}
	\label{eq:wrench}
	_b\screw{w}^b =
	\begin{pmatrix}
		{}_{b}\vect{f} \\
		{}_{b}\vect{m}^b
	\end{pmatrix}.
\end{equation}
Transforming the wrench from the body frame $\refframe{b}$ to the world frame $\refframe{w}$ is done using the same $6\times6$ \textit{screw transformation matrix} $\mat{S}(\T{b}{w})$ as for the twist:
\begin{align}
	\label{eq:screw_transformtation}
	_w\screw{w}^w & = \mat{S}(\T{b}{w}) \ _b\screw{w}^b.
\end{align}
This analogy between twists and wrenches is also referred to as the \textit{duality between velocity and force}. Both twist and wrench can be interpreted as a 6-dimensional \textit{screw} \cite{ball1900treatise}, meaning they can always be represented by a vector lying along a line in space and the moment of that vector around the line.

\subsection{Time-invariance using a geometric progress variable}
\label{subsec:progress}

In the previous subsections, motion and force trajectories were assumed to be a function of time $t$. To obtain a \textit{time-invariant trajectory representation}, i.e., a trajectory representation that is invariant with respect to the motion profile, a geometric progress variable $\xi$ must be chosen.

For tasks where translation is dominant, such as in contour following, the arc length is a natural choice. The arc length is calculated by integrating the norm of the translational velocity vector $\vect{v}$ over time. For tasks where rotation is dominant, such as in opening a door, a natural progress variable would be the integral of the norm of the rotational velocity vector $\vect{\omega}$. If translation and rotation are both important, then a progress parameter can be defined combining the translation and rotation, for example, using a linear combination \cite{DeSchutter2010}:
\begin{equation}
	\xi(t) = \int_{0}^{t} w \norm{\vect{\omega}} + (1-w) \norm{\vect{v}} \textup{d}t,
\end{equation}
where $w$ weights the relative contribution. In \cite{DeSchutter2010}, further details are provided for obtaining a progress value that is invariant for the choice of reference point on the body for the translation $\vect{v}$.

For the pose $\mat{T}$ and wrench $\screw{w}$, the reparameterization from time $t$ to progress $\xi$ is straightforward:
\begin{align}
	\T{}{}(\xi) = \T{}{}( t(\xi) ) ~~~\text{and}~~~	{}_{}^{}\screw{w}(\xi)   = {}_{}^{}{\screw{w}}( t(\xi) ).
	\label{eq:reparam}
\end{align}
In practice, since measurement data is often only available in a discretely sampled form, the reparameterization is achieved through numerical interpolation.

For the velocity, the reparameterization from time $t$ to progress $\xi$ requires an additional step since velocities represent derivatives of the trajectory. As an example, the reparameterization of the translational velocity vector $\vect{v}$, as the derivative of the position $\vect{p}$, can be worked out as:
\begin{equation}
	\vect{v}(\xi) = \frac{\textup{d}\vect{p}(\xi)}{\textup{d}\xi} = \frac{\textup{d} \vect{p}(t(\xi))}{\textup{d}t(\xi)} \cdot \frac{\textup{d}t(\xi)}{\textup{d}\xi} =  \frac{\vect{v}(t(\xi))}{\dot{\xi}(t(\xi))},
\end{equation}
where the chain rule of differentiation was applied in the second equality to obtain the relation with time $t$, and the \textit{progress rate} $\dot{\xi}=\frac{\textup{d}\xi}{\textup{d}t}$ is the derivative of the progress with respect to time. Similar derivations can be made for the rotational velocity vector $\vect{\omega}$ and the twist $\screw{t}$:
\begin{align}
	\vect{\omega}(\xi) = \frac{\vect{\omega}( t(\xi) )}{\dot{\xi}( t(\xi) )}\text{~~~and~~~} \screw{t}(\xi) = \frac{\screw{t}( t(\xi) )}{\dot{\xi}( t(\xi) )}.
	\label{eq:reparam4}
\end{align}

Equations \eqref{eq:reparam}-\eqref{eq:reparam4} can be used in the other direction to apply a motion profile $\xi(t)$ and obtain the pose, wrench, translational velocity, rotational velocity or twist as a function of time. This is relevant, for example, when a trajectory was planned in the progress domain $\xi$ and needs to be executed with a specific timing ${\xi(t)}$. A special choice for the progress variable is one where $\xi=0$ at the beginning of the trajectory and $\xi=1$ at the end of the trajectory. This results in a reparameterization that is independent of the scale of the trajectory \cite{DeSchutter2010}.

\section{Invariant descriptor for vector trajectories}
\label{sec:invars_vec}

This section derives a local invariant descriptor for a general \textit{vector trajectory} $\vect{c}(\xi)$. The trajectory $\vect{c}(\xi)$ is assumed to be time-invariant, i.e. parameterized as a function of a geometric progress $\xi$ as discussed in Section~\ref{subsec:progress}.
To keep the formulas concise, the explicit dependency on $\xi$ is omitted in the following formulas. The derivative of $\vect{c}$ with respect to the progress $\xi$ is represented using the prime notation: $\vect{c}' = \frac{\textup{d}\vect{c}}{\textup{d}\xi}$.

\subsection{Definition of invariants}

The vector invariants are based on a generalization of the \textit{Frenet-Serret equations} for translation \cite{gray2006modern}, where $\vect{c}$ corresponded to the translational velocity vector. While in \cite{vochten2015,lee2017bidirectional} the Frenet-Serret equations were extended to rotational velocity trajectories, the aim here is to exploit the duality between velocity and force so that the invariants can be presented in a general form, applicable to rotational and translational velocity trajectories, as well as to force and moment trajectories.

\begin{figure}[t]
	\centering
	\subfloat[Object invariant along tangent]{\includegraphics[width=0.44\linewidth]{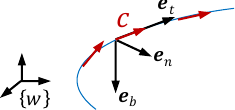}%
		\label{fig:vector_b}}
	\hfil
	\subfloat[Moving frame invariants]{\includegraphics[width=0.44\linewidth]{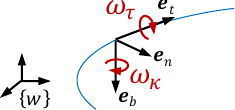}%
		\label{fig:vector_c}}
	\caption{Invariant descriptor for a vector trajectory: (a) object invariant $c$ defined in the moving FS frame given by the unit vectors $\vect{e}_t$, $\vect{e}_n$ and $\vect{e}_b$; (b) moving frame invariants $\omega_\tau$ and $\omega_\kappa$. The figure depicts the special case where $\vect{c}$ corresponds to a translational velocity vector with respect to reference frame $\{w\}$ and where, for visualization purposes, the moving frame is depicted with its origin chosen at the corresponding position along the point trajectory (indicated in blue).}
	\label{fig:vector}
\end{figure}

The vector invariants are defined and expressed in a local moving frame, referred to as the \textit{Frenet-Serret frame} (FS frame), which is constructed as follows.
Its first axis, the tangent $\vect{e}_t$, corresponds to the unit vector in the direction of $\vect{c}$. The second axis corresponds to the normal $\vect{e}_n$ of the trajectory. The third axis, the binormal $\vect{e}_b$, follows directly from the cross-product of $\vect{e}_t$ and $\vect{e}_n$. These unit vectors are defined as follows from $\vect{c}$ and $\vect{c}'$\footnote{
	To be consistent with the common notation of the FS frame and
	equations,
	the authors adapted the order of the columns $\vect{e}_t$, $\vect{e}_n$, and $\vect{e}_b$ of
	$\widetilde{\boldsymbol{R}}$ in their
	previous papers such as
	\cite{DeSchutter2010,vochten2018,vochten2019generalizing}. Therefore, this paper has a slightly different notation than the authors' previous works. This adaptation affects only the appearance of the equations, not the concept.
}:
\begin{equation}
	\vect{e}_t = \frac{\vect{c}}{\norm{\vect{c}}},
	~~~\vect{e}_n = \frac{ (\vect{c} \times \vect{c}') \times \vect{c}}{\norm{(\vect{c} \times \vect{c}') \times \vect{c}}},
	~~~\vect{e}_b = \vect{e}_t \times \vect{e}_n.
	\label{eq:tnb1}
\end{equation}
In differential geometry, the linear span of $\vect{e}_t$ and $\vect{e}_n$ is referred to as the \textit{osculating plane}, i.e. the plane in which the vector $\vect{c}$ rotates instantaneously, while $\vect{e}_b$ represents the axis about which this rotation occurs.
The complete orientation of the FS frame is expressed using a rotation matrix $\matmf{R}$:
\begin{equation}
	\matmf{R} =
	\begin{bmatrix}
		\vect{e}_t & \vect{e}_n & \vect{e}_b
	\end{bmatrix},
	\label{eq:tnb}
\end{equation}
where the tilde $\sim$ indicates it is a local moving frame, in this case the FS frame. See Fig. \ref{fig:vector} for a visualization of the frame.

The invariants are then defined as follows. The first invariant $c$ is the magnitude of the vector $\vect{c}$ along the tangent:
\begin{equation}
	c = \vect{c} \cdot \vect{e}_t.
	\label{eq:vec_invars_obj}
\end{equation}
This invariant is referred to as the \textit{object invariant} since it relates to the object. For example, $c$ can represent the magnitude of the translational or rotational velocity vector of the object, or the magnitude of the force or moment vector applied to the object. The vector $\vect{c}$ can be retrieved from the object invariant $c$ using:
\begin{equation}
	\vect{c} =  \matmf{R}~ \vectmf{c}, ~~~\text{with}~~~\vectmf{c} =  \begin{pmatrix}
		c & 0 & 0
	\end{pmatrix}^{\text{T}}.
	\label{eq:FS_obj}
\end{equation}
The second and third invariants $\omega_{\kappa}$ and $\omega_{\tau}$ govern the first-order kinematics of the local FS frame\footnote{The relation between \eqref{eq:kin_R} and the Frenet-Serret equation as usually presented in literature is derived in the appendix.}:
\begin{equation}
	\label{eq:kin_R}
	\matmf{R}' = \matmf{R}
	\begin{bmatrix}
		0               & -\omega_{\kappa} & 0              \\
		\omega_{\kappa} & 0                & -\omega_{\tau} \\
		0               & \omega_{\tau}    & 0
	\end{bmatrix},
\end{equation}
where $\omega_{\kappa}$ is known as the curvature rate and $\omega_{\tau}$ as the torsion rate, with the following explicit formulas \cite{gray2006modern}:
\begin{equation}
	\omega_{\kappa} = \frac{||\vect{c} \times \vect{c}'||}{{||\vect{c}||}^2}, ~~~~
	\omega_{\tau} = \frac{((\vect{c} \times \vect{c}') \times (\vect{c} \times \vect{c}'')) \cdot \vect{c}}{{||\vect{c} \times \vect{c}'||}^2 ||\vect{c}||}.
	\label{eq:vec_invars_mf}
\end{equation}
For a naturally parameterized point trajectory, with $\vect{c}$ the point's velocity, $\omega_{\kappa}$ and $\omega_{\tau}$ correspond to the local curvature $\kappa$ and torsion $\tau$ of the point's trajectory.

Equation \eqref{eq:kin_R} can be written compactly using \eqref{eq:skew}:
\begin{equation}
	\label{eq:kin_R2}
	\matmf{R}' = \matmf{R}\begin{bmatrix} \vect{i} \end{bmatrix}_\times,
	~~~\text{with}~~~ \vect{i} = \begin{pmatrix}
		\omega_\tau \\ 0 \\ \omega_\kappa
	\end{pmatrix},
\end{equation}
where $\vect{i}$ is referred to as the \textit{moving frame invariant} since it models the first-order kinematics of the moving frame.

Only the orientation is relevant for the FS frame, not its position. Therefore it is sometimes referred to as an \textit{orientation frame} \cite{delaet2013}. For visualization purposes, its evolution is often shown by representing it at the point on the motion trajectory associated with the progress variable $\xi$. This was done in Fig.~\ref{fig:vector}, which shows the definition of the moving frame and the corresponding invariants for a point trajectory.

\textbf{Singularities}: Singularities are defined as instances where axes of the moving frame and the corresponding invariants are not uniquely defined anymore.
A first type of singularity is when  $\vect{c} = \vect{0}$. The object invariant $c$ is then zero, while the moving frame invariants $\omega_\kappa$ and $\omega_\tau$ are undefined. In other words, the complete moving frame is arbitrary.

A second type of singularity is when the vector $\vect{c}$ has an unchanging orientation at some instance ($\vect{c} \times {\vect{c}}' = \vect{0}$), remaining parallel to itself. The first moving frame invariant $\omega_\kappa$ is then zero while the second moving frame invariant $\omega_\tau$ is undefined. In other words, the tangent of the moving frame is well-defined, but the normal and binormal axes are arbitrary.

\textbf{Invariant properties}:
If the trajectory $\vect{c}(\xi)$ is expressed as a function of a geometric progress variable $\xi$ as explained in Section~\ref{subsec:progress}, then the descriptor ($c$,~$\vect{i}$) is also \textit{time-invariant}.
Furthermore, the descriptor is \textit{invariant with respect to the choice of world frame} $\refframe{w}$ (both origin and orientation) in which the coordinates of the given vector trajectory $\vect{c}(\xi)$ are expressed. It is evident that a change of $\refframe{w}$ merely results in the same change of orientation for the vectors $\vect{c}$, $\vect{c}'$, and $\vect{c}''$, resulting in the same object invariant $c$, the same FS frame for all $\xi$, and hence also the same moving frame invariant $\vect{i}$. If the coordinates of $\vect{c}$ are instead expressed in $\refframe{b}$, the descriptor is still \textit{invariant for changes in orientation of the body frame} $\refframe{b}$. However, not all considered instances of the descriptor are invariant with respect to the origin of body frame $\refframe{b}$. As pointed out in \cite{vochten2015}, the translational velocity vector $\vect{v}$ depends on the reference point on the moving object chosen to express its velocity, hence on the origin of $\refframe{b}$. Similarly, the moment vector $\vect{m}$ depends on the reference point for expressing the moment acting on the object, hence on the origin of $\refframe{b}$. The dependency of $\vect{v}$ and $\vect{m}$, and hence also of their invariant descriptors, on the origin of $\refframe{b}$ is a disadvantage compared to the screw invariants discussed in Section~\ref{sec:invars_screw}, which do not exhibit this dependency on the chosen reference point on the moving body.

\textbf{Trajectory reconstruction}:
Reconstructing the trajectory $\vect{c}(\xi)$ from the invariant descriptor $(\vectmf{c}(\xi),\vect{i}(\xi))$ requires integrating the differential equation \eqref{eq:kin_R2} from an initial value of the moving frame $\matmf{R}$. In general, there exists no closed-form solution for the integration. Instead, the problem is discretized so that $\matmf{R}_{k+1}$ at sample $k+1$ is found from $\matmf{R}_{k}$ at sample $k$ as follows:
\begin{equation}
	\matmf{R}_{k+1} = \matmf{R}_{k} \ \exp\left(  [\vect{i}_k]_{\times} \Delta \xi \right),
	\label{eq:FS_rec}
\end{equation}
where we assume that $\vect{i}_k$ remains constant over the integration step $\Delta \xi$. The matrix exponential operator $\exp(\cdot)$ maps the skew-symmetric matrix $ [\vect{i}_k]_{\times} \Delta \xi$, which is part of the Lie algebra $so(3)$, into the corresponding change in orientation, which is part of the Lie group $SO(3)$. An explicit expression of the matrix exponential can be formulated using Rodrigues' rotation formula \cite{murray1994}. Given the initial value $\matmf{R}_{1}$ at $k=1$, the moving frame can be reconstructed for all remaining samples $k=2...N$. From the moving frames $\matmf{R}_k$, the corresponding vector trajectory $\vect{c}_k$ at $k:1...N$ is found using \eqref{eq:FS_obj}:
\begin{equation}
	\vect{c}_k = \matmf{R}_k ~ \vectmf{c}_k, ~~\forall k \in [1,N].
	\label{eq:FS_rec2}
\end{equation}

Equations \eqref{eq:FS_rec} and \eqref{eq:FS_rec2} are referred to as the \emph{moving frame equation} and \emph{object trajectory equation}, respectively. Table~\ref{tab:invars_vec_and_screw} on page~\pageref{tab:invars_vec_and_screw} provides an overview of the vector invariants framework with application to translational velocity $\vect{v}$, rotational velocity $\vect{\omega}$, force $\vect{f}$, and moment $\vect{m}$ trajectories.

\subsection{Robust calculation of invariants using optimal control}
\label{sec:robust_calc_vector}

Using the analytical formulas to calculate the vector invariants ${c}$ and $\vect{i}$ has three pitfalls: (1) near singularities, some of the invariants and axes of the moving frame $\matmf{R}$ are ill-defined, hence the behavior of ${c}$ and $\vect{i}$ is not well-captured in such case; (2) the high-order derivatives cause sensitivity to noise, especially near singularities; and (3) the open-loop reconstructed trajectory using \eqref{eq:FS_rec}-\eqref{eq:FS_rec2} is not guaranteed to match the original trajectory due to integration drift.

 In addition, the invariants $c$ and $\omega_{\kappa}$ are always positive by definition from \eqref{eq:vec_invars_obj}-\eqref{eq:vec_invars_mf} resulting in \textit{frame flipping} issues: the FS frame must flip $180^\circ$ when the direction of $\vect{c}$ reverses or when the curve traced by the vector $\vect{c}$ has an inflection point. Other definitions of moving frames exist to minimize the frame's rotation, such as Bishop frames \cite{bishop1975there}, but these frames are not locally defined anymore.

To deal with these problems, we formulate an \textit{Optimal Control Problem} (OCP) to calculate the vector invariants ${c}$ and $\vect{i}$ in a robust way. An OCP is a type of optimization problem that, for a given dynamical system, tries to find control inputs that minimize a specific cost function over a horizon. Here, the objective of the optimization problem is to find vector invariants that reconstruct the given measured vector trajectory. The latter can be quantified by introducing a trajectory reconstruction cost $\Delta \vect{c}_{\text{MSE}}$ containing the mean-squared error (MSE) between the reconstructed trajectory $\vect{c}_k$ and measured trajectory $\vect{c}^{meas}_k$ for $k=1...N$:
\begin{equation}
	\Delta \vect{c}_{\text{MSE}}(\vect{c},\vect{c}^{meas}) = \frac{1}{N} \sum\limits_{k=1}^{N}{\norm{\vect{c}_k -  \vect{c}^{meas}_k}^2}.
	\label{eq:obj_meas}
\end{equation}

To deal with measurement noise and the effects of singularities, a regularization cost is introduced.
In general, the regularization cost can be any function of invariants, including derivatives of invariants. In \cite{vochten2018}, the terms in the regularization cost were chosen as the derivatives of all invariants and the absolute values of the moving frame invariants. The trajectory reconstruction cost and regularization cost were added in the objective function of the OCP, and as a result they had to be weighted with appropriate weighting factors, requiring an extensive tuning process.

To reduce tuning efforts and improve interpretability of chosen parameters, this paper proposes a new formulation of the OCP.  The trajectory reconstruction cost is moved to the constraints of the OCP and is bounded by a chosen tolerance value $\epsilon$. For vector trajectories, this is worked out as follows:
\begin{equation}
	\epsilon_c^2 \geq \Delta \vect{c}_{\text{MSE}}(\vect{c},\vect{c}^{meas}),
	\label{eq:traj_recon_cons}
\end{equation}
in which $\epsilon_c$ is interpreted as the desired tolerance on the trajectory reconstruction cost over the entire horizon. The subscript in $\epsilon_c$ refers to the type of trajectory that is considered. The value of $\epsilon$ can be set by the user based on the expected accuracy of the measurement system or the tolerable reconstruction error for the given application.

For the regularization term in the objective function, we take the squared norm of the moving frame invariants $\vect{i}^{}$. Since $\vect{i}^{}$ signifies the rotational change of the moving frame, this will result in a more stable and smooth evolution of the moving frame $\matmf{R}$. To avoid frame flipping issues, all invariants are allowed to become positive or negative, resulting in a smoother evolution of the FS frame.

The complete OCP to calculate the vector invariants $\vectmf{c}$ and $\vect{i}$ from a given sequence of $N$ measured vectors $\vect{c}^{meas}_k$ is:
\begin{align}
	\begin{aligned}
		\underset{%\begin{subarray}{l}
		\widetilde{c}_{[\cdot]}, \vect{i}_{[\cdot]},
		\vect{c}_{[\cdot]}, \matmf{R}_{[\cdot]}
		%\end{subarray}
		}{\text{minimize}} ~~
		 & \sum\limits_{k=1}^{N-1}{\norm{ \vect{i}^{}_k}^2_{} },
		\label{eq:obj_vec}
	\end{aligned}
\end{align}
\text{subject to:~}
\begin{alignat}{2}
	\epsilon_c^2                    & \geq \Delta \vect{c}_{\text{MSE}}(\vect{c},\vect{c}^{meas}), \tag{r.\ref{eq:traj_recon_cons}}                                                   \\
	\matmf{R}_{k+1}                 & = \matmf{R}_{k} \ \exp\left(  [\vect{i}_k]_{\times} \Delta \xi \right), \quad                 && \forall k \in [1,N] \tag{r.\ref{eq:FS_rec}}  \\
	\vect{c}_k                      & = \matmf{R}_k  \ \vectmf{c}_k,                                                                && \forall k \in [1,N] \tag{r.\ref{eq:FS_rec2}} \\
	\matmf{R}_{1}^{T} \matmf{R}_{1} & = \mat{I}_{3\times3}  \label{eq:orthog_vec}.
\end{alignat}
Besides the trajectory error constraint \eqref{eq:traj_recon_cons}, the constraints consist of the trajectory reconstruction equations \eqref{eq:FS_rec}-\eqref{eq:FS_rec2}, while constraint \eqref{eq:orthog_vec} imposes orthonormality conditions on the initial moving frame. The orthonormality of $\matmf{R}$ is preserved throughout the entire horizon through the matrix exponential integrator in \eqref{eq:FS_rec}.

\textbf{Initialization}: The OCP \eqref{eq:obj_vec}-\eqref{eq:orthog_vec} is a non-convex optimization problem and therefore multiple local minima may exist. The solution will converge towards one of the local minima depending on the initialization of the unknown variables. To obtain invariant solutions regardless of the reference frame in which the variables are represented, the variables must be initialized in an invariant way.
One possibility is to calculate initial values for the invariants and moving frames using the analytical formulas \eqref{eq:tnb}-\eqref{eq:vec_invars_mf}, as was done in \cite{vochten2018}. However, this initialization then suffers from the same three pitfalls mentioned at the beginning of this subsection.

We propose an alternative initialization approach here based on calculating an average moving frame, first introduced in \cite{asapaper} for screw trajectories, but adapted here to vector trajectories in three steps. First, calculate the covariance matrix $\mat{C}_c$ associated with the measured vector trajectory $\vect{c}_k^{meas}$:
\begin{equation}
	\mat{C}_c = \frac{1}{N}\sum_{k=1}^{N}\vect{c}_k^{meas}\left(\vect{c}_k^{meas}\right)^\text{T}.
\end{equation}
Second, the orthogonal matrix of eigenvectors $\mat{U}_c$ is calculated from $\mat{C}_c$. This matrix $\mat{U}_c$ is invariant for changes in reference frame by definition and can be interpreted as a sort of average FS frame for the whole trajectory. The signs of the first two eigenvectors are chosen unambiguously by making the average value of the measurement vector $\vect{c}_k^{meas}$ positive along these directions. The third axis then follows from the cross product of the first two axes. All moving frames $\matmf{R}_k$ are initialized with the average moving frame $\mat{U}_c$.

Third, to be consistent with the average moving frame, the moving frame invariants $\omega_\kappa$ and $\omega_\tau$ are initialized with zero values. The object invariant $c_k$ is initialized by projecting the vector trajectory $\vect{c}_k^{meas}$ onto the average moving frame and selecting the component along the first axis for all samples $k$.

\textbf{Application to motion and force data}:
The calculation of vector invariants using OCP \eqref{eq:obj_vec}-\eqref{eq:orthog_vec} can directly be applied to translational velocity trajectories $\vect{v}(\xi)$, rotational velocity trajectories $\vect{\omega}(\xi)$, force trajectories $\vect{f}(\xi)$, and moment trajectories $\vect{m}(\xi)$.

However, in many applications involving rigid-body motion, the measured data corresponds to position and orientation coordinates of the rigid body, instead of translational and rotational velocity vectors. In this case, one option is to use the following two-step approach: (1) estimate the rigid-body velocities from the measured position and orientation coordinates; (2) apply OCP \eqref{eq:obj_vec}-\eqref{eq:orthog_vec} to the estimated velocity vectors. A downside of this first option is that integration of the optimized velocity vectors to reconstruct position and orientation coordinates will inevitably result in integration drift, such that there will be a deviation between the measured and reconstructed positions and orientations.

A second option is to adapt optimization problem \eqref{eq:obj_vec}-\eqref{eq:orthog_vec} to include position or orientation measurements. Below, this is worked out assuming the measured positions are given by position vectors and the measured orientations by rotation matrices. The proposed approach can be adapted to deal with other orientation representations such as quaternions.

\textbf{Including position measurements for case $\vect{c}=\vect{v}$}:
The reconstructed position vectors $\vect{p}_k$ at each sample are introduced as additional variables in the OCP. The trajectory reconstruction cost $\Delta \vect{c}_{\text{MSE}}(\vect{c},\vect{c}^{meas})$ in \eqref{eq:obj_meas} is now replaced by the MSE error between the reconstructed position $\vect{p}_k$ and the measured position $\vect{p}^{meas}_k$ over $k=1...N$:
\begin{equation}
	\label{eq:cost_position}
	\Delta \vect{p}_{\text{MSE}}(\vect{p},\vect{p}^{meas}) = \frac{1}{N} \sum\limits_{k=1}^{N}{\norm{\vect{p}_k -  \vect{p}^{meas}_k}^2}.
\end{equation}

The reconstructed positions are defined by extending the integrator in \eqref{eq:FS_rec} with a fourth row and column:
\begin{equation}
	\label{eq:integrator_position}
	\resizebox{.89\hsize}{!}{$
			\begin{bmatrix}
				\matmf{R}_{k+1}  & \vect{p}_{k+1} \\
				\zerom{1\times3} & 1
			\end{bmatrix} = \begin{bmatrix}
				\matmf{R}_{k}    & \vect{p}_{k} \\
				\zerom{1\times3} & 1
			\end{bmatrix} \ \exp\left(
			\begin{bmatrix}
				\left[\vect{i}_k\right]_\times & \vectmf{c}_k \\
				\zerom{1\times3}               & 0
			\end{bmatrix} \Delta \xi \right)$.
	}
\end{equation}
The effect of these additions is that $\vectmf{c}_k$, which in this case signifies the translational velocity in the moving frame, is used to update the position $\vect{p}_k$.

\textbf{Including orientation measurements for case $\vect{c} = \vect{\omega}$}:
The workflow is similar as for position. Variables for the reconstructed rotation matrices $\mat{R}_k$ at each sample are introduced. The trajectory reconstruction cost $\Delta \vect{c}_{\text{MSE}}(\vect{c},\vect{c}^{meas})$ in \eqref{eq:obj_meas} is replaced by a first-order approximation of the mean-squared rotation angle \cite{MetricsRotations2009} between the reconstructed rotation matrix $\mat{R}_k$ and the measured rotation matrix $\mat{R}^{meas}_{k}$ over $k=1...N$:
\begin{equation}
	\label{eq:cost_rotation}
	\scalemath{0.94}{\Delta \mat{R}_{\text{MSE}}(\mat{R},\mat{R}^{meas})} = \scalemath{0.94}{\frac{1}{N} \sum\limits_{k=1}^{N}{\norm{\left(\mat{R}^{meas}_{k}\right)^T\hspace{-3 pt}\mat{R}_{k}  - \mat{I}_{3\times3}}_{F_u}^2}},
\end{equation}
where $\norm{\cdot}_{F_u}$ corresponds to a Frobenius norm that is applied to the upper-triangular part of the matrix. The reconstructed rotation matrices $\mat{R}_k$ are defined by integrating $\vect{c}_k$, which now signifies the rotational velocity, in an additional reconstruction constraint:
\begin{equation}
	\label{eq:integrator_rotation}
	\mat{R}_{k+1} = \exp\left([\vect{c}_k]_{\times} \Delta \xi \right)\mat{R}_{k},~~~~ \forall k \in [1,N].
\end{equation}
Finally, adding the constraint $\mat{R}^T_{1}  \mat{R}_{1} = \mat{I}_{3\times3}$  ensures that orthonormality is preserved throughout the trajectory.

\section{Invariant descriptor for screw trajectories}
\label{sec:invars_screw}

This section derives a local invariant descriptor for a screw trajectory $\screw{s}(\xi)$, parameterized with respect to a progress variable $\xi$:
\begin{equation}
	\screw{s}(\xi) = \begin{pmatrix}\vect{a}(\xi) \\ \vect{b}(\xi)\end{pmatrix},~~~\text{with}~~~\vect{a},\vect{b}\in\mathbb{R}^{3\times1},
\end{equation}
where $\vect{a}$ is the direction part of the screw and $\vect{b}$ the moment part. Again we omit the explicit dependency on $\xi$ for conciseness while the derivative of $\screw{s}$ with respect to the progress
$\xi$ is denoted as: $\screw{s}' = \frac{\textup{d}\screw{s}}{\textup{d}\xi}$.

\subsection{Definition of invariants}

The screw invariants are based on a generalization of \textit{screw theory} for rigid-body kinematics and statics. In 1900, Ball \cite{ball1900treatise} published a complete \textit{theory of screws} based on the theorems of Chasles and Poinsot. \textit{Chasles' theorem} states that any rigid-body displacement can always be represented as a rotation and translation along an axis in space. \textit{Poinsot's theorem} is the dual to Chasles' theorem and it states that all forces and moments acting upon a rigid body can always be replaced by a single force and moment along an axis in space. This duality allows us to calculate the screw invariants of a wrench trajectory similar to how it was done for a twist trajectory in \cite{DeSchutter2010}. Hence, $\screw{s}$ can represent either a twist $\screw{t}$ or a wrench $\screw{w}$. In screw theory, the \textit{screw axis} refers to the axis along which the twist or wrench can be expressed \cite{featherstone2014rigid}, but commonly it is also referred to as the \textit{instantaneous screw axis} (ISA) to highlight its instantaneous nature.

The screw invariants are expressed in a local moving frame, referred to as the \textit{Instantaneous Screw Axis frame} (ISA frame), which is constructed as follows.
The orientation of the ISA frame is completely determined by $\vect{a}$, the direction part of the screw, i.e. $\vect{\omega}$ for a twist and $\vect{f}$ for a wrench. Hence, this orientation, represented by the rotation matrix $\matmf{R}$, is calculated in a similar way as the FS frame using \eqref{eq:tnb} after replacing $\vect{c}$ by $\vect{a}$. Now, the tangent $\vect{e}_t$ is along the ISA, while the binormal $\vect{e}_b$ is in the direction in which the ISA is instantaneously rotating.

\begin{figure}[t]
	\centering
	\subfloat[Object invariants]{\includegraphics[width=0.40\linewidth]{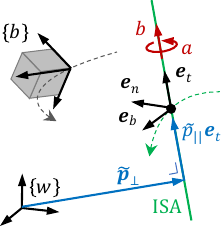}%
		\label{fig:screw_b}}
	\hfill
	\subfloat[Moving frame invariants]{\includegraphics[width=0.40\linewidth]{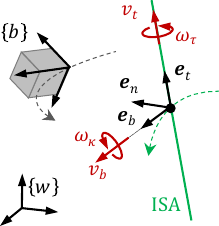}%
		\label{fig:screw_c}}
	\caption{Invariant descriptor for a screw trajectory: (a) object invariants $a$ and $b$ defined in the moving ISA frame given by the unit vectors $\vect{e}_t$, $\vect{e}_n$, $\vect{e}_b$, and the origin $\vectmf{p}=\vectmf{p}_\perp + \widetilde{p}_\parallel \vect{e}_t$; (b) the moving frame invariants $\omega_\kappa$, $\omega_\tau$, $v_b$, and $v_t$. The figure shows the special case where the screw $\screw{s}$ corresponds to the screw twist $\screw{t}$, representing the motion of the rigid body $\refframe{b}$ w.r.t $\refframe{w}$.}
	\label{fig:screw}
\end{figure}

On the other hand, the origin of the ISA frame $\vectmf{p}$ is defined as the point on the ISA about which the ISA is instantaneously rotating. An analytic procedure to derive this origin was proposed in \cite{DeSchutter2010}. We summarize it here as a two-step procedure, illustrate it schematically in Fig.~\ref{fig:screw_b}, and refer to \cite{DeSchutter2010} for more detail.
First, the point $\vectmf{p}_\perp$ on the ISA closest to the origin of the reference frame is determined:
\begin{equation}
	\vectmf{p}_\perp = \frac{\vect{a} \times \vect{b}}{\norm{\vect{a}}^2}.
\end{equation}
Second, the parallel displacement $\widetilde{p}_\parallel$ along the ISA (from $\vectmf{p}_\perp$ to $\vectmf{p}$ ) is defined \cite{DeSchutter2010}:
\begin{equation}
	\widetilde{p}_\parallel = \frac{\Vert \vect{a} \Vert^2}{\Vert\vect{a} \times \vect{a}'\Vert}\left(\vectmf{p}_\perp^{~\prime}\cdot\vect{e}_n\right),
\end{equation}
such that the \textit{origin} of the ISA frame is found by their addition:
\begin{equation}
	\vectmf{p} = \vectmf{p}_\perp + \widetilde{p}_\parallel \vect{e}_t.
\end{equation}
As opposed to the FS frame, the ISA frame is a \textit{complete} moving frame having both a position and orientation, which can be expressed using the pose matrix $\matmf{T}$:
\begin{equation}
	\matmf{T} = \begin{bmatrix}
		\matmf{R} & \vectmf{p} \\ \vect{0}_{3\times1}  &  1
	\end{bmatrix}.
	\label{eq:isa_frame}
\end{equation}
With the ISA frame determined, the screw invariants can be defined. The first two invariants $a$ and $b$ are defined, respectively, as the components of $\vect{a}$ and $\vect{b}$ along the ISA:
\begin{align}
	\label{eq:screw_invars_obj}
	a & = \vect{a} \cdot \vect{e}_t, &
	b & = \vect{b} \cdot \vect{e}_t.
\end{align}
These two invariants are referred to as \textit{object invariants} since they relate to the object: for motion, they correspond to the rotation and translation of the object along the ISA, while for force, they correspond to the force and moment applied to the object along the ISA.

The screw trajectory is reconstructed from the object invariants using a screw transformation matrix $\mat{S}(\mat{T})$ as in \eqref{eq:screwtransformation}:
\begin{equation}
	\screw{s} =  \mat{S}(\matmf{T})~ \screwmf{s}, ~~~\text{with}~~~\screwmf{s} =  \left(
	a ~ 0 ~ 0 ~ b ~ 0 ~ 0
	\right)^{\text{T}}.
	\label{eq:isa_obj}
\end{equation}

The first-order kinematics $\matmf{T}'$ of the ISA frame $\matmf{T}$ is completely determined by four invariants:
\begin{align}
	\label{eq:kin_T}
	\matmf{T}' & = \matmf{T}  \begin{bmatrix}
		                          0             & -\omega_\kappa & 0            & v_t \\
		                          \omega_\kappa & 0              & -\omega_\tau & 0   \\
		                          0             & \omega_\tau    & 0            & v_b \\
		                          0             & 0              & 0            & 0
	                          \end{bmatrix}.
\end{align}
Here, $\omega_\kappa$ and $\omega_\tau$ correspond to the curvature rate and torsion rate of the vector $\vect{a}$, which is the direction part of the screw $\screw{s}$. Hence, the invariants $\omega_\kappa$ and $\omega_\tau$ are calculated using similar formulas as in \eqref{eq:vec_invars_mf}, with $\vect{c}$ replaced by $\vect{a}$. This is evident because, for all $\xi$, the orientation part $\matmf{R}$ in \eqref{eq:isa_frame} is calculated in exactly the same way as $\matmf{R}$ in \eqref{eq:tnb} for the FS frame. Hence, also its derivative is the same.
The invariants $v_b$ and $v_t$ correspond to the translation of the origin of the ISA frame along $\vect{e}_b$ and $\vect{e}_t$ and are defined as \cite{DeSchutter2010}:
\begin{equation}
	v_b = \vectmf{p}_\perp^{~\prime} \cdot \vect{e}_b, ~\text{and}~
	v_t = \vectmf{p}_\perp^{~\prime}\cdot \vect{e}_t + \widetilde{p}_\parallel^{~\prime}.
\end{equation}
Equation~\eqref{eq:kin_T} can be written compactly using \eqref{eq:skew2}:
\begin{align}
	\matmf{T}' =  \matmf{T} \left[\screw{i}\right]_{\times}, ~~~\text{with}~~~ \screw{i} = \left(\omega_\tau ~ 0 ~ \omega_\kappa ~ v_t ~ 0 ~ v_b \right)^\text{T},
	\label{eq:SAI_mf}
\end{align}
where $\screw{i}$ is referred to as the \textit{moving frame invariant} since it models the first-order kinematics of the ISA frame.
Fig. \ref{fig:screw} visualizes the definition of the moving frame $\matmf{T}$, the object invariants $a$ and $b$, and the moving frame invariant $\screw{i}$.

\textbf{Singularities}:
The singularities of the screw invariants bear a strong relationship with the singularities of the vector invariants.
The first type of singularity occurs when the direction vector is zero ($\vect{a} = \vect{0}$). The object invariant $a$ is zero while the moving frame invariants $\omega_\kappa$, $\omega_\tau$, $v_t$, and $v_b$ are undefined. In other words, the complete moving frame is arbitrary. In this case, the moment vector $\vect{b}$, i.e., $\vect{v}$ for twist and $\vect{m}$ for wrench, can still be used to construct the orientation of the moving frame using the vector invariants defined in Section~\ref{sec:invars_vec}. This singular case is also referred to as \textit{infinite pitch} or \textit{pure moment} for motion and force, respectively.

The second type of singularity occurs when the screw axis has an unchanging direction $(\vect{a} \times {\vect{a}}' = \vect{0}$) so that it remains parallel to itself. As a result, the invariant $\omega_\kappa$ is zero, while the other moving frame invariants $\omega_\tau$, $v_t$, and $v_b$ are undefined. The perpendicular distance $\vectmf{p}_\perp$ is still defined. In other words, the direction of the ISA and perpendicular distance to the ISA are well-defined in this case, but the parallel distance $\widetilde{p}_\parallel$ and normal and binormal axes are arbitrary. In this case, the derivative of the perpendicular distance $\vectmf{p}_\perp^{~\prime}$ can be used instead of $\vect{a}$ to construct the normal and binormal axes. Thus, it is possible to determine the complete orientation and perpendicular distance of the ISA frame, but not its position along the ISA since $\widetilde{p}_\parallel$ is undefined. %\vspace{3pt}

\textbf{Invariant properties}:
All the invariant properties of the vector invariants in Section~\ref{sec:invars_vec} also hold for the screw invariants. The key advantage of the screw invariants compared to vector invariants is their additional invariance to the choice of the location of the reference point on the object. In other words, screw invariants are invariant to the choices of \textit{both} the world frame $\refframe{w}$ and body frame $\refframe{b}$ in which the screw coordinates $\screw{s}$ may be expressed. This property is also referred to as \textit{bi-invariance} \cite{murray1994}.

\textbf{Trajectory reconstruction}:
Reconstructing the trajectory $\screw{s}(\xi)$ from the invariant descriptor $(\screwmf{s}(\xi),\screw{i}(\xi))$ requires numerical integration of the differential equation \eqref{eq:SAI_mf}. Discretizing the problem again, the moving frame $\matmf{T}_{k+1}$ at sample $k+1$ can be found from $\matmf{T}_{k}$ at sample $k$ as follows:
\begin{align}
	\matmf{T}_{k+1} & =  \matmf{T}_k \exp \left( \left[ \screw{i}_k \right]_{\times} \Delta \xi \right), %\\
	\label{eq:SAI_rec}
\end{align}
where we assume that $\screw{i}_k$ remains constant over the integration step $\Delta \xi$. The matrix exponential operator $\exp(\cdot)$ maps the matrix $\left[\screw{i}\right]_{\times}\Delta\xi$ which is part of the Lie algebra $se(3)$ into the corresponding change in pose which is part of the Lie group $SE(3)$, and has a closed-form expression under the given assumptions \cite{murray1994}.
Given the initial value for $\matmf{T}_1$ at $k=1$, the moving frame $\matmf{T}_k$ can be reconstructed for all remaining samples $k=2...N$. From the moving frames, the corresponding screw trajectory $\screw{s}_k$ is retrieved using \eqref{eq:isa_obj}:
\begin{equation}
	\screw{s}_k =  \mat{S}(\matmf{T}_k) ~ \screwmf{s}_k, ~~\forall k \in [1,N].
	\label{eq:pose_dyn_disc2}
\end{equation}

Table \ref{tab:invars_vec_and_screw} (bottom half) provides an overview of the screw invariants framework with applications to rigid-body twist $\screw{t}$ and wrench $\screw{w}$ trajectories.

\begin{table*}[ht!]
	\centering
	\caption{Overview of equations for vector invariants (top half) and screw invariants (bottom half). The first column shows the type of vector and screw trajectories from which invariants can be calculated. In the second column, the local moving frame is defined in which the invariants are expressed. The third column defines the moving frame invariant that characterizes the motion of the moving frame, and the object invariant that characterizes the object's trajectory in the moving frame. The fourth column details how to reconstruct the trajectory from the invariant descriptor, possibly including the object's position and rotation.}
	\label{tab:invars_vec_and_screw}
	\resizebox{\linewidth}{!}{
	\renewcommand{\arraystretch}{1.1}
	\begin{tabular}{llll}
		\toprule
		\textbf{Trajectory coordinates} & \textbf{Moving frame definition} & \textbf{Invariants definition} & \textbf{Reconstruction}                        \\
		\toprule
		\begin{tabular}{l}
			\textbf{Vectors}	$\vect{c}$:                   \\
			~~$\vect{c} = \vect{\omega}$ for rot. velocity \\
			~~$\vect{c} = \vect{v}$ for transl. velocity   \\
			~~$\vect{c} = \vect{f}$ for force              \\
			~~$\vect{c} = \vect{m}$ for moment             \\
														   \\
			derivatives $\vect{c}'$ and $\vect{c}''$
		\end{tabular}
		                                &
		\begin{tabular}{l}
			\textbf{orientation frame} \\ $\matmf{R} = \left[\vect{e}_t ~~ \vect{e}_n ~~ \vect{e}_b \right]$, \\
			~~
			$\begin{aligned}
					  & \vect{e}_t =\frac{\vect{c}}{\norm{\vect{c}}}                                                                        \\
					  & \vect{e}_n = \frac{(\vect{c} \times \vect{c}') \times \vect{c}}{\norm{(\vect{c} \times \vect{c}') \times \vect{c}}} \\
					  & \vect{e}_b = \vect{e}_t \times \vect{e}_n
				 \end{aligned}$
			\vspace{3em}
		\end{tabular}
		                                &
		\begin{tabular}{l}
			\textbf{moving frame invariant}                                                                                                                                                                                       \\
			$\vect{i} = \left(\omega_\tau ~~ 0 ~~ \omega_\kappa \right)^T$: \vspace{0.5em}                                                                                                                                        \\
			$\begin{aligned}
					 ~~~~ & \omega_\kappa = \frac{||\vect{c} \times \vect{c}'||}{{||\vect{c}||}^2}                                                                                \\
					      & \omega_\tau = \frac{((\vect{c} \times \vect{c}') \times (\vect{c} \times \vect{c}'')) \cdot \vect{c}}{{||\vect{c} \times \vect{c}'||}^2 ||\vect{c}||}
				 \end{aligned}$ \\
			\textbf{object invariant}                                                                                                                                                                                             \\
			$\vectmf{c} = \left( c ~~ 0 ~~ 0 \right)^T$                                                                                                                                                                           \\
			~~~~$c = \vect{c} \cdot \vect{e}_t$
		\end{tabular}
		                                &
		\begin{tabular}{l}
			\textbf{moving frame equation}                                             \\
			$\matmf{R}' = \matmf{R} \left[ \vect{i}\right]_\times$                     \\
																					   \\
			\textbf{object trajectory equation}                                                   \\
			$\vect{c} = \matmf{R}~ \vectmf{c}$ \vspace{0.5em}                          \\
			object orientation $\mat{R}$ equation:                                    \\
			$\mat{R}' = [\vect{c}]_{\times} \mat{R}$, when $\vect{c}=\vect{\omega}$ \\
			object position $\vect{p}$ equation:                                       \\
			$\vect{p}' = \vect{c}$, when $\vect{c}=\vect{v}$
		\end{tabular}   \\

		\midrule

		\begin{tabular}{l}
			\textbf{Screws}	$\screw{s} =
			\begin{pmatrix}
					\vect{a} \\
					\vect{b}
				\end{pmatrix}$:                              \\
			~~$\screw{s} = \screw{t} =				\begin{pmatrix}
					                    \vect{\omega} \\
					                    \vect{v}
				                    \end{pmatrix}$ \\ 	~~~~~~ for twist \\
			~~$\screw{s} = \screw{w} = 	\begin{pmatrix}
					                   \vect{f} \\
					                   \vect{m}
				                   \end{pmatrix}$ \\ 	~~~~~~ for wrench \\
			                   		\\
			derivatives $\screw{s}'$ and $\screw{s}''$
		\end{tabular}
		                                &
		\begin{tabular}{l}
			\textbf{pose frame}                                                                                                      \\
			$\matmf{T} = \begin{bmatrix} \vect{e}_t ~~ \vect{e}_n ~~ \vect{e}_b ~~ \vectmf{p} \\ 0 ~~~ 0 ~~~ 0 ~~~ 1 \end{bmatrix}$: \\
			~~~~
			$\begin{aligned}
					  & \vect{e}_t = \frac{\vect{a}}{\norm{\vect{a}}}                                                                                                          \\
					  & \vect{e}_n = \frac{(\vect{a} \times \vect{a}') \times \vect{a}}{\norm{ (\vect{a} \times \vect{a}') \times \vect{a}}}                                   \\
					  & \vect{e}_b = \vect{e}_t \times \vect{e}_n                                                                                                              \\
					  & \vectmf{p} = \vectmf{p}_\perp + \widetilde{p}_\parallel \vect{e}_t                                                                                     \\
					  & ~~~~\vectmf{p}_\perp = \frac{\vect{a} \times \vect{b}}{\norm{\vect{a}}^2}                                                                              \\
					  & ~~~~\widetilde{p}_\parallel = \frac{\Vert \vect{a} \Vert^2}{\Vert\vect{a} \times \vect{a}'\Vert}\left(\vectmf{p}_\perp^{~\prime}\cdot\vect{e}_n\right) \\
				 \end{aligned}$
		\end{tabular}
		                                &
		\begin{tabular}{l}
			\textbf{moving frame invariant}                                                                                                                                                                                     \\
			$\screw{i} = \left(\omega_\tau ~ 0 ~ \omega_\kappa ~ v_t ~0 ~ v_b\right)^T$: \vspace{0.5em}                                                                                                                         \\
			$\begin{aligned}
					 ~~ & \omega_\kappa = \frac{||\vect{a} \times \vect{a}'||}{{||\vect{a}||}^2}                                                                                \\
					    & \omega_\tau = \frac{((\vect{a} \times \vect{a}') \times (\vect{a} \times \vect{a}'')) \cdot \vect{a}}{{||\vect{a} \times \vect{a}'||}^2 ||\vect{a}||} \\
					    & v_b = \vectmf{p}_\perp^{~\prime} \cdot \vect{e}_b                                                                                                     \\
					    & v_t = \vectmf{p}_\perp^{~\prime}\cdot \vect{e}_t + \widetilde{p}_\parallel^{~\prime}
				 \end{aligned}$	\vspace{0.5em} \\
			\textbf{object invariant}                                                                                                                                                                                           \\
			$\screwmf{s} = \left( a ~ 0 ~ 0 ~ b ~ 0 ~ 0\right)^T$: \vspace{0.5em}                                                                                                                                               \\
			$\begin{aligned}
					 ~~ & a = \vect{a}\cdot\vect{e}_t \\
					    & b = \vect{b}\cdot\vect{e}_t \\
				 \end{aligned}$
		\end{tabular}
		                                &
		\begin{tabular}{l}
			\textbf{moving frame equation}                               \\
			$\matmf{T}' = \matmf{T} \left[ \screw{i} \right]_\times$     \\
																		 \\
			\textbf{object trajectory equation}                                     \\
			$\screw{s} = \mat{S}(\matmf{T})~ \screwmf{s}$ \vspace{2.5em} \\
			object pose $\mat{T}$ equation:                             \\
			$\mat{T}' = \left[{}_{}\screw{s} \right]_{\times} \T{}{}$,  when $\screw{s}=\screw{t}$ \vspace{4em}
		\end{tabular} \\
		\bottomrule
	\end{tabular}
	}
\end{table*}

\subsection{Robust calculation of invariants using optimal control}
\label{sec:robust_calc_screw}

Calculating screw invariants $\screwmf{s}$ and $\screw{i}$ with the analytical formulas has similar problems related to singularities, sensitivity to noise, and integration drift during reconstruction, as mentioned in Section~\ref{sec:robust_calc_vector} for vector invariants.

To address these problems, we again formulate an OCP, of which the objective is to find the screw invariants that reconstruct the given measured screw trajectory in a robust way. This is reflected by introducing a cost $\Delta\screw{s}_k$ that contains the difference between the reconstructed $\screw{s}_k$ and measured screw trajectory $\screw{s}^{meas}_k$ at sample $k$:
\begin{equation}
	\Delta \screw{s}_k = \screw{s}_k -  \screw{s}^{meas}_k =  \begin{pmatrix}
		\Delta \vect{a}_k \\
		\Delta \vect{b}_k
	\end{pmatrix}.
	\label{eq:obj_meas_screw}
\end{equation}

Similarly as in Section~\ref{sec:robust_calc_vector}, a mean-squared trajectory reconstruction error is formulated, but the errors on the directional part of the screw $\Delta \vect{a}_{\text{MSE}}$ and the moment part of the screw $\Delta \vect{b}_{\text{MSE}}$ are considered separately:
\begin{align}
	\epsilon_a^2 \geq \Delta \vect{a}_{\text{MSE}}(\screw{s},\screw{s}^{meas}) & = \frac{1}{N} \sum\limits_{k=1}^{N}{\norm{\Delta \vect{a}_k}^2}, \label{eq:traja} \\
	\epsilon_b^2 \geq \Delta \vect{b}_{\text{MSE}}(\screw{s},\screw{s}^{meas}) & = \frac{1}{N} \sum\limits_{k=1}^{N}{\norm{\Delta \vect{b}_k}^2}, \label{eq:trajb}
\end{align}
bounded by tolerance $\epsilon_a$ for the direction part of the screw and tolerance $\epsilon_b$ for the moment part of the screw.

A regularization cost is introduced for the moving frame invariants to achieve a more stable and smooth evolution of the moving frame $\matmf{T}$ in the presence of measurement noise and singularities:
\begin{equation}
	\scalemath{0.94}{\sum\limits_{k=1}^{N-1}{ {\norm{\screw{i}_k}}^2_{L}}} = \scalemath{0.94}{\sum\limits_{k=1}^{N-1}{ \omega^2_{\kappa}[k] +  \omega^2_{\tau}[k] +  \frac{1}{L^2} \left( v^2_{n}[k] + v^2_{b}[k] \right) }},
\end{equation}
where $L$ signifies a chosen length scale with units [m] to properly compare rotational and translational invariants.

The complete OCP to calculate the screw invariants $\screwmf{s}$ and $\screw{i}$ from a given sequence of $N$ measured screws $\screw{s}^{meas}_k$ is:
\begin{align}
	\begin{aligned}
		\underset{				          \screwmf{s}_{[\cdot]}, \screw{i}_{[\cdot]},
		\screw{s}_{[\cdot]}, \matmf{T}_{[\cdot]}
		}{\text{minimize}} ~~
		 &
		\sum\limits_{k=1}^{N-1}{ {\norm{\screw{i}_k}}^2_{L}
		},
		\label{eq:obj_screw}
	\end{aligned}
\end{align}
\vspace{-0.4cm}
\nopagebreak
\text{~subject to:}
\nopagebreak
\vspace{0.3cm}
\begin{alignat}{2}
	\epsilon^2_a                  & \geq \Delta \vect{a}_{\text{MSE}}(\screw{s},\screw{s}^{meas}),  \tag{r.\ref{eq:traja}}                                                             \\
	\epsilon^2_b                  & \geq \Delta \vect{b}_{\text{MSE}}(\screw{s},\screw{s}^{meas}), \tag{r.\ref{eq:trajb}}                                                              \\
	\matmf{T}_{k+1}               & =  \matmf{T}_k \exp\left( \left[ \screw{i}_k \right]_{\times}  \Delta \xi \right),     &  & ~~~\forall k \in [1,N] \tag{r.\ref{eq:SAI_rec}}        \\
	\screw{s}_k                   & = \mat{S}(\matmf{T}_k) ~ \screwmf{s}_k,                                                &  & ~~~\forall k \in [1,N] \tag{r.\ref{eq:pose_dyn_disc2}} \\
	\matmf{T}_1^{-1}\matmf{T}_{1} & = \mat{I}_{4\times4}. \label{eq:orthog_screw}
\end{alignat}
Besides the trajectory error constraints \eqref{eq:traja}-\eqref{eq:trajb}, the constraints consist of the trajectory reconstruction equations \eqref{eq:SAI_rec}-\eqref{eq:pose_dyn_disc2}, supplemented with an orthonormality constraint on the initial moving frame $\matmf{T}_{1}$.

\textbf{Initialization}:
Similarly as for vector invariants, an invariant initialization of the variables in the OCP can be done using either the analytical formulas or an approach based on an average moving frame \cite{asapaper}. The latter is summarized as follows. The orientation of the moving frames $\matmf{T}_k$ is initialized using the same approach as for vector trajectories, based on the directional part of the measured screw trajectory $\screw{s}_k^{meas}$. The origin of the moving frames $\matmf{T}_k$ is initialized using the average intersection point of all the instantaneous screw axes corresponding to the measured screw trajectory $\screw{s}^{meas}$ \cite{asapaper}.
The moving frame invariants $\omega_\kappa$, $\omega_\tau$, $v_b$, and $v_t$ are initialized with zero values. The object invariants $a_k$ and $b_k$ are initialized by transforming the screw trajectory $\screw{s}_k^{meas}$ to the average moving frame and selecting the components along the first axis for each sample $k$.

\textbf{Application to motion and force data}:
The calculation of screw invariants using OCP \eqref{eq:obj_screw}-\eqref{eq:orthog_screw} can directly be applied to twist trajectories $\screw{t}(\xi)$ and wrench trajectories $\screw{w}(\xi)$.

Similarly as in Section~\ref{sec:robust_calc_vector}, if the measured motion data consists of positions and orientations of the rigid body, instead of twists, the optimization problem \eqref{eq:obj_screw}-\eqref{eq:orthog_screw} can be adapted to include the object's pose. Below, this is worked out assuming the positions and orientations are given by homogeneous transformation matrices. The proposed approach can be adapted to other pose representations, such as dual quaternions.

\textbf{Including pose measurements for case $\screw{s}=\screw{t}$}:
Additional variables are introduced for the reconstructed pose matrices $\mat{T}_k = \begin{bmatrix} \mat{R}_k & \vect{p}_k \\ \vect{0}_{1\times3} & 1 \end{bmatrix}$ at each sample $k$. The trajectory reconstruction costs $\Delta \vect{a}_{\text{MSE}}$ and $\Delta \vect{b}_{\text{MSE}}$ in \eqref{eq:traja}-\eqref{eq:trajb} are replaced by the MSE errors between the reconstructed and the measured rotation matrices and positions:
\begin{align}
	\label{eq:cost_pose}
	\scalemath{0.91}{\Delta \mat{R}_{\text{MSE}}(\mat{R},\mat{R}^{meas})}    & = \scalemath{0.91}{\frac{1}{N} \sum\limits_{k=1}^{N}{\norm{\left(\mat{R}^{meas}_{k}\right)^T\hspace{-3 pt}\mat{R}_{k}  - \mat{I}_{3\times3}}_{F_u}^2}}, \\
	\Delta \vect{p}_{\text{MSE}}(\vect{p},\vect{p}^{meas}) & = \frac{1}{N} \sum\limits_{k=1}^{N}{\norm{\vect{p}_k -  \vect{p}^{meas}_k}^2}.
\end{align}
The reconstructed pose matrices $\mat{T}_k$ are defined by integrating $\screw{s}_k$, in this case signifying the object's twist, in additional reconstruction constraints:
\begin{equation}
	\label{eq:integrator_pose}
	\T{}{}_{k+1} = \exp\left(  \left[{}_{}\screw{s}_k \right]_{\times} \Delta \xi \right) \T{}{}_k, ~~~ \forall k \in [0,N].
\end{equation}
Finally, adding an orthonormality constraint on the initial pose $\mat{T}_1$, like in \eqref{eq:orthog_screw}, ensures that orthonormality is preserved.

\section{Application to 3D contour following}
\label{sec:experiments}

\begin{figure}[t]
	\centering
	\subfloat[Setup]{\includegraphics[width=0.44\linewidth]{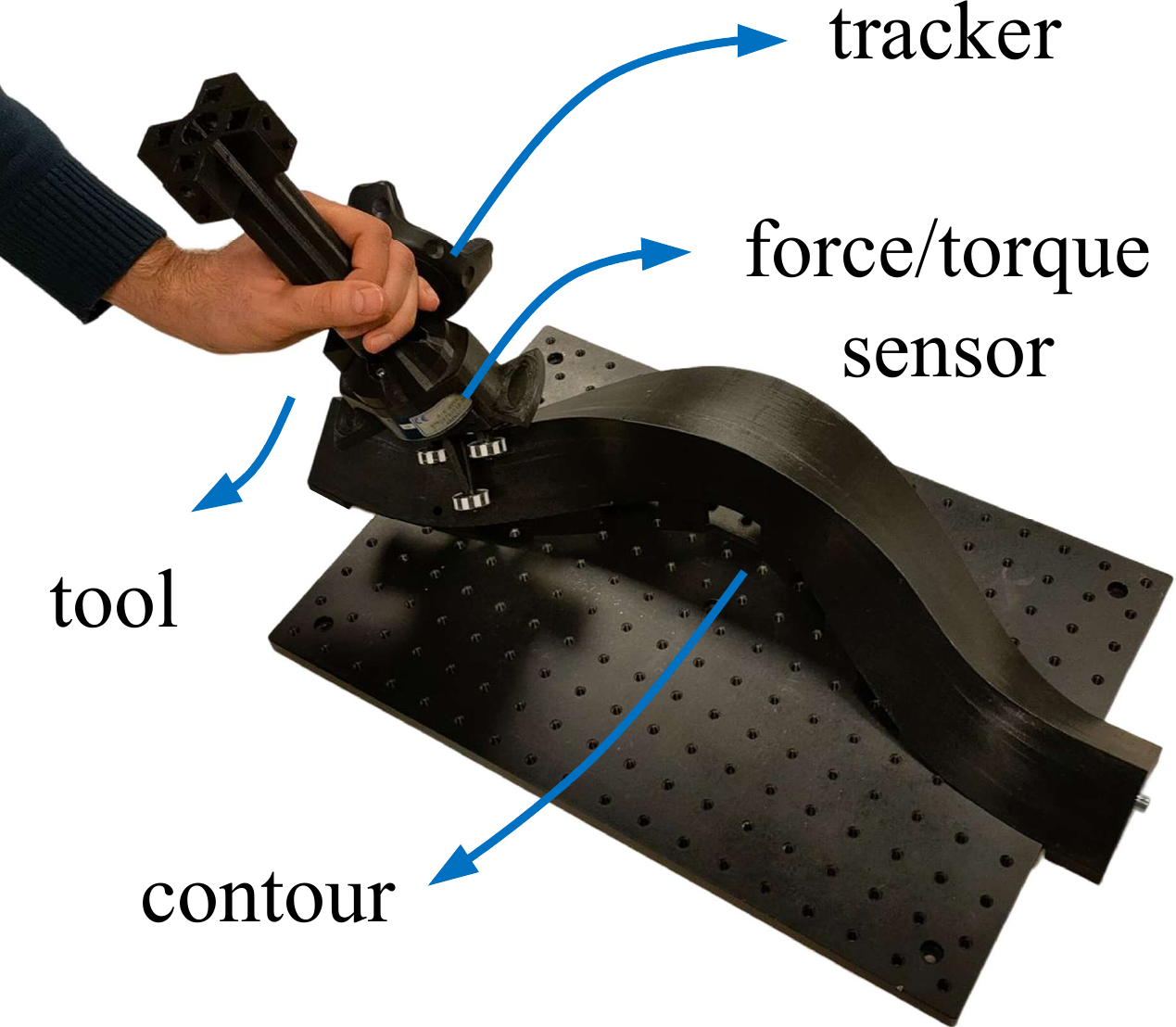}%
		\label{fig:setup_pic}}
	\hfill
	\subfloat[Frames]{\includegraphics[width=0.24\linewidth]{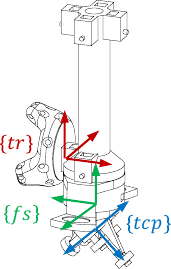}%
		\label{fig:setup_frames}}
	\hfill
	\subfloat[Cross-section]{\includegraphics[width=0.22\linewidth]{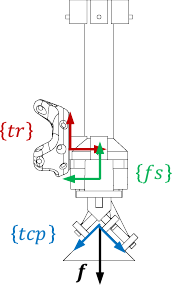}%
		\label{fig:setup_frames2}}
	\caption{The 3D contour following setup: (a) demonstration setup showing tool and contour, (b) 3D drawing showing the three assigned frames on the tool: tracker $\{tr\}$, force/torque sensor $\{fs\}$, and tool center point (TCP) $\{tcp\}$, (c) cross-section of the contour showing the reference contact force $\protect{\vect{f}}$.}
	\label{fig:setup}
\end{figure}

This section demonstrates the use and benefits of screw and vector invariant descriptors for both motion and force trajectories by analyzing a human-demonstrated 3D contour following task, shown in Fig. \ref{fig:setup}.
The human operator holds the tool while following the contour and pressing the tool's five contact wheels (diameter $22 ~\text{mm}$) against the contour.

The contour consists of the edge between two contact surfaces that form an angle of $90^\circ$ in each point along the contour (see Fig. \ref{fig:setup_frames2}). Each of the five contact wheels takes away one degree of freedom for the motion of the tool relative to the contour, while creating one degree of freedom to apply a contact force. Hence, every point along the contour is characterized by a 1-dof vector space of possible twists and a 5-dof  vector space of possible contact wrenches. This makes the application interesting, as we expect to obtain a very good repeatability of the invariant descriptors for motion among different demonstrations, while we expect to see the effect of human variation among the demonstrations in the invariant descriptors for force. Furthermore, the contour is designed to have a symmetric curvature profile and an antisymmetric torsion profile, both with respect to its midpoint. The contour starts and ends with a straight line segment (no curvature or torsion).

\subsection{Experimental set-up}
\label{sec:exp-setup}

An HTC VIVE motion capture system, consisting of a tracker mounted on the tool and a camera (not shown in Fig.~\ref{fig:setup}), records the pose trajectory of the tool with respect to a reference frame attached to the camera. This frame is chosen as the world frame $\{w\}$. The system records the pose trajectory with a frequency of $200~\mathrm{Hz}$ and an accuracy in the order of a few millimeters and a few degrees. The contact force/torque is measured with the same frequency using a 6-axis JR3 force/torque sensor. The nominal accuracy of this force/torque sensor is $\pm1\%$ of its standard measurement range which is $\pm800\ \text{N}$ along the central axis (Z-axis) and $\pm400\ \text{N}$ in the horizontal plane (XY-plane) for the force, and $\pm24\ \text{Nm}$ in all directions for the moment.

Three frames are rigidly attached to the tool: the motion tracker frame $\{tr\}$, the force/torque sensor frame $\{fs\}$, and the \textit{tool-center-point }(TCP) frame $\{tcp\}$. The $\{tcp\}$ frame is defined such that, when the tool is tracking a straight-line segment, the origin of $\{tcp\}$ corresponds to the intersection of the contour's edge and the symmetry axis of the cylindrical part of the tool (see Fig. \ref{fig:setup_frames2}). One axis of the $\{tcp\}$ frame is along the edge and the remaining two axes are along the two surfaces. (For completeness: if the edge has curvature and/or torsion, the $\{tcp\}$ frame will deviate slightly from this contour-centered definition due to the complex contact geometry).

To introduce variation in the measurements, the tracker can be physically attached to the tool at different locations, further discussed in Section \ref{sec:exp-design}. However, the location of the force sensor relative to the tool cannot be changed in the current set-up. Instead, and without loss of generality, we can artificially transform the physically measured wrenches to a different, virtual frame on the tool using a screw transformation matrix $\mat{S}$ as in \eqref{eq:screw_transformtation}. We can take this one step further and in a similar manner transform the physically measured wrenches to the world frame $\{w\}$. This way we can, again without loss of generality, simulate an alternative experimental set-up in which the force sensor is fixed to the world instead of the tool, i.e., representing a situation where the force sensor is placed underneath the support to which the contour is fixed.

\subsection{Objectives}
\label{sec:exp-obj}

A first objective is to show that the invariant descriptors have a physical interpretation and to obtain evidence that they are calculated correctly using the OCPs. It is however very difficult to calculate the ground truth invariants given the complex contact geometry involving five contact wheels with finite radii and finite mutual distances. Instead, we calculated reference values to compare the invariants with, for the limit case of an infinitely small set of contact wheels and a contact wrench which corresponds to a pure force with constant magnitude $\norm{\boldsymbol{f}}=25~\text{N}$, as shown in Figure \ref{fig:setup_frames2}. In this limit case, the pose of $\{tcp\}$ is easily derived from the contour's CAD data. The wrench is then also uniquely defined and constant in any frame attached to the tool. From the reference pose and reference wrench we calculate the corresponding invariant descriptors using the same OCPs as for the human-demonstrated data. 

A second objective is to confirm the invariant properties of the vector invariants and the screw invariants, as discussed in Sections \ref{sec:invars_vec} and \ref{sec:invars_screw}, respectively. Furthermore, we want to point out the practical benefits of these properties in terms of relaxed calibration needs.  We also want to show the difference between the invariant descriptors depending on the perspective of the input data: as the motion and wrench trajectories can be defined either with respect to the world or with respect to the moving body, the corresponding moving frame invariants will also model the motion of the moving frame with respect to the world or the moving body, and hence they will be different.

A third objective is to confirm that the new formulation of the OCPs for calculating the invariants, as introduced in Sections~\ref{sec:robust_calc_vector} and \ref{sec:robust_calc_screw}, allows for an intuitive and effective parameter tuning.

A fourth and final objective is to confirm that the trajectories reconstructed from the calculated invariants are accurate and driftless, even when the trajectories are reconstructed in a different region in space.

\subsection{Experiment design}
\label{sec:exp-design}

Twelve demonstrations of the 3D contour following task were recorded. According to the experimental protocol, the operator was asked to repeat the task twelve times by starting from approximately the same location (at the end of the initial straight-line segment) and approximately ending at the same location (at the beginning of the final straight-line segment). The operator was also asked to try to maintain a constant pure contact force as shown in Fig. \ref{fig:setup_frames2}, because this can be shown to be a very good strategy to ensure that the five wheels remain in contact with the contour. Both instructions contributed to limiting human variation in the task execution.

\begin{figure}[t]
	\centering
	\subfloat[]{\includegraphics[width=0.105\linewidth]{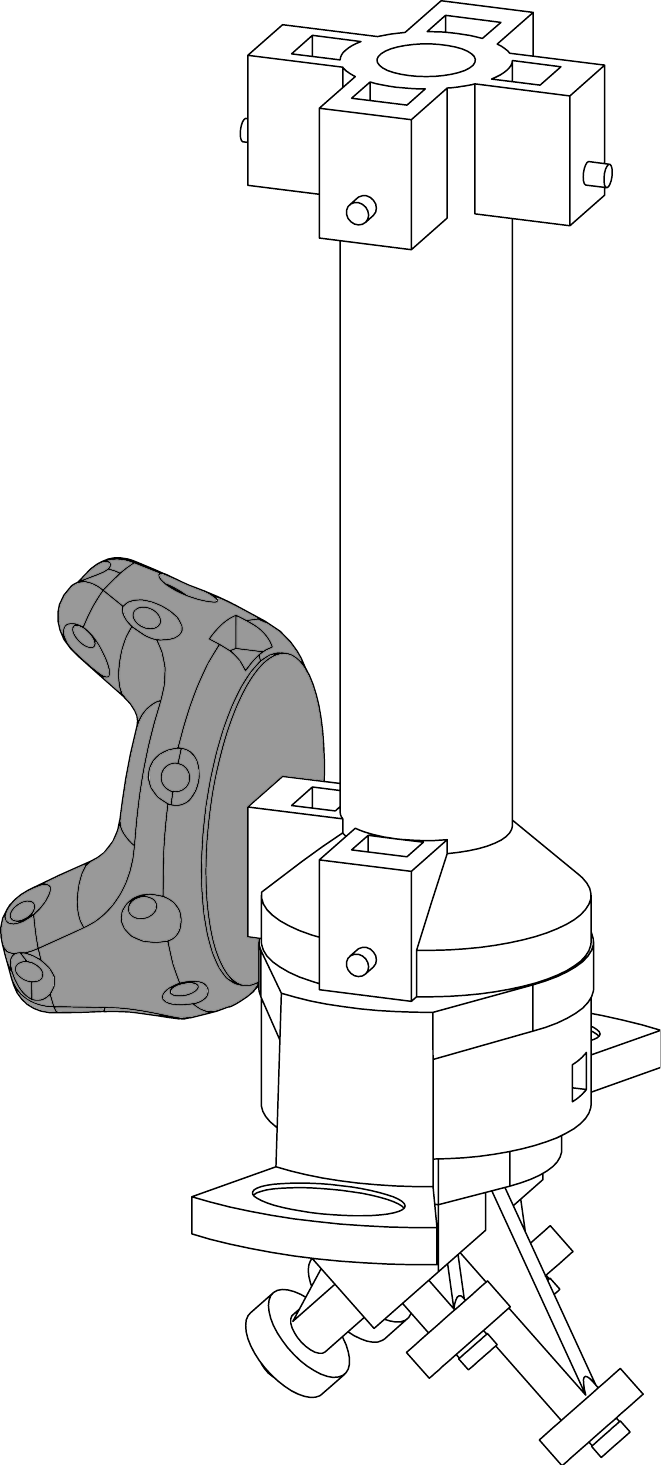}}%
	\hfil
	\subfloat[]{\includegraphics[width=0.105\linewidth]{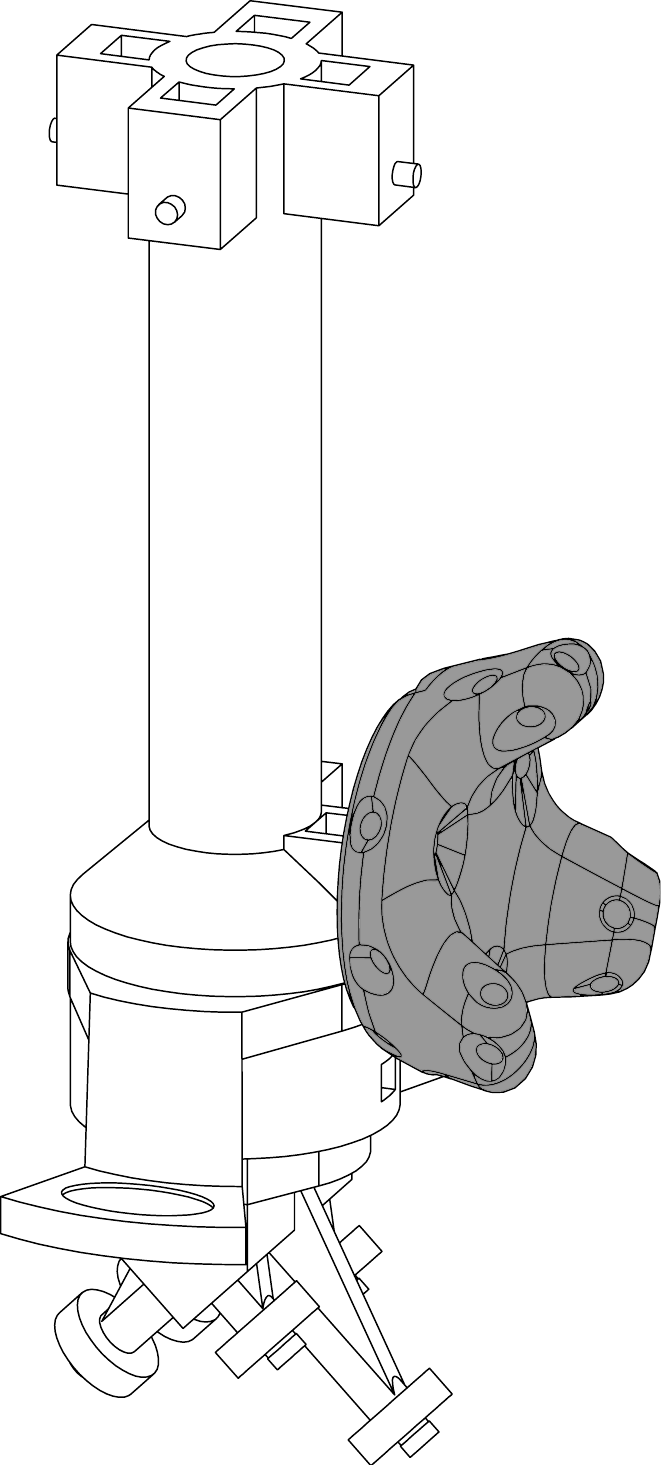}}%
	\hfil
	\subfloat[]{\includegraphics[width=0.100\linewidth]{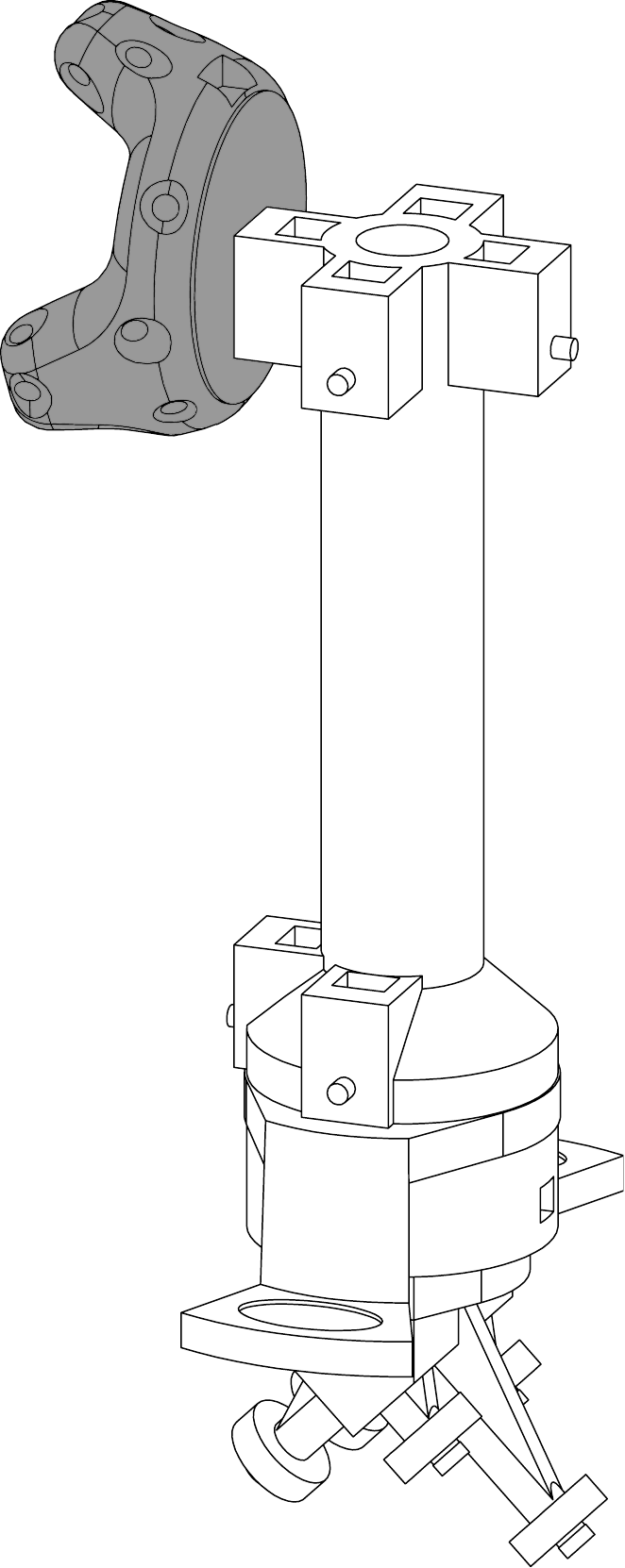}}%
	\hfil
	\subfloat[]{\includegraphics[width=0.085\linewidth]{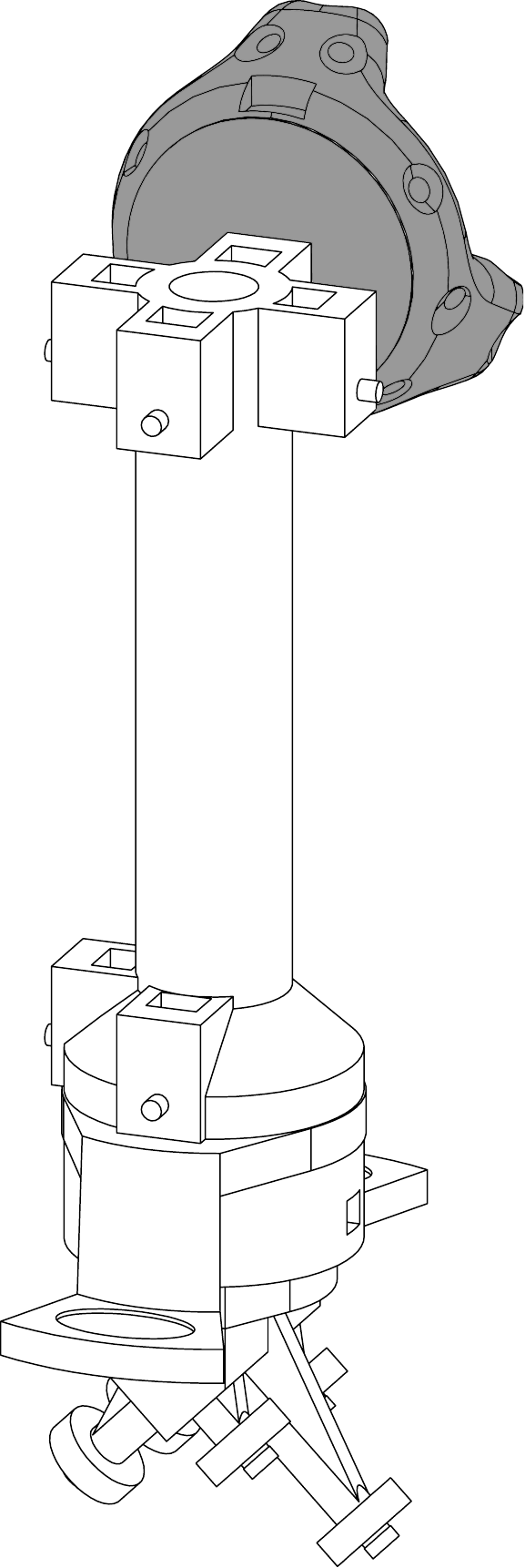}}%
	\hfil
	\subfloat[]{\includegraphics[width=0.100\linewidth]{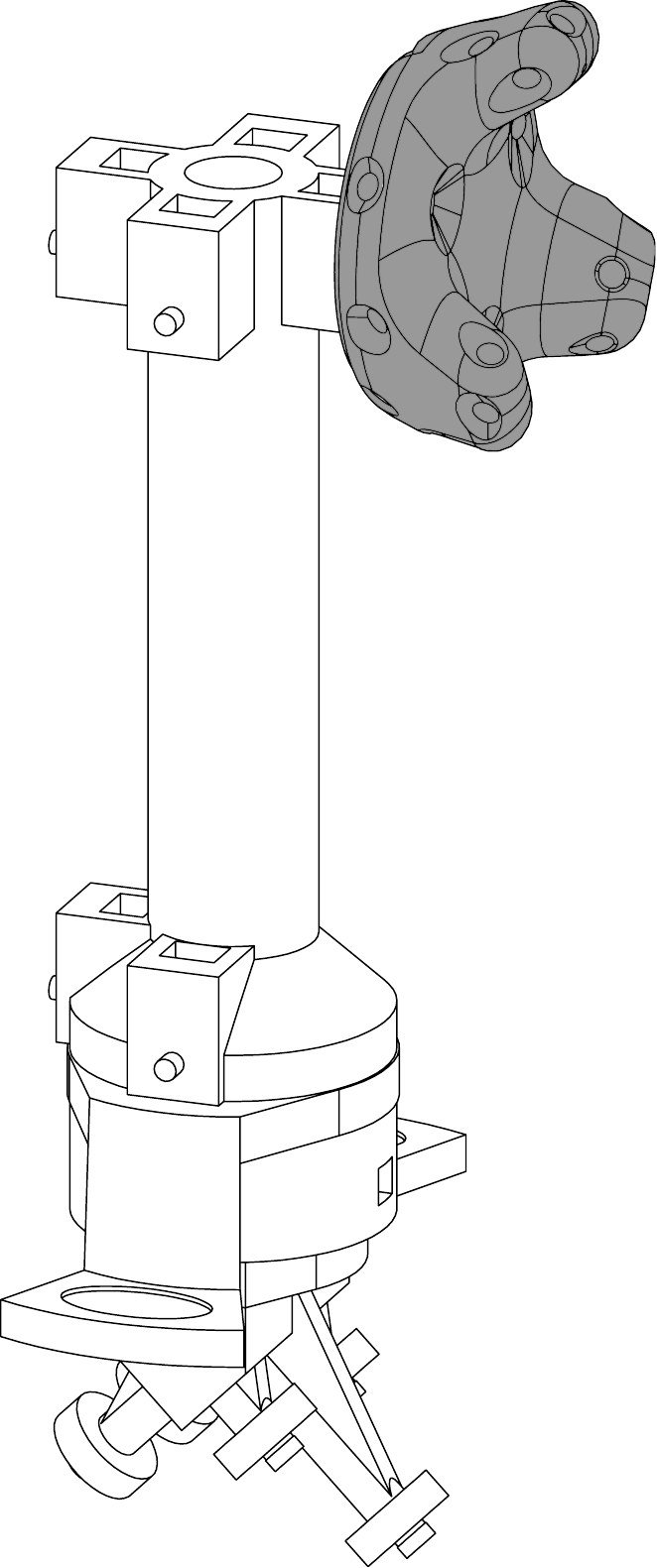}}%
	\hfil
	\subfloat[]{\includegraphics[width=0.090\linewidth]{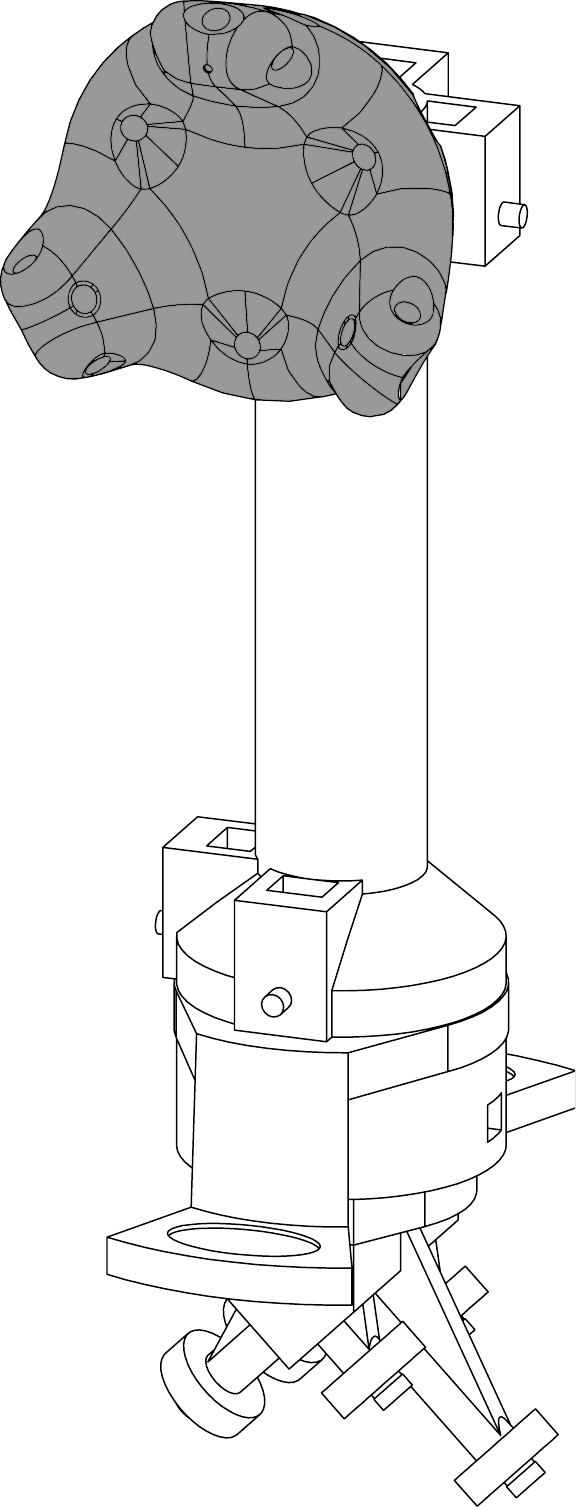}}%
	\caption{The six different configurations of the motion tracker on the tool.}
	\label{fig:configurations}
\end{figure}

Each of the twelve demonstrations was recorded with the tracker physically attached to the tool in one of six different locations, shown in Fig. \ref{fig:configurations} (i.e. two demonstrations per tracker location), resulting in significantly different pose trajectories.
Additionally, the physically measured contact wrenches were artificially transformed from the force sensor frame $\{fs\}$ to the tracker frame $\{tr\}$ used in the demonstration, simulating a change in location of the force sensor and resulting in significantly different wrench trajectories.

The screw invariants are expected to be invariant for the previous variations, but the vector invariants are not. To verify the repeatability of the vector invariants, it is necessary to pick a well-chosen reference point on the tool to express the position and moment trajectories, the same reference point for all demonstrations. The obvious choice was the origin of $\{tcp\}$, both for the position and moment trajectories. Hence, for this specific case, the measured pose and wrench trajectories of the twelve demonstrations were additionally transformed to the $\{tcp\}$ frame. Note that this requires calibration of the origins of $\{tr\}$ and $\{fs\}$ with respect to $\{tcp\}$, but this is required solely for the vector invariants, not the screw invariants.

The twelve resulting pose trajectories were further modified by changing the pose of the world reference frame relative to the contour, simulating a change in camera viewpoint. Table \ref{tab:table_demonstrations} lists the different artificial transformations that were applied to each of the twelve trials together with their corresponding tracker configuration. Changing the world reference frame is not expected to affect the screw or vector invariants for motion. Obviously, such change also does not affect the wrench trajectories and the resulting wrench invariants since the force sensor is attached to the tool, not to the world.

\begin{table}
	\caption{Each of the twelve demonstration trials is listed with one of the six configurations of the motion tracker (see Fig. \ref{fig:configurations}) and with the artificial transformation that was applied to the measured poses to simulate a large variation in the world (i.e. camera) frame. These transformations are specified using XYZ Euler angles and a translation vector.}
	\label{tab:table_demonstrations}
	\resizebox{\linewidth}{!}{
	\begin{tabular}{@{}lll|lll@{}}
		\toprule
		 \textbf{Trial \#} & \multicolumn{2}{c|}{\textbf{Artificial transformation}} & \textbf{Trial \#} & \multicolumn{2}{c}{\textbf{Artificial transformation}} \\ \textbf{+config.}  &   angles [°]   & transl. [m]   &   \textbf{+config.}   &  angles [°]    & transl. [m]  \\
		\midrule
		1~(a)                      & [0, -60, 0]                                   & [-3, 2, -1]                                   & 7  ~(d)                      & [-90, 0, 45]                                  & [1, 0.5, -0.5]                                \\
		2 (a)                      & [30, 0, -180]                                 & [-1, 2.5, -1]                                 & 8  ~(d)                      & [0, 90, -135]                                 & [-1, 0, -0.5]                                 \\
		3 (b)                      & [0, -60, -180]                                & [-1, 2, -2]                                   & 9  ~(e)                      & [60, 0, -150]                                 & [1, -2, 2]                                    \\
		4 (b)                      & [45, 0, 0]                                    & [-3, 2.5, -2]                                 & 10 (e)                     & [0, 45, -60]                                  & [3, -2, 2]                                    \\
		5 (c)                      & [-90, 45, 90]                                 & [-1, 0.5, 0.5]                                & 11 (f)                     & [0, -45, -180]                                & [3, -1.5, 1]                                  \\
		6 (c)                      & [0, 90, 45]                                   & [1, 0, 0.5]                                   & 12 (f)                     & [-60, 0, -45]                                 & [1, -1.5, 1]                                  \\
		\bottomrule
	\end{tabular}
	}
\end{table}

To simulate the alternative experimental set-up in which the force sensor is fixed to the world, the twelve measured wrench trajectories were artificially transformed to a single frame, fixed to the world underneath the contour, about halfway the contour. The exact location of this frame relative to the contour is not known or calibrated, but this does not affect the screw invariants.

Finally, to show that the trajectories reconstructed from the invariants are accurate and driftless, the motion and force trajectories were reconstructed at a new location and were compared with the measured trajectories, after they were globally transformed to the same location. 

\subsection{Data processing}
\label{sec:exp-processing}

This section explains how the screw and vector invariants are calculated from the measured pose and wrench trajectories and the reference pose and wrench trajectories.

\textbf{Preprocessing:} There are two preprocessing steps: segmentation and re-parameterization.

The purpose of segmentation is to extract the motion-in-contact part from a complete demonstration, which also includes an approach-to-contact motion, a retract motion, and two stationary parts while in contact, at the start and the end of the actual contour tracking motion. Segmentation is achieved by using a threshold for the magnitude of the force ($5 ~\text{N}$) to detect contact, and a threshold for the magnitude of the translational velocity of the TCP ($0.05 ~\text{m/s}$) to remove the stationary parts of the demonstration.

Re-parameterization involves changing the time-based input trajectories to trajectories that are function of an alternative progress variable. As discussed in Section \ref{subsec:progress}, for tasks with dominant translation like contour following, the integral of the translational velocity is a natural choice. Accordingly, we chose the progress variable $\xi$ as the integral of the TCP's velocity, because it best reflects the arc length along the contour. Finally, we rescaled the progress variable to a dimensionless variable ranging from $0$ to $1$, making it independent of the scale of the contour.

\textbf{Calculation of the invariants:}
The force invariants were calculated from the wrench measurements using the original OCPs: \eqref{eq:obj_vec}-\eqref{eq:orthog_vec} for vector invariants and \eqref{eq:obj_screw}-\eqref{eq:orthog_screw} for screw invariants.
The motion invariants were calculated from the pose measurements using the adapted versions of the OCPs using \eqref{eq:cost_position}-\eqref{eq:integrator_rotation} for the vector invariants and \eqref{eq:cost_pose}-\eqref{eq:integrator_pose} for the screw invariants. All OCPs were specified using the nonlinear optimization framework CasADi \cite{Andersson2019} and solved using a primal-dual Newton method with IPOPT \cite{ipopt}.

\textbf{Initialization of the OCP:}
For initializing the moving frames, we followed the initialization approaches based on an average moving frame as explained in Sections~\ref{sec:robust_calc_vector} and \ref{sec:robust_calc_screw} with a minor variation. Due to the symmetry and antisymmetry in the contour, averages calculated over the trajectory might become approximately zero.
To avoid confusion in determining the signs, we calculated the average over the last two-thirds of the trajectory.
The motion invariants are calculated using pose measurements as input. However, the initialization routine uses velocity and twist data as input. Hence, for motion, numerical differentiation is performed to estimate the velocities and twists as inputs for the initialization routine.

\textbf{Tuning of the OCPs:} Vector and screw invariants for motion and force were calculated using measured position, orientation, force, and moment trajectories as inputs. For every type of trajectory, a tolerance $\epsilon$ related to the desired accuracy of the reconstructed trajectory had to be chosen, such as for example in \eqref{eq:traj_recon_cons} for the vector trajectories. To maintain an intuitive relation between the values for these tolerances and the accuracy of the measurement systems, we expressed these tolerances in the respective sensor frames. All reported results were obtained using the following intuitive values: $\epsilon_p = 2 \text{ mm}$ for position and $\epsilon_R = 2^{\circ}$ for orientation, which corresponded to a rough estimation of the accuracy of the motion capture system, and $\epsilon_f = 0.8\text{ N}$ for force and $\epsilon_m = 0.16 \text{ Nm}$ for moment, which corresponded to the estimated noise levels of the force/torque measurements.  To calculate screw invariants, a length scale $L$ must be chosen to properly compare rotational and translational invariants. For the application we chose $L = 0.5$ m, i.e. the length of the contour's edge.

\textbf{Reconstructing trajectories from invariants:} \label{subsec:recon} The pose trajectory $\mat{T}(\xi)$ of the tool and the contact wrench trajectory $\screw{w}(\xi)$ were reconstructed from the screw invariant descriptors at a new location in space. This was done by supplying new initial moving frames $\matmf{T}(0)$ for both motion and wrench, and a new initial pose of the tool $\mat{T}(0)$ in the trajectory reconstruction equations \eqref{eq:SAI_rec}, \eqref{eq:pose_dyn_disc2} and \eqref{eq:integrator_pose}. The frames were specified by globally transforming the original calculated initial frames to the new location. Time-based trajectories $\mat{T}(t)$ and $\screw{w}(t)$ were obtained by re-applying the time profile $\xi(t)$ that was extracted from the demonstration using the inverse of the reparameterization in \eqref{eq:reparam}.

\subsection{Results and discussion}
\label{sec:exp-results}

\begin{figure}[t]
	\centering
	\subfloat[Screw invariants calculated from measured pose of tracker]{%
		\includegraphics[width=1.0\linewidth]{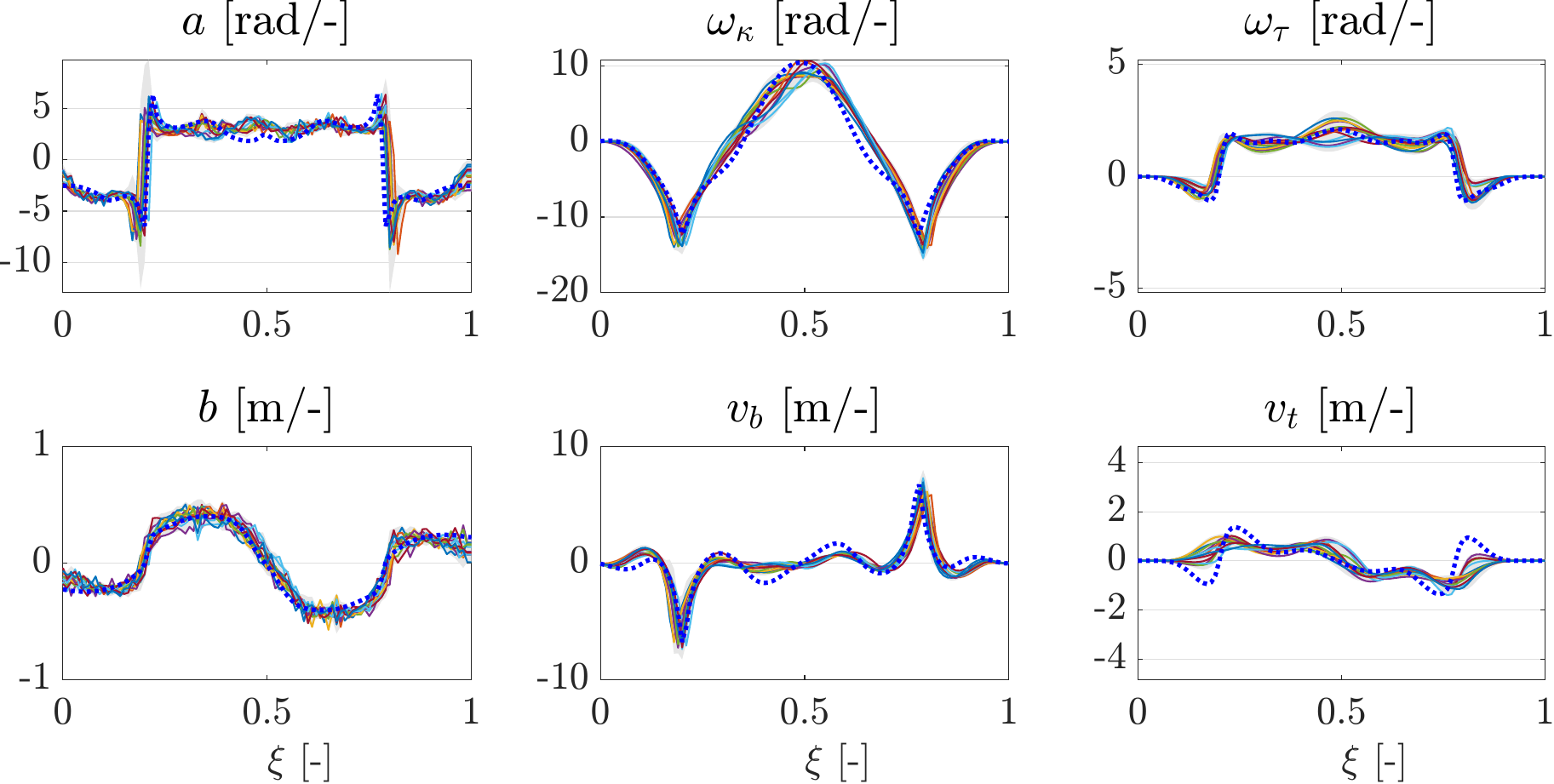}%
		\label{fig:motion_results_SAI}} 
	\\
	\subfloat[Vector invariants calculated from measured orientation of tracker]{%
		\includegraphics[width=1.0\linewidth]{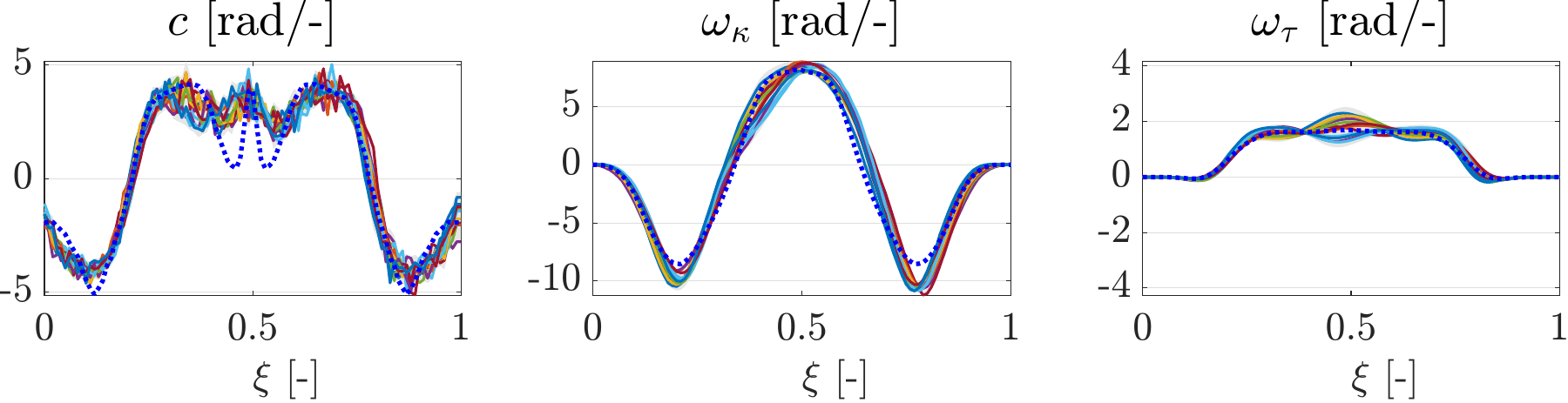}%
		\label{fig:motion_results_orientation}} 
	\\
	\subfloat[Vector invariants calculated from measured position of tracker]{%
		\includegraphics[width=1.0\linewidth]{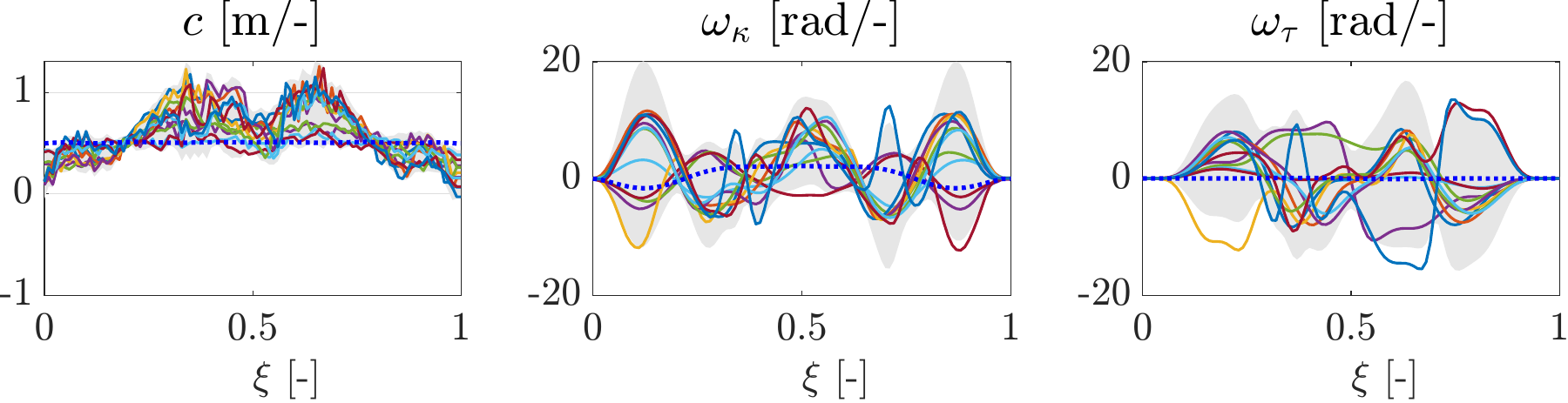}%
		\label{fig:motion_results_positiontracker}} 
	\\
	\subfloat[Vector invariants calculated from estimated position of tool point]{%
		\includegraphics[width=1.0\linewidth]{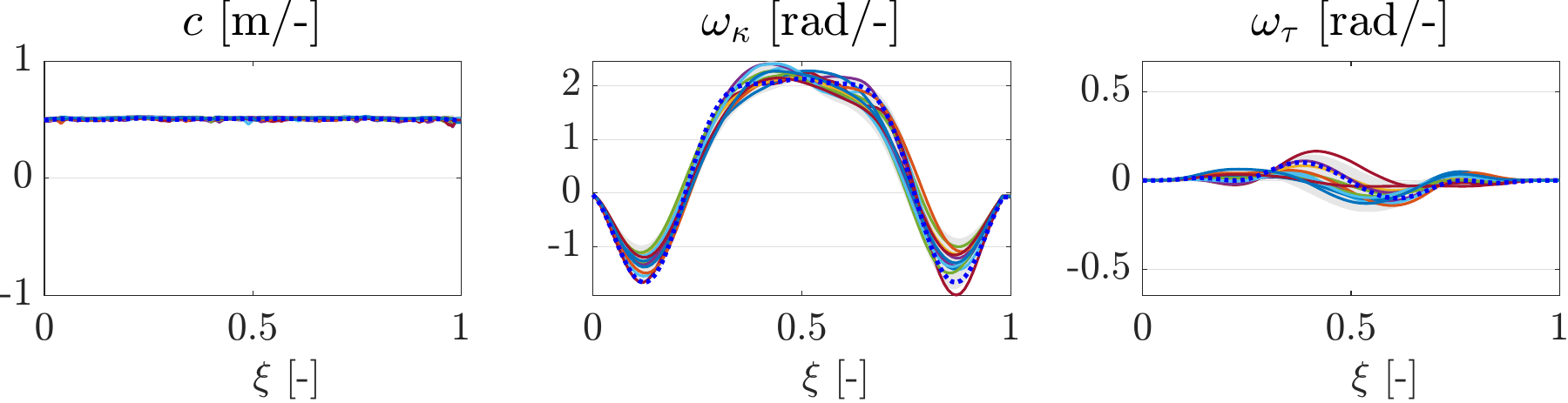}%
		\label{fig:motion_results_positiontoolpoint}}
	\caption{Screw and vector invariants calculated from position and orientation of tracker or tool point on manipulated tool, measured in the world frame.}
	\label{fig:motion_results}
\end{figure}

\begin{figure}[t]
	\centering
	\subfloat[Screw invariants calculated from wrench transformed to tracker frame]{%
		\includegraphics[width=1.0\linewidth]{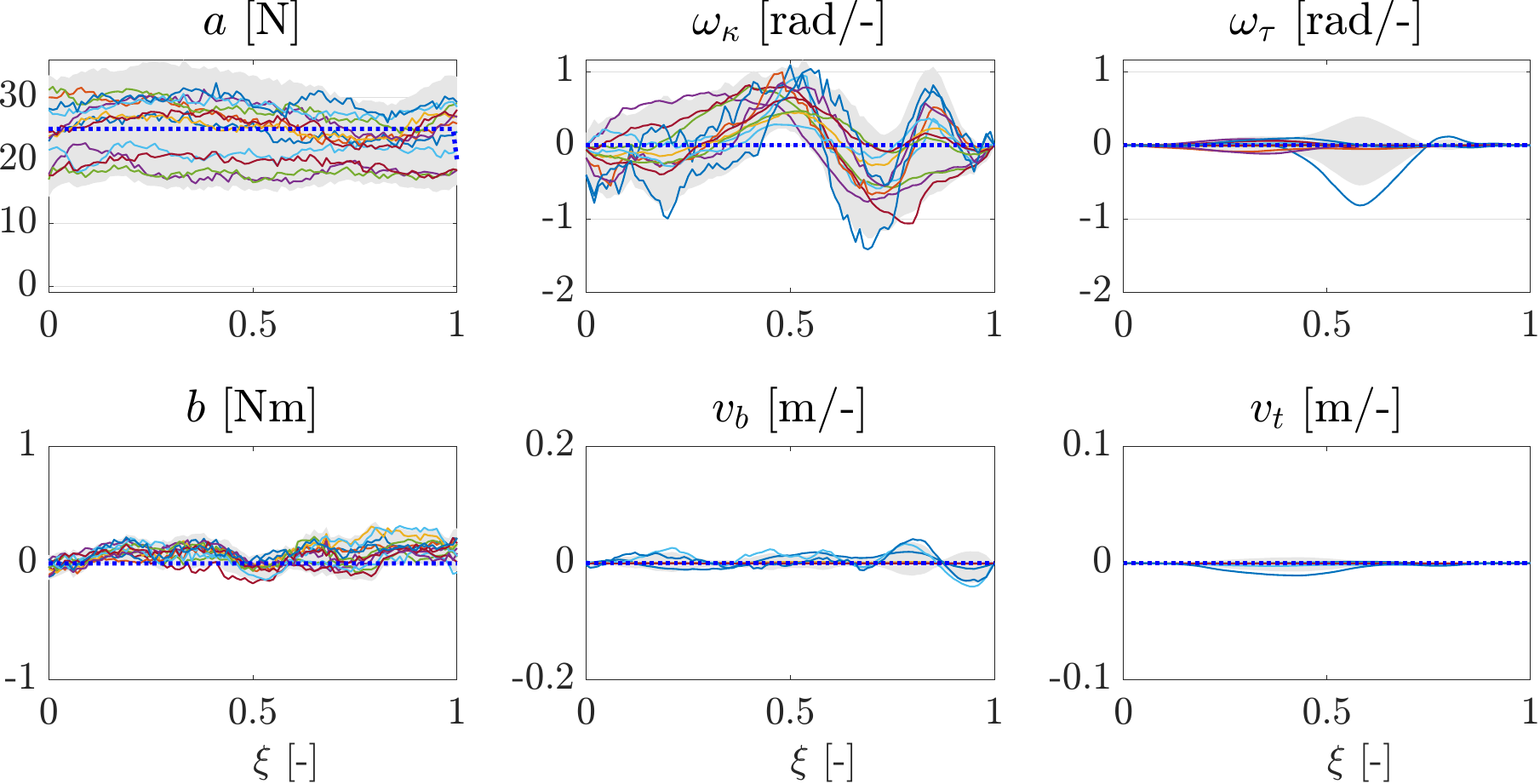}%
		\label{fig:force_results_wrench}} 
	\\
	\subfloat[Vector invariants calculated from force transformed to tracker frame]{%
		\includegraphics[width=1.0\linewidth]{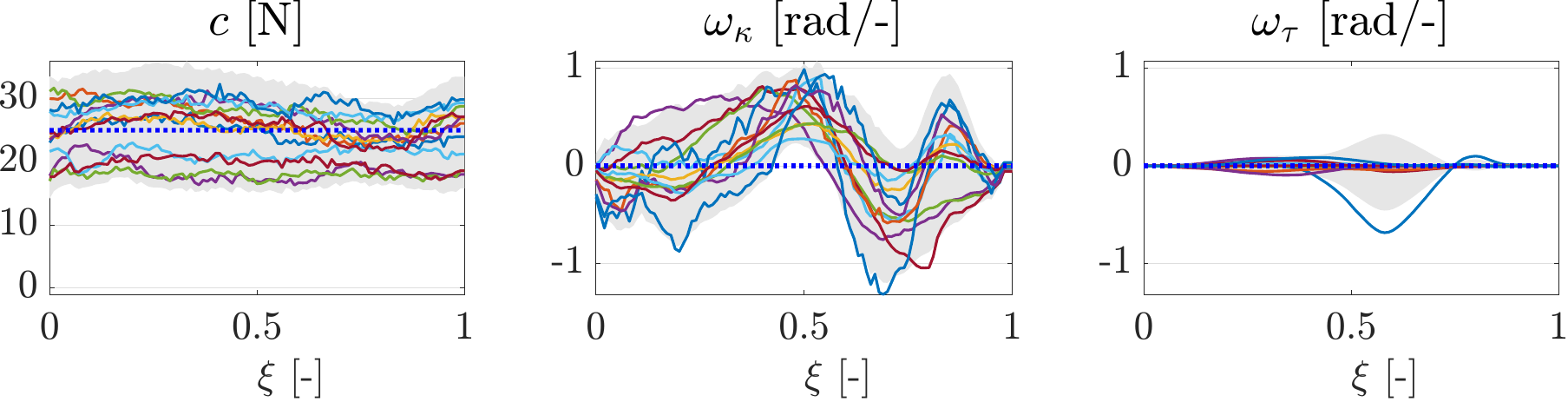}%
		\label{fig:force_results_force}} 
	\\
	\subfloat[Vector invariants calculated from moment transformed to tracker frame]{%
		\includegraphics[width=1.0\linewidth]{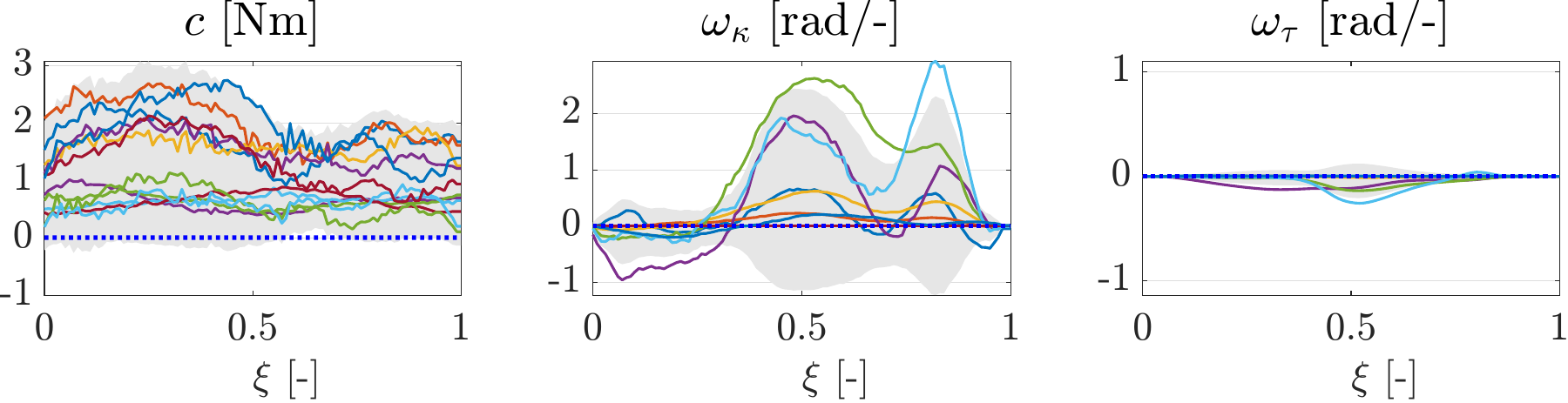}%
		\label{fig:force_results_momenttracker}} 
	\\
	\subfloat[Vector invariants calculated from moment transformed to tool center frame]{%
		\includegraphics[width=1.0\linewidth]{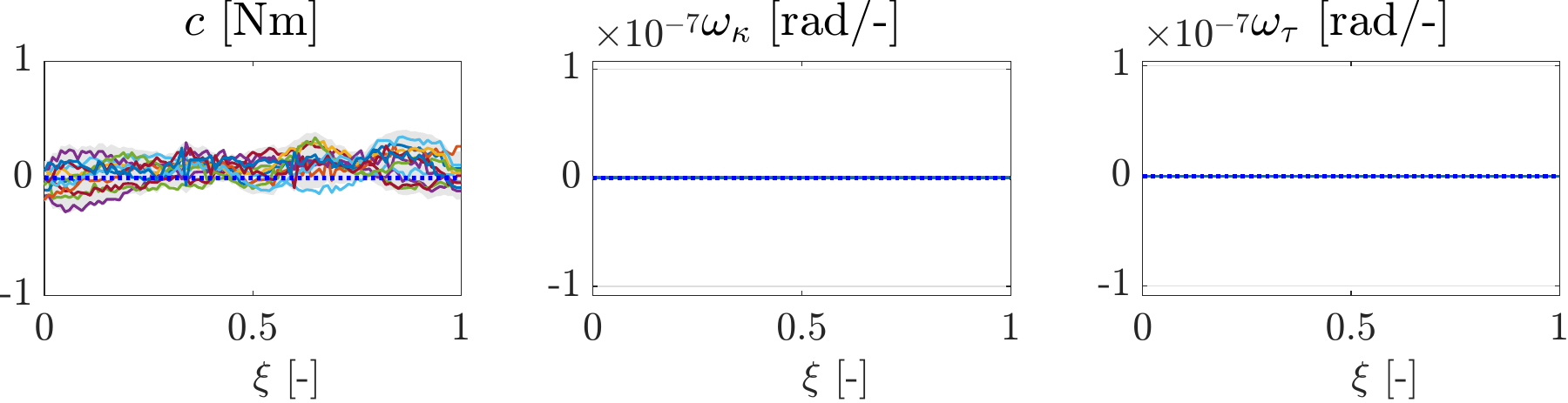}%
		\label{fig:force_results_momenttoolpoint}}
	\caption{Screw and vector invariants calculated from measured force and moment trajectories, transformed to either the tracker or the tool center frame.}
	\label{fig:force_results}
\end{figure}

\begin{figure}[t]
	\centering
	\includegraphics[width=0.95\linewidth]{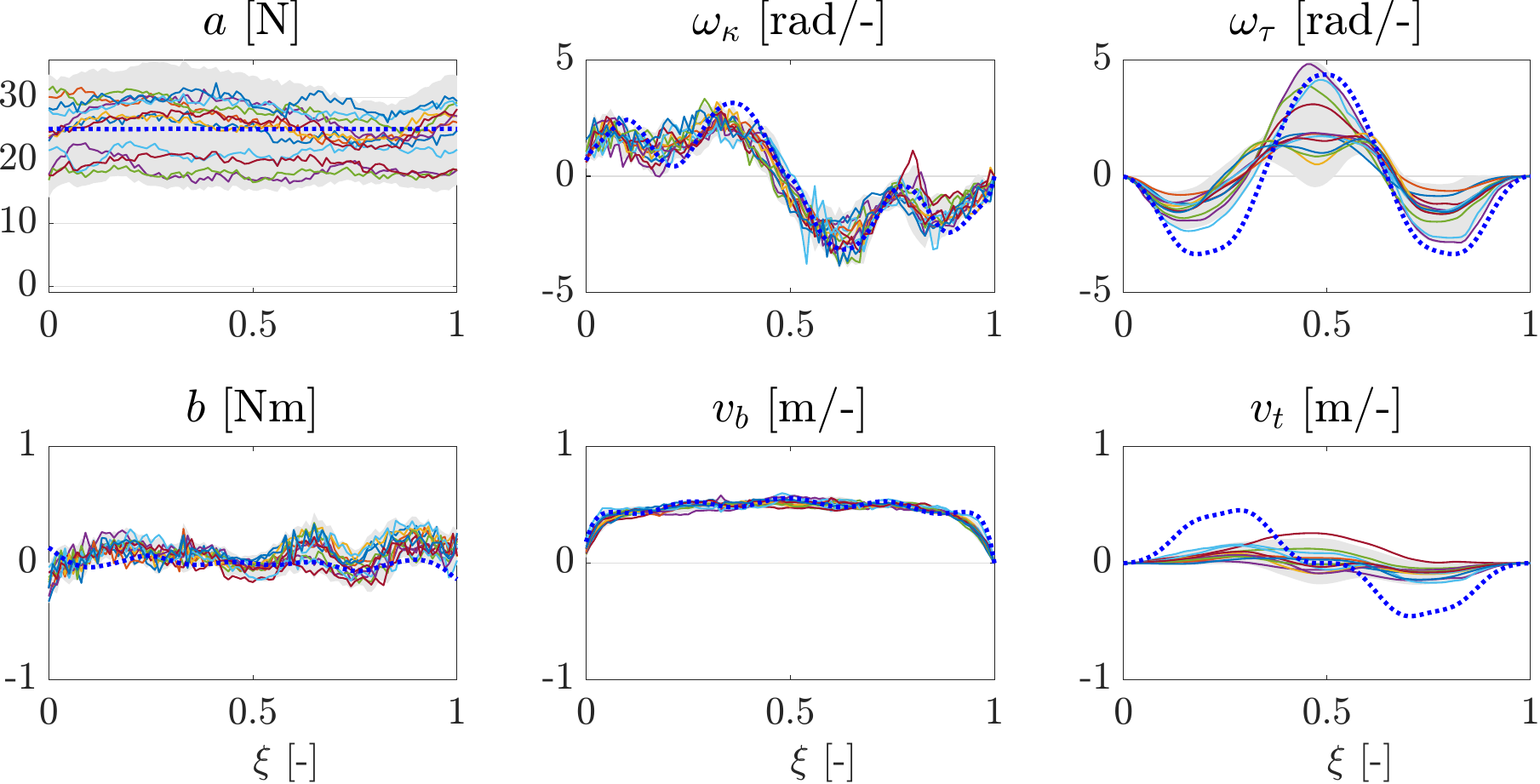}%
	\caption{Screw invariants calculated from measured wrench transformed to the world frame.}%
	\label{fig:force_results_world_meas}
\end{figure}

\begin{figure}[t]
	\centering
	\includegraphics[width=0.95\linewidth]{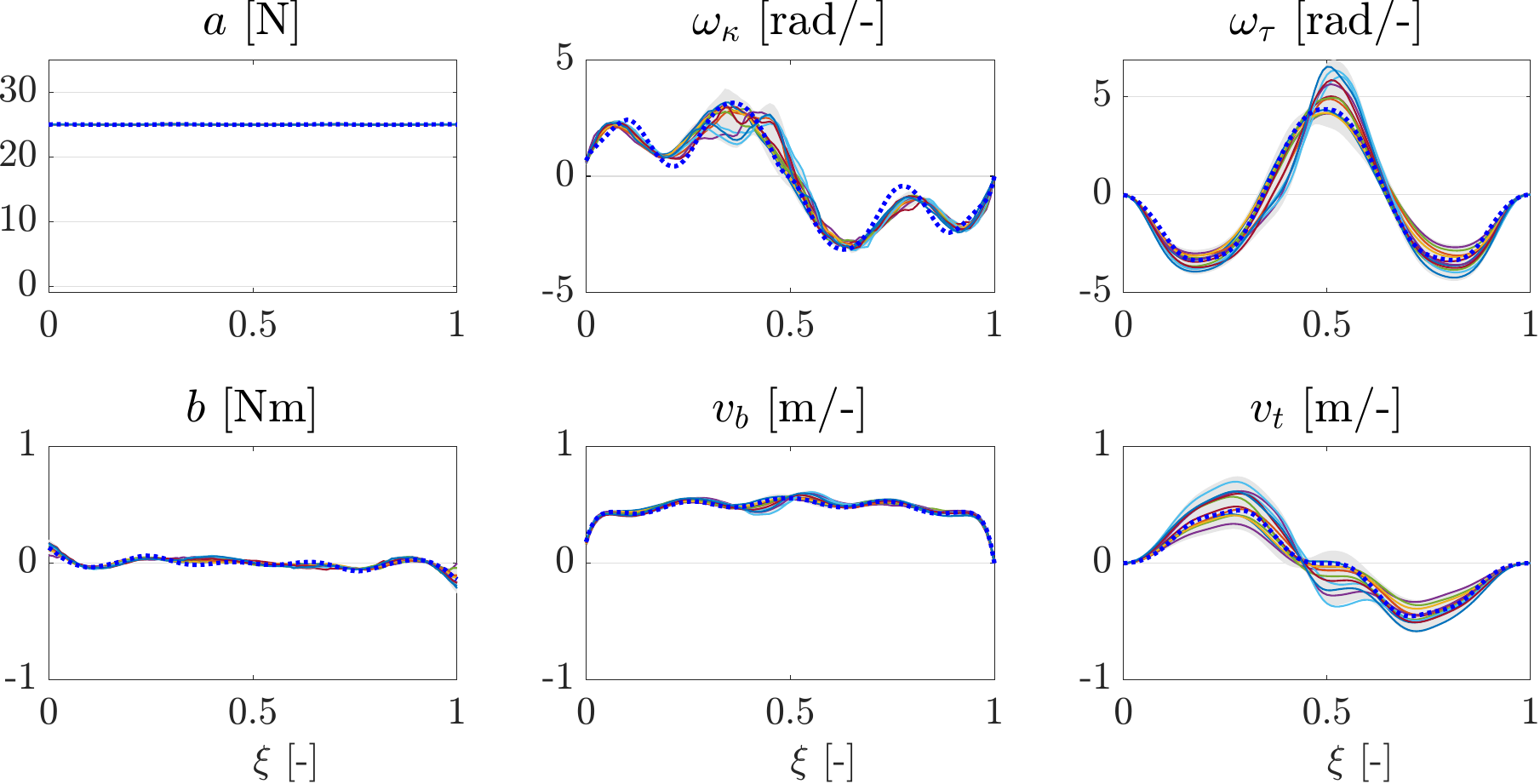}%
	\caption{Screw invariants calculated from a \textit{simulated wrench} in the world frame to study the reduction in variation with respect to Figure \ref{fig:force_results_world_meas}.}%
	\label{fig:force_results_world_simul}
\end{figure}

\begin{figure}[t]
	\centering
	\subfloat[Motion ISA frames in world]{%
		\includegraphics[width=0.50\linewidth]{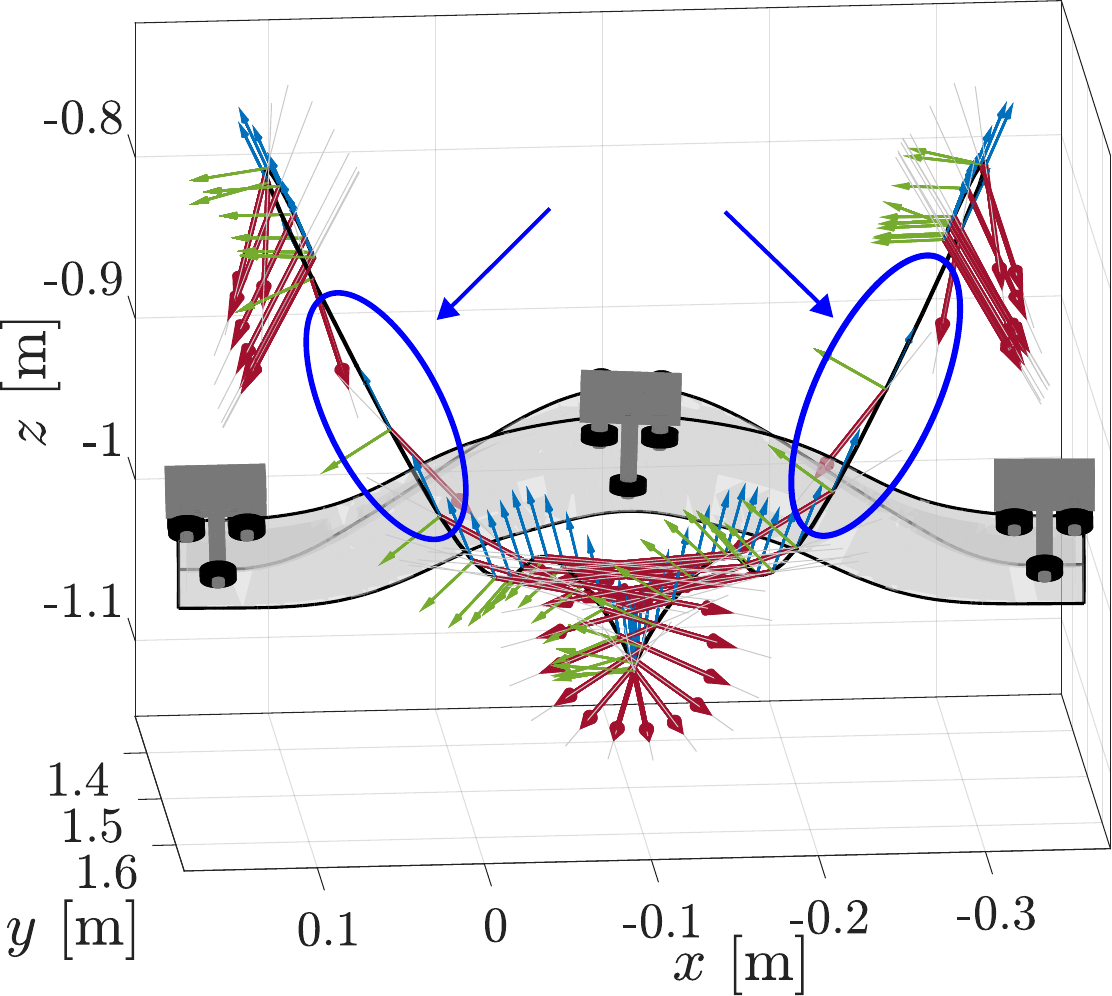}%
		\label{fig:moving_frames_ISA_motion}}
	\hfil
	\subfloat[Force ISA frames in tool]{%
		\includegraphics[width=0.38\linewidth]{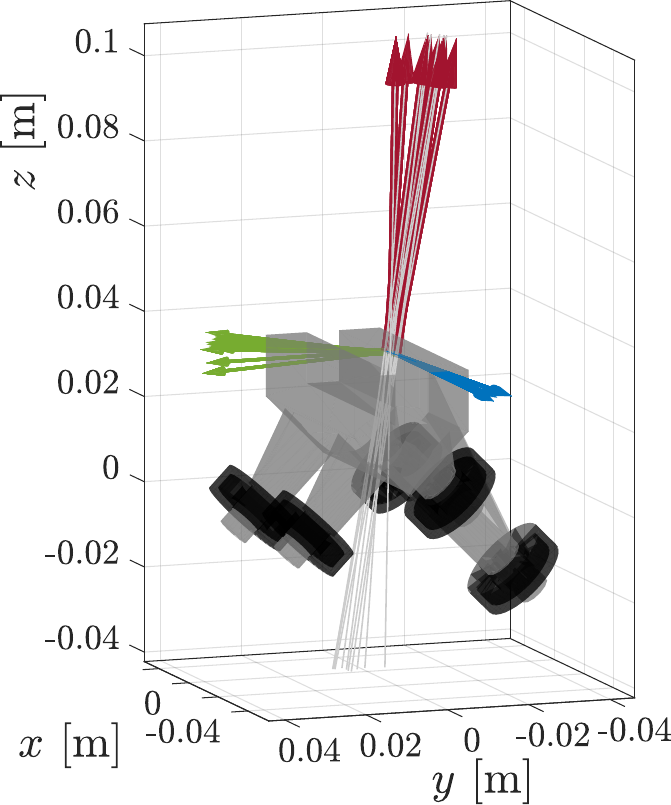}%
		\label{fig:moving_frames_ISA_force}}
	\\
	\subfloat[Force ISA frames in world]{%
	\includegraphics[width=0.495\linewidth]{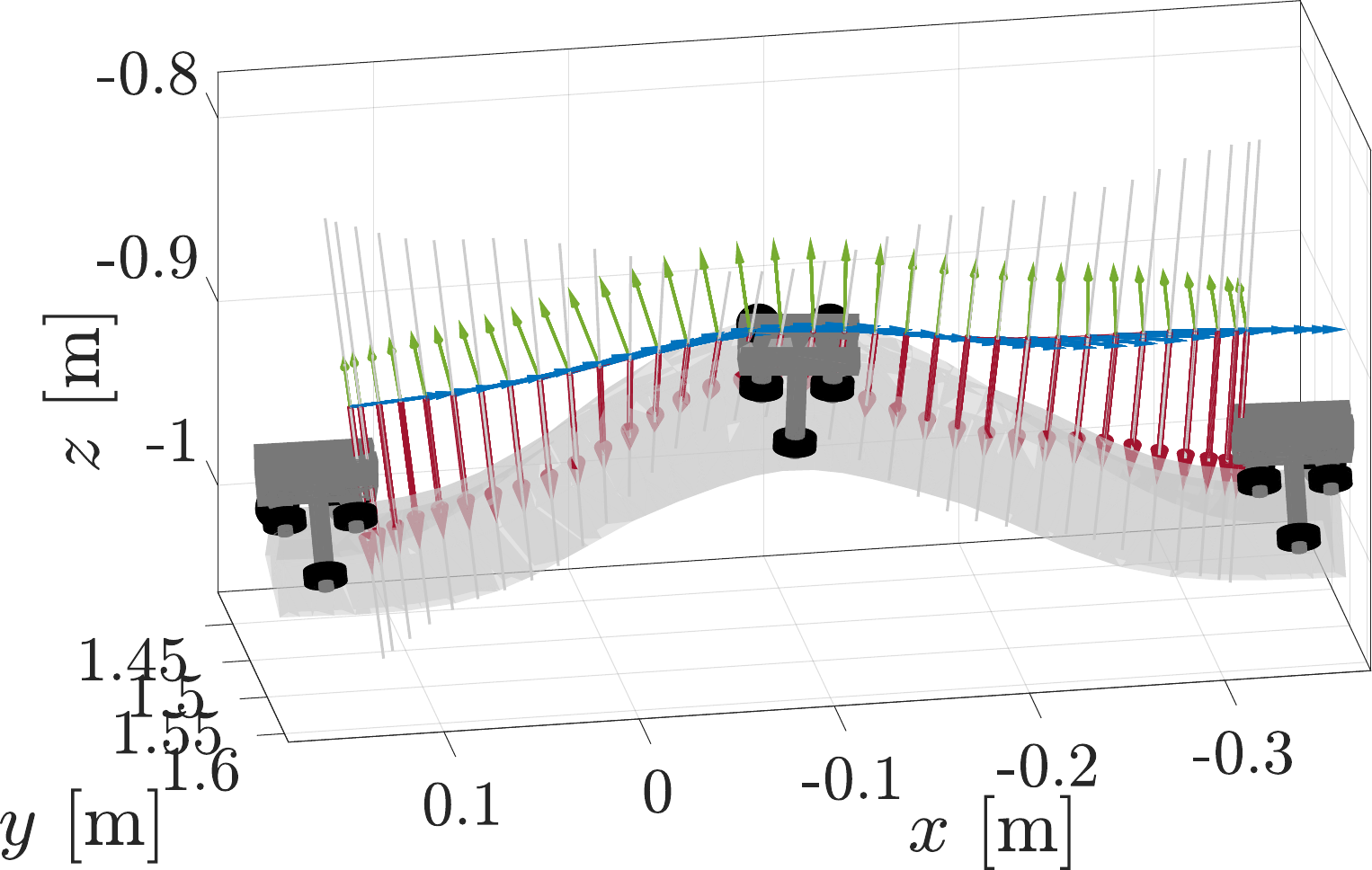}%
	\label{fig:moving_frames_ISA_force_world}}
	\hfill
	\subfloat[Transl. velocity FS frames in world]{%
	\includegraphics[width=0.495\linewidth]{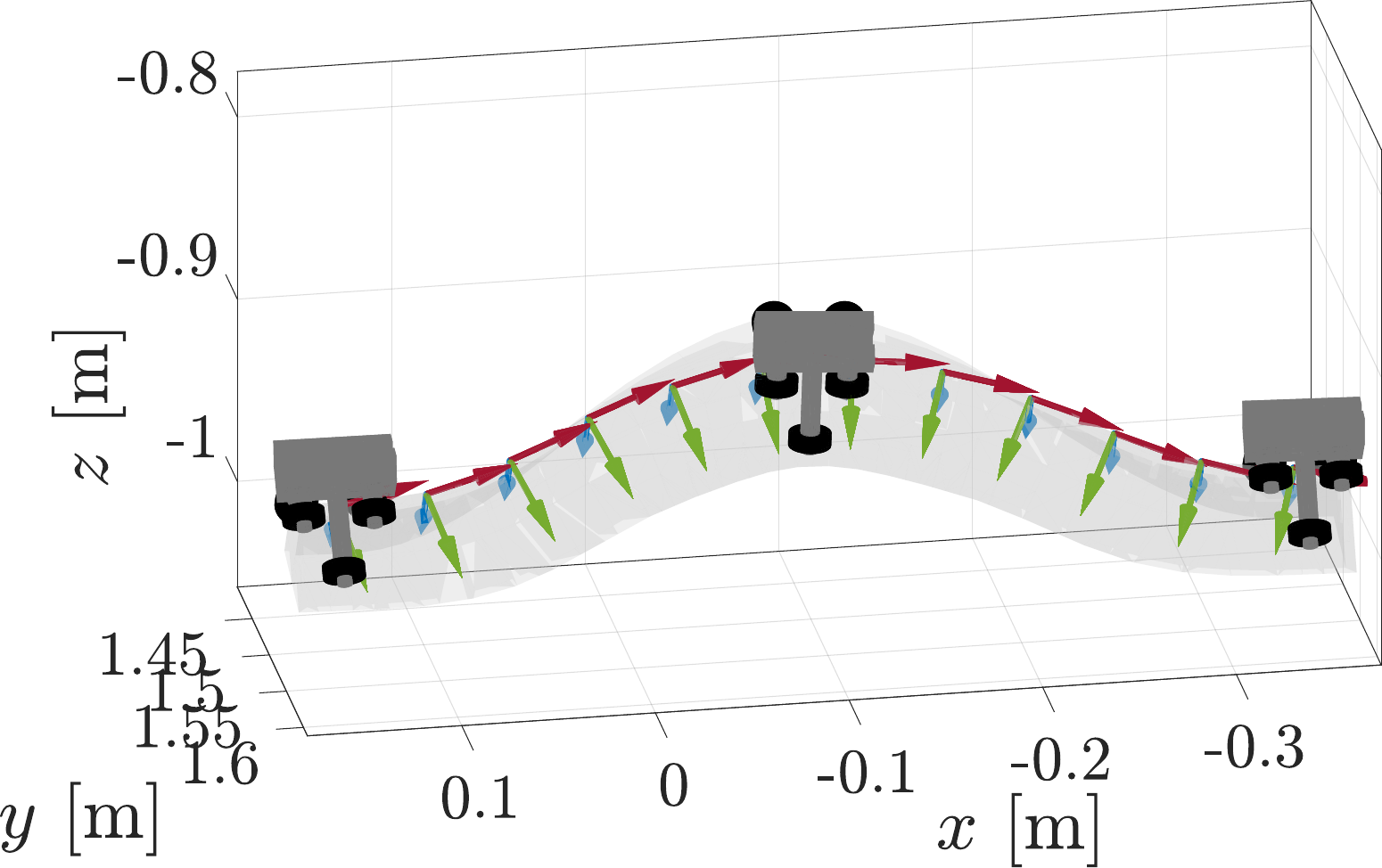}%
	\label{fig:moving_frames_FS_transl}}
	\caption{Moving frames calculated from data of Trial 5: (a) ISA frames calculated from the tool's motion in the world. The red arrows correspond to the ISA itself, which is the first axis of the moving frame. The green and blue arrows correspond to the second and third axes, respectively. The two blue circles indicate the translation of the ISA frames towards the other side of the contour at the inflection points; (b) ISA frames calculated from interaction wrench, relative to the tool; (c) ISA frames calculated from wrench w.r.t the world; (d) Frenet-Serret (FS) frames calculated from the position of the tool center point (TCP). The red arrows correspond to the tangent along the contour, which is the first axis of the moving frame.}
	\label{fig:moving_frames}
\end{figure}

\begin{figure}[t]
	\centering
	\includegraphics[width=0.80\linewidth]{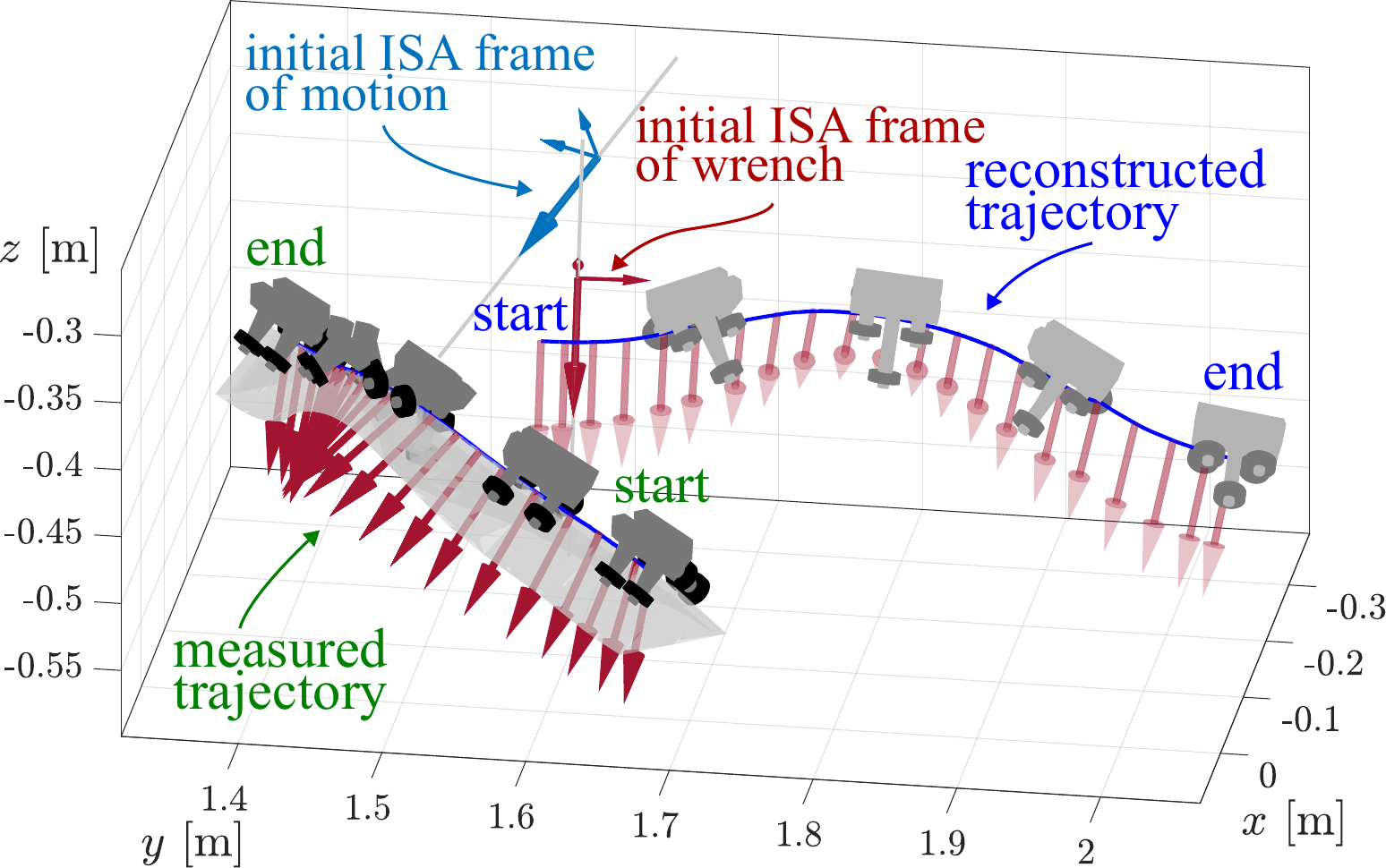}
	\caption{Reconstruction of the motion and wrench trajectory of Trial 7 at a new location given the screw invariants of the measured trajectory seen from the world. The motion of the follower is shown using the position of the TCP and the orientation of the follower. The force on the follower is depicted by the red arrows at the corresponding position of the TCP. The initial ISA frames for motion and wrench, necessary for reconstruction, are also depicted.}
	\label{fig:reconstruction}
\end{figure}

Figures \ref{fig:motion_results}-\ref{fig:reconstruction} summarize the experimental results. Figures \ref{fig:motion_results}-\ref{fig:force_results_world_meas} contain screw and vector invariants: motion invariants with respect to the world (Fig. \ref{fig:motion_results}), force invariants with respect to the tool (Fig. \ref{fig:force_results}), and force invariants with respect to the world (Fig. \ref{fig:force_results_world_meas}). Additional results for other cases can be generated in the provided software \cite{software}. Each plot contains twelve thin lines corresponding to each of the twelve demonstrations. The thick dotted blue line represents the reference values for the limit case. The variation among the demonstrations is given by the two-sigma band depicted in gray. For all invariants, the dimensionless arc length $\xi$ defined in Section~\ref{sec:exp-processing} is chosen as the progress variable. Figure \ref{fig:moving_frames} visualizes the corresponding evolution of the different moving frames for one demonstration. Note that in the subfigures of Fig. \ref{fig:moving_frames}, the contour is only positioned approximately\footnote{This approximation is based on the spatial alignment of the recorded trajectory of $\{tcp\}$ with the edge of the contour \cite{sorkine2017}.} because its true pose was not recorded. Figure \ref{fig:reconstruction} visualizes motion and force trajectories reconstructed from the invariants at a new location. Below, the results are discussed in detail.

\textbf{Motion invariants with respect to world:}
Fig. \ref{fig:motion_results_SAI} shows the screw invariants for motion. The object invariants $a$ and $b$ describe the rotation and translation of the object along the screw axis, while the moving frame invariants $\omega_\kappa$, $\omega_\tau$, $v_b$, and $v_t$ describe the rotation and translation of the moving frame attached to the screw axis. Recalling that each demonstration has different tracker and world frames, the relatively small variations between demonstrations confirm the invariance properties of the screw invariants. Even in the presence of human variations, relatively small variations between the demonstrations and good correspondence with the reference values from the limit case are obtained.  This is explained by the constrained motion, allowing variations in only one dof. Due to the special design of the contour, clear symmetric or antisymmetric profiles with respect to the midpoint were expected and were obtained. The motion of the moving frames for one trial is visualized in Figure \ref{fig:moving_frames_ISA_motion}.

Fig. \ref{fig:motion_results_orientation} shows the vector invariants for orientation. According to the analytical formulas, they should correspond to the invariants for the directional component of the screw for motion (top row in Fig. \ref{fig:motion_results_SAI}). The reason why they are not exactly the same is explained by the OCP-based calculation approach. In this approach, the calculation for orientation and translation are coupled into a single OCP for the screw invariants and hence they are subject to a joint regularization. Nevertheless, the overall evolution of the vector invariants for orientation and the invariants for the directional component of the screw are still similar.

Turning to the vector invariants for translation, Fig. \ref{fig:motion_results_positiontracker} confirms that they are not invariant for changes of the tracker location, serving as the reference point to define the translation of the tool. This non-invariance can be resolved by picking the same reference point for all demonstrations, in this case the origin of $\{tcp\}$, as shown in Fig. \ref{fig:motion_results_positiontoolpoint}. In this figure, $c$ represents the magnitude of the translational velocity vector, while $\omega_{\kappa}$ and $\omega_{\tau}$ represent the dimensionless curvature and torsion of the point curve tracked by the origin of $\{tcp\}$. From all the figures, Fig. \ref{fig:motion_results_positiontoolpoint} exhibits the best correspondence between the reference limit case and the demonstrations. This is to be expected since the limit case models the translation of the $\{tcp\}$-frame along the edge. For this limit case, $c$ is approximately constant (about 0.5 m/-). Its integral (also close to 0.5 m) represents the arc length of the point curve tracked by the origin of $\{tcp\}$. The small values for the torsion rate indicate that this point curve lies, at least locally, approximately in a plane. The small differences between the limit case and the demonstrations are due to the finite dimensions of the tracking tool and due to measurement errors. The moving FS frames corresponding to these vector invariants for one demonstration are visualized in Fig. \ref{fig:moving_frames_FS_transl}.

Given the geometry-based definition of the invariants, they can be used to segment the trajectory into meaningful parts. For example, in the screw invariants in Fig. \ref{fig:motion_results_SAI}, we notice two inflection points at $\xi\approx 0.2$ and at $\xi\approx 0.8$: while the rotational velocity $a$ changes sign at these progress values, the peaks in $v_b$ show that the position of the ISA is rapidly shifting from a center of curvature on one side of the contour to the other side of the contour. This is confirmed by the visualization of the corresponding ISAs in Fig. \ref{fig:moving_frames_ISA_motion}.

\textbf{Force invariants with respect to tool:}
More variation among demonstrations was expected in the wrench invariants because, although the operator was instructed to apply a pure force in a specified direction and through a specific point of the tool, the operator could at any time apply a wrench in a local 5-dof space. The higher noise levels associated with force measurements may cause additional variation compared to the motion invariants. Note that a pure force that remains parallel to itself corresponds to the second type of singularity, mentioned in Section \ref{sec:invars_screw}. This singularity applies to both vector and screw invariants.

Fig. \ref{fig:force_results_wrench} shows the screw invariants. Recalling that each demonstration has a different transformed force sensor frame, these plots confirm that the screw invariants are invariant for such transformation. The plot of invariant $a$ shows that the operator was able to maintain relatively constant magnitudes of the force along the contour, although the magnitudes varied between demonstrations. The plots of $\omega_{\kappa}$ and $\omega_{\tau}$ reveal a limited change of direction of the force (hence of the ISA), almost in a single plane ($\omega_{\tau}$ is very small). This plane is orthogonal to the edge of the contour. This can be found by analyzing the direction of the ISAs and is confirmed in Fig. \ref{fig:moving_frames_ISA_force}, where the first (red) and second axes (green) of the successive moving frames lie approximately in a plane that is orthogonal to the tangent of the contour.

The second row of Fig. \ref{fig:force_results_wrench} shows that there is no significant moment about the ISA ($b$ is very small) and that the origin of the moving frame does not translate much, both parallel to and along  the ISA  (both $v_b$ and $v_t$ remain very small). The latter can also be seen in Fig. \ref{fig:moving_frames_ISA_force}.

Fig. \ref{fig:force_results_force} shows the vector invariants for force, which are nearly identical to the directional component of the screw invariants (top row in Fig. \ref{fig:force_results_wrench}). As for the vector invariants for moment, Fig. \ref{fig:force_results_momenttracker} confirms that they are not invariant for a change in position of the force sensor frame, resulting in moment magnitudes up to 2.5 Nm. Similarly as for translational velocity, this can be resolved by picking the TCP as the common reference point for expressing the moments in all demonstrations, as shown in Fig. \ref{fig:force_results_momenttoolpoint}. Since these moment values are relatively close to zero w.r.t. the specified tolerance $\epsilon_m=0.16~\text{Nm}$, they can be perfectly described by a stationary Frenet-Serret frame. Therefore, the solutions of $\omega_\kappa$ and $\omega_\tau$ are below $10^{-8}$, the stopping criterion in the OCP solver.

\textbf{Force invariants with respect to world:}
Figure \ref{fig:force_results_world_meas} shows the screw invariants for the simulated, alternative experimental set-up in which the force sensor is assumed fixed to the world. This simulation, in particular the transformation of the physical force measurements to the world, is subject to two additional errors: the calibration between the force sensor and the motion tracker, and the camera measurements. Again, the invariants are shown to be insensitive to this transformation.
Fig. \ref{fig:moving_frames_ISA_force_world} shows the evolution of the corresponding moving frame. During execution, the force ISA remains approximately perpendicular to the contour edge while it approximately intersects the contour edge. To do this for the entire trajectory, the force axis has to translate along the contour with a velocity $v_b$, which is approximately equal to the translation of the TCP per dimensionless arc length (about $0.5$ m/-). Integrating this value over the total horizon again results in the length of the entire contour ($\sim$0.5 m).

To further study the variation in the force invariants with respect to the world, we replaced the real wrench measurements in each demonstration with a constant simulated wrench $\boldsymbol{w}=25\left[0~~\mathrm{cos}(\frac{\pi}{4})~~\mathrm{sin}(\frac{\pi}{4})~~0~~0~~0\right]^T$ expressed in the TCP frame. By using this simulated wrench, the effects of human variations and wrench measurement noise were eliminated. Consequently, the corresponding force invariants with respect to the world are only influenced by errors in the calibration between the TCP and the motion tracker, and noise in the motion measurements.  Invariant descriptors of this case are shown in Fig. \ref{fig:force_results_world_simul}. We observe greater repeatability in force invariants compared to Fig. \ref{fig:force_results_world_meas} as a result of eliminating the human variation and wrench measurement noise. This analysis confirms that the larger variation in the force invariants compared to motion invariants is mainly due to human variation in a 5-DOF space, not due to errors in the computation.

\textbf{Reconstructed motion and force trajectories:} Figure~\ref{fig:reconstruction} visualizes a reconstruction of the motion and force trajectory at a new location starting from the invariants of one of the trials. Table~\ref{tab:rec_acc} compares these reconstructed trajectories with the measured trajectories on which a global transformation has been applied so that they are in the same location. The RMS-differences between the trajectories were found to be in agreement with the chosen tolerances for the trajectory accuracy constraint in the OCPs. The small difference is due to the convergence tolerance of the numerical solver of the OCPs, which was set to $10^{-8}$. These results confirm that the trajectories reconstructed from the calculated invariants have the accuracy specified in the OCPs and are also driftless. Evidently, these results also hold for the special case of a reconstruction at the same location as the measurements.

\begin{table}[t]
	\centering
	\caption{Reconstruction error for the motion and force trajectory of Figure \ref{fig:reconstruction}. The chosen tolerances are given as a reference.}
	\label{tab:rec_acc}
	\begin{tabular}{ lll }
		\toprule
		\textbf{Trajectory}  & \textbf{RMS reconstruction error} & \textbf{Tolerance in OCP} \\
		\midrule
		position                      & $2.00 + 7\times10^{-9}~\mathrm{mm}$  & $\epsilon_p = 2.00~\mathrm{mm}$   \\
		orientation 			      & $2.00 + 9\times10^{-9}~{}^{\circ}$  & $\epsilon_R = {2.00}^{\circ}$     \\
		force                         & $0.80 + 4\times10^{-9}~\mathrm{N}$  & $\epsilon_f = 0.80~\mathrm{N}$    \\
		moment                        & $0.16 + 8\times10^{-9}~\mathrm{Nm}$  & $\epsilon_m = 0.16~\mathrm{Nm}$   \\
		\bottomrule
	\end{tabular}
\end{table}

\section{Application to Peg-on-Hole Alignment}
\label{sec:experiments2}

\begin{figure}[t]
	\centering
	\subfloat[Setup]{%
		\includegraphics[width=0.36\linewidth]{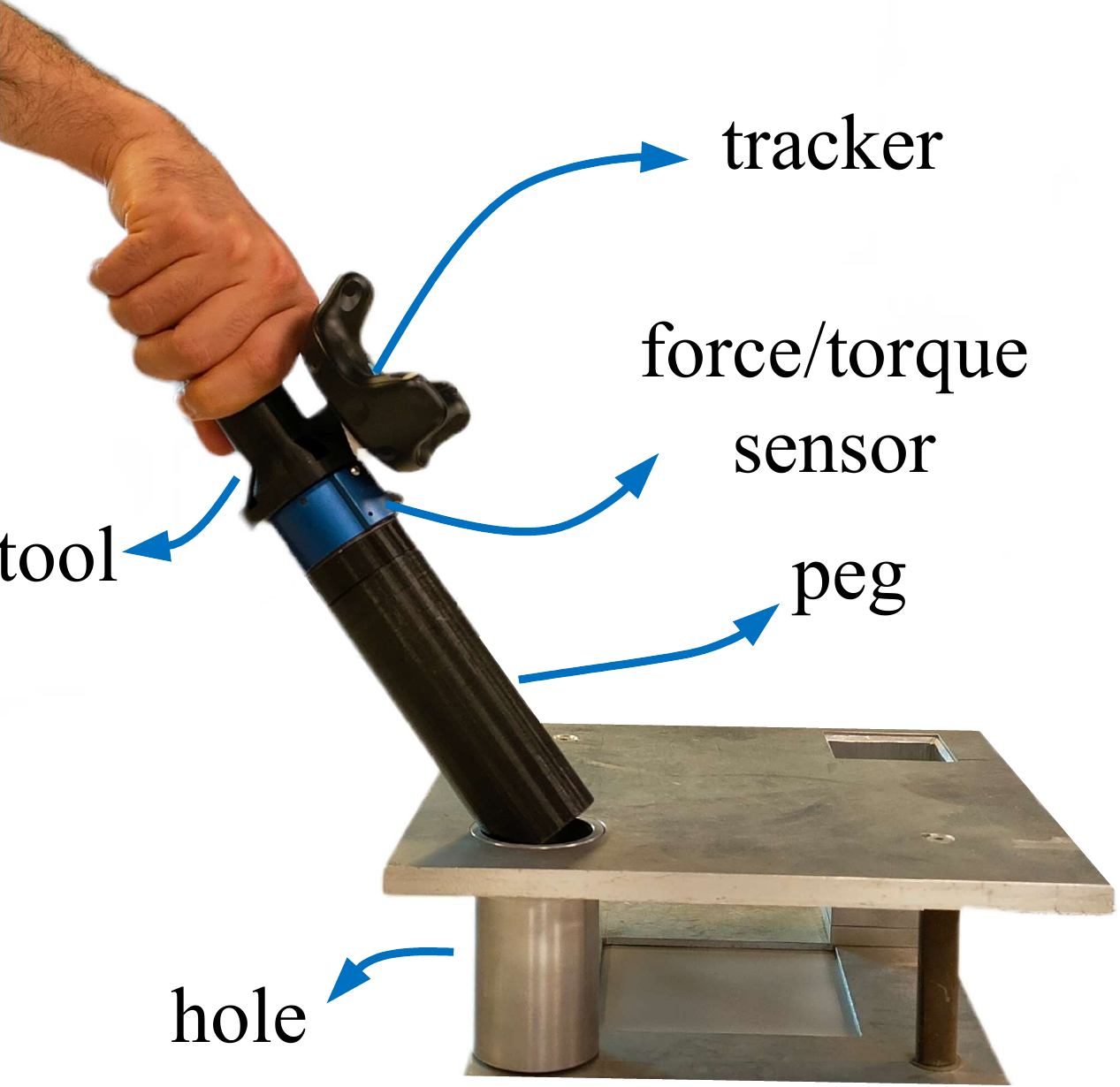}%
		\label{fig:setup_peg_demo}}
	\hfill
	\subfloat[Frames]{%
		\includegraphics[width=0.170\linewidth]{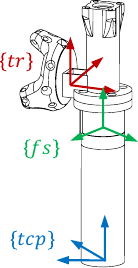}%
		\label{fig:setup_peg_frames}}
	\hfill
	\subfloat[Three-point contact]{%
		\includegraphics[width=0.38\linewidth]{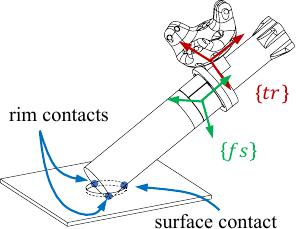}%
		\label{fig:3pointcontact}}
	\caption{Peg-on-hole setup with defined frames and illustration of the contact.}
	\label{fig:setup_peg}
\end{figure}

This section examines motion and force trajectories resulting from a human-demonstrated peg-on-hole task in which the operator holds a tool to which the peg with diameter $50~\text{mm}$ is attached, as shown in Fig. \ref{fig:setup_peg}. 
The task aims to ensure that peg and hole are aligned before peg insertion starts, even if the hole is inaccurately located. To accomplish this, the operator first establishes a stable
three-point contact between the hole and a purposefully misaligned peg, as shown
in Fig. \ref{fig:3pointcontact}. This contact involves two rim-rim contact points and one
surface-rim contact point. We recorded 12 demonstrations of the alignment, each starting from such initial three-point contact and finishing at full alignment. The operator was instructed to perform alignment while maintaining the three-point contact.
All demonstrations were performed by the same operator, but in two batches, respectively of 7 and of 5 trials, at different points in time.

Although this task appears to involve a pure rotation, previous
theoretical studies have shown that, apart from a rotation, the motion involves a small, but non-zero translation, see e.g. \cite{bruyninckx1995peg}. It is evident that, due to the three-point contact and neglecting friction, this task involves two instantaneous 3-dof vector spaces: one for the wrench and one for the motion. So, the operator can use different motion and wrench strategies during the demonstration.

\subsection{Experimental set-up and data processing}

The motion capture system is the same as for the contour following task of Section \ref{sec:experiments}. The force/torque sensor is again a 6-axis JR3 sensor with a nominal accuracy of $\pm1\%$, but its measurement ranges are smaller: $\pm200\ \text{N}$ along $Z$ and $\pm100\ \text{N}$ along $X$ and $Y$ for force, and $\pm5\ \text{Nm}$ in all directions for moment.
The same three frames as on the contour following tool were defined on the demonstration tool here (see Fig. \ref{fig:setup_peg_frames}), with the TCP frame located at the tip of the peg.

Unlike in the contour following application, we did not use different tracker positions on the tool and did not artificially transform the pose measurements by changing the world frame, because invariance with respect to such variations was sufficiently proven in the contour following experiment. We also did not transform the wrench measurements to different locations on the tool, but we did transform them to a frame fixed to the world to investigate the wrench invariants with respect to the world.

Data was processed in roughly the same way as in the contour following application, with the following changes or additions. The segmentation thresholds were set to $1 ~\text{N}$ and $0.35 ~\text{rad/s}$ for force and angular velocity, respectively. Since the task is predominantly rotational, the integral of the rotational velocity was chosen as the progress variable to re-parameterize the time-based input trajectories, and it was also rescaled to a dimensionless variable ranging from $0$ to $1$. The initialization of the OCPs based on average moving frames (Section \ref{sec:robust_calc_vector} and \ref{sec:robust_calc_screw}) was done on the complete measured trajectories. 
Furthermore, since the force/torque sensor was more accurate than the one used for contour following (same relative accuracy for a smaller range), we lowered tolerances $\epsilon_f$ and $\epsilon_m$ in the OCPs to $0.3~\text{N}$ and $0.1~\text{Nm}$, respectively.

Finally, to investigate if the operator used different wrench strategies in the two batches, we calculated the invariants corresponding to the average of the wrench trajectories for each of the two batches and compared them to the invariants corresponding to the average of all demonstrations. This was done both with respect to the tool and with respect to the world. Three remarks have to made here: 1) we took the average of the wrench trajectories rather than the average of the invariants, because averaging is a linear operation, but the subsequent calculation of the invariants is not; 2) we can just average the measured wrench trajectories, because all wrench trajectories were recorded at the same physical sensor location on the tool and were additionally transformed to the same virtual location in the world\footnote{If the wrench trajectories are recorded at different locations, we can always calculate the invariants of each wrench trajectory and then reconstruct all wrench trajectories at a common new location, as explained in Section \ref{subsec:recon}.}; 3) because  averaged wrench trajectories are less noisy, we could further lower $\epsilon_f$ to $0.17~\text{N}$ and $\epsilon_m$ to $0.05~\text{Nm}$.

\subsection{Results and discussion}

\begin{figure}[t]
	\centering
	\subfloat[Screw invariants for motion with respect to world]{%
		\includegraphics[width=1.0\linewidth]{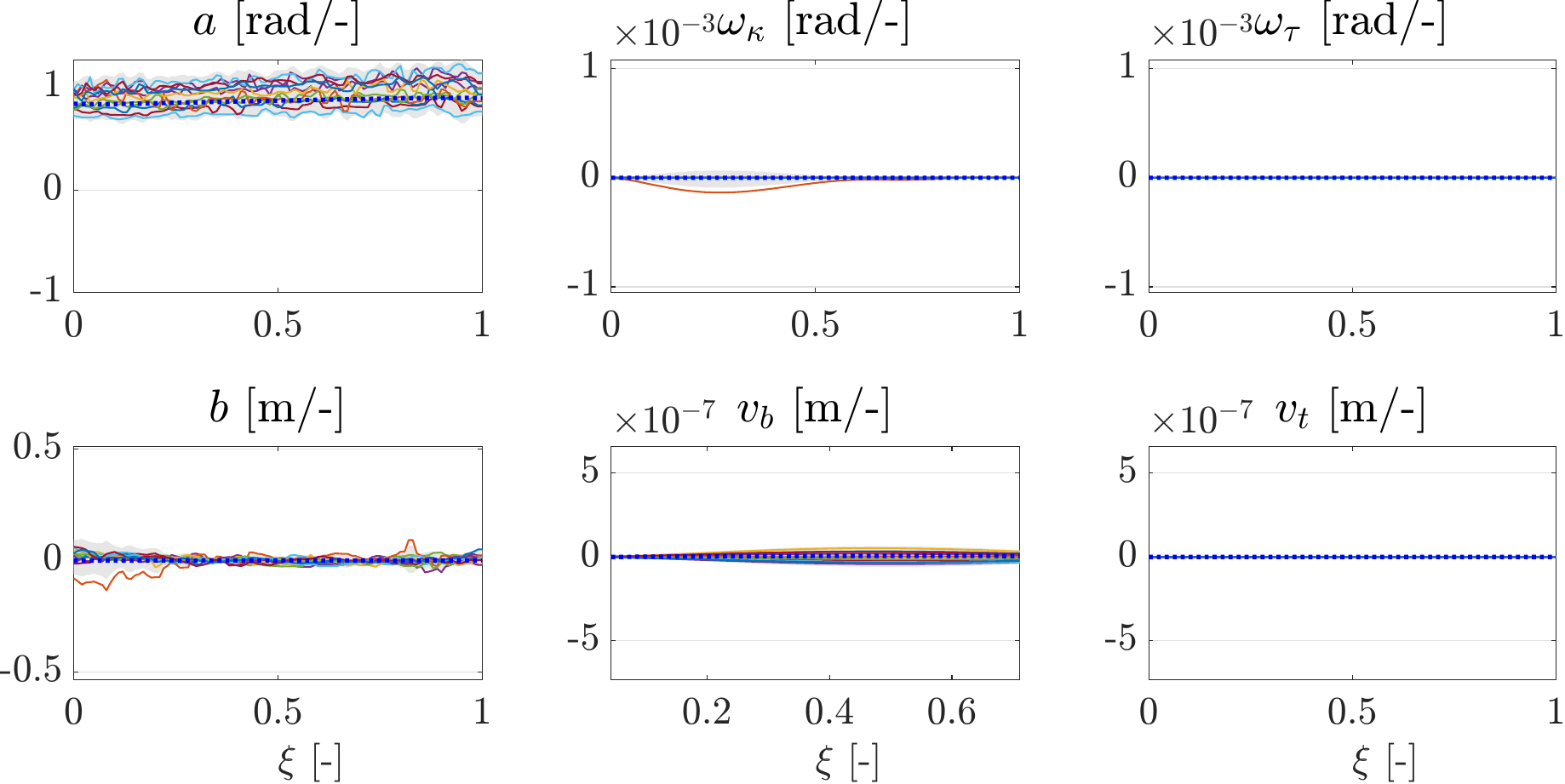}%
		\label{fig:results_peg_motion}} 
	\hfil
	\subfloat[Screw invariants for force with respect to world]{%
		\includegraphics[width=1.0\linewidth]{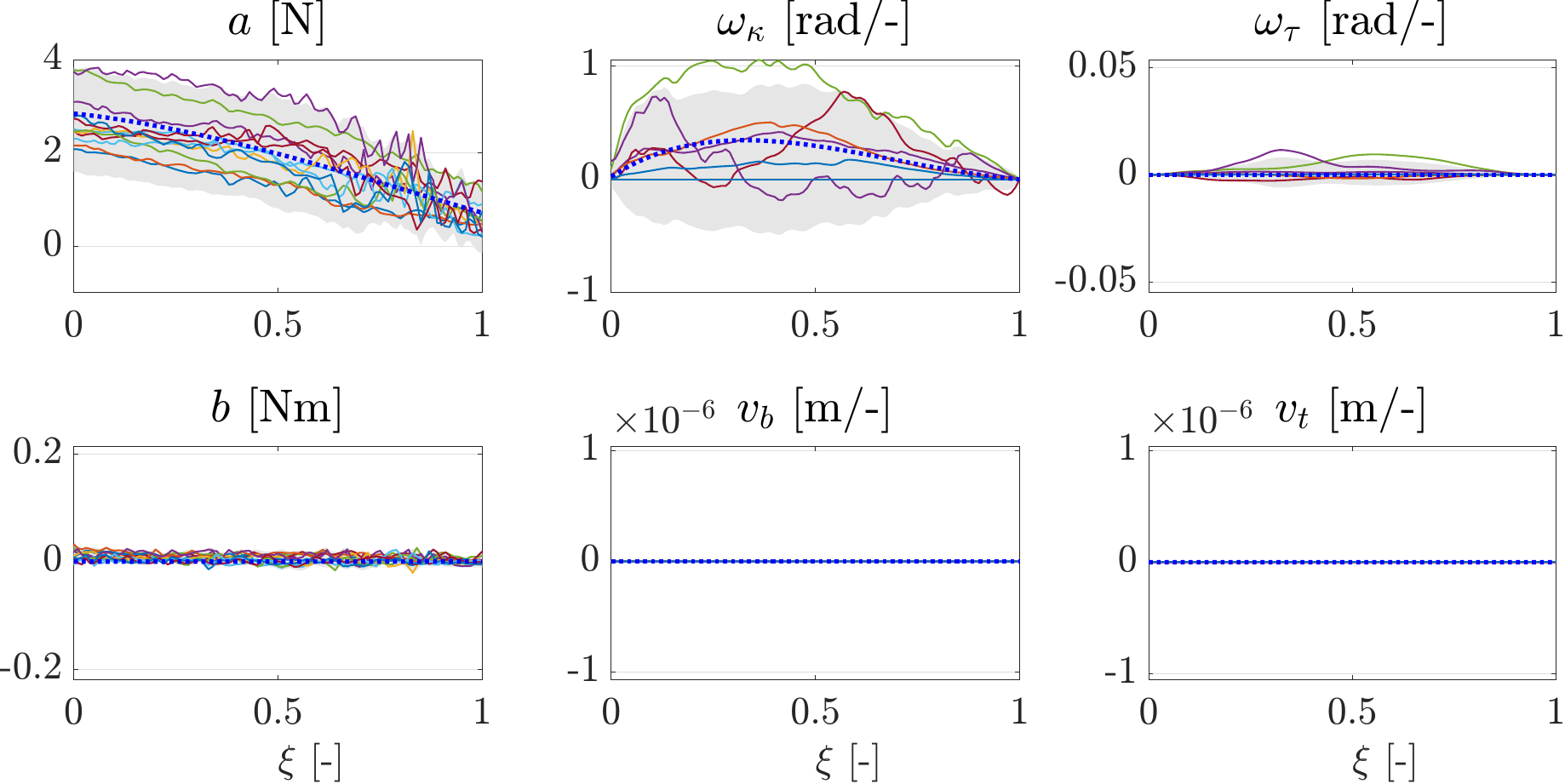}%
		\label{fig:results_peg_force}}
	\hfil
	\subfloat[First two screw invariants for the averaged wrench trajectories of measurement batch 1 and 2, with respect to either world or tool.]{%
		\includegraphics[width=1.0\linewidth]{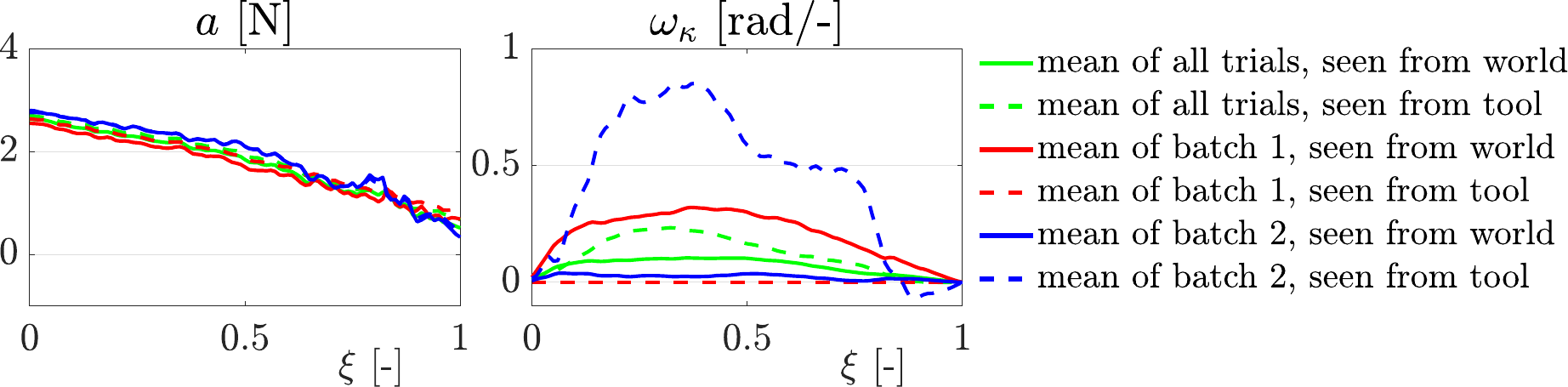}%
		\label{fig:results_peg_average}}
	\caption{Screw invariants for motion and force for the peg-on-hole task.}
	\label{fig:results_peg}
\end{figure}

\begin{figure}[t]
	\centering
	\subfloat[Motion ISA Trial 1]{%
		\includegraphics[width=0.32\linewidth]{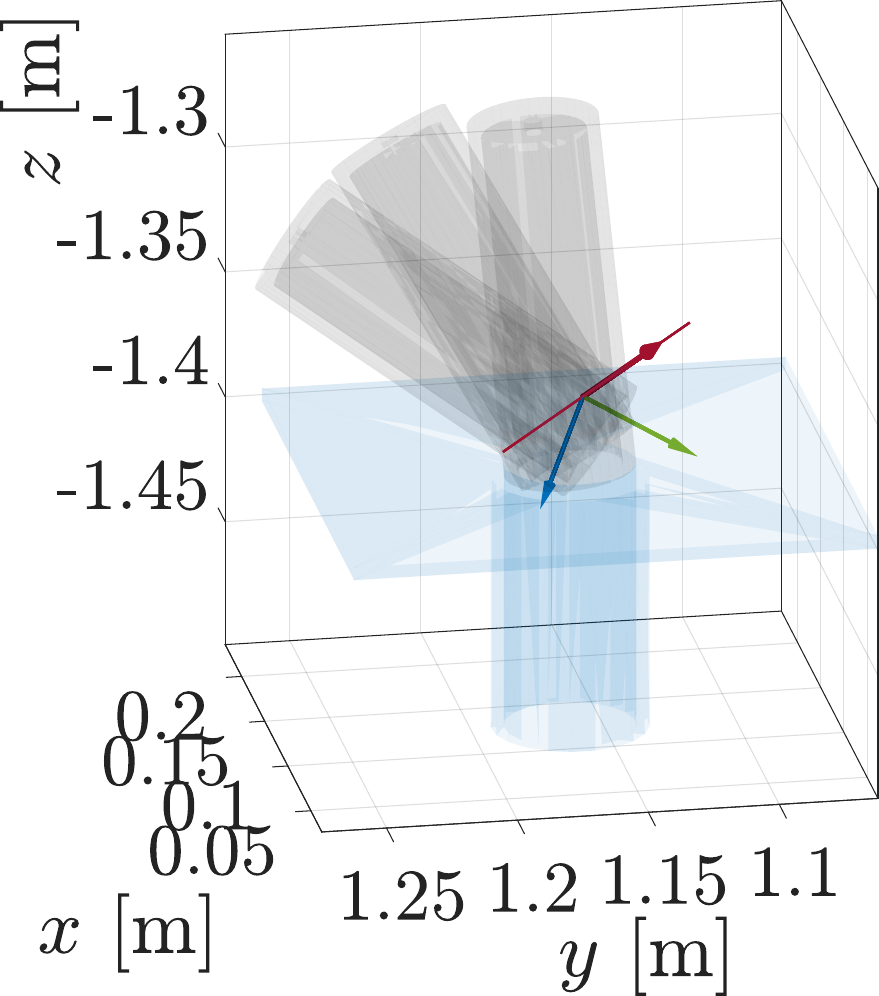}%
		\label{fig:moving_frames_peg_motion}}
	\hfil
	\subfloat[Force ISA Trial 1]{%
		\includegraphics[width=0.32\linewidth]{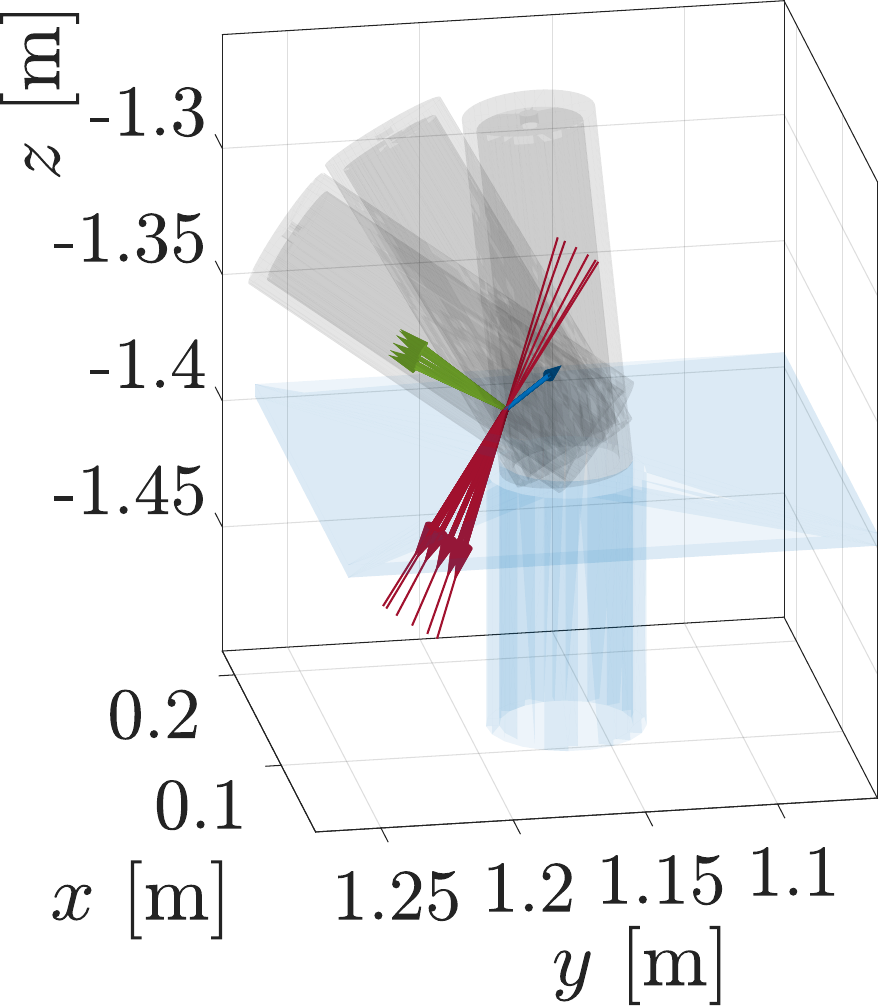}%
		\label{fig:moving_frames_peg_force_batch1}}
	\hfil
	\subfloat[Force ISA Trial 12]{%
		\includegraphics[width=0.33\linewidth]{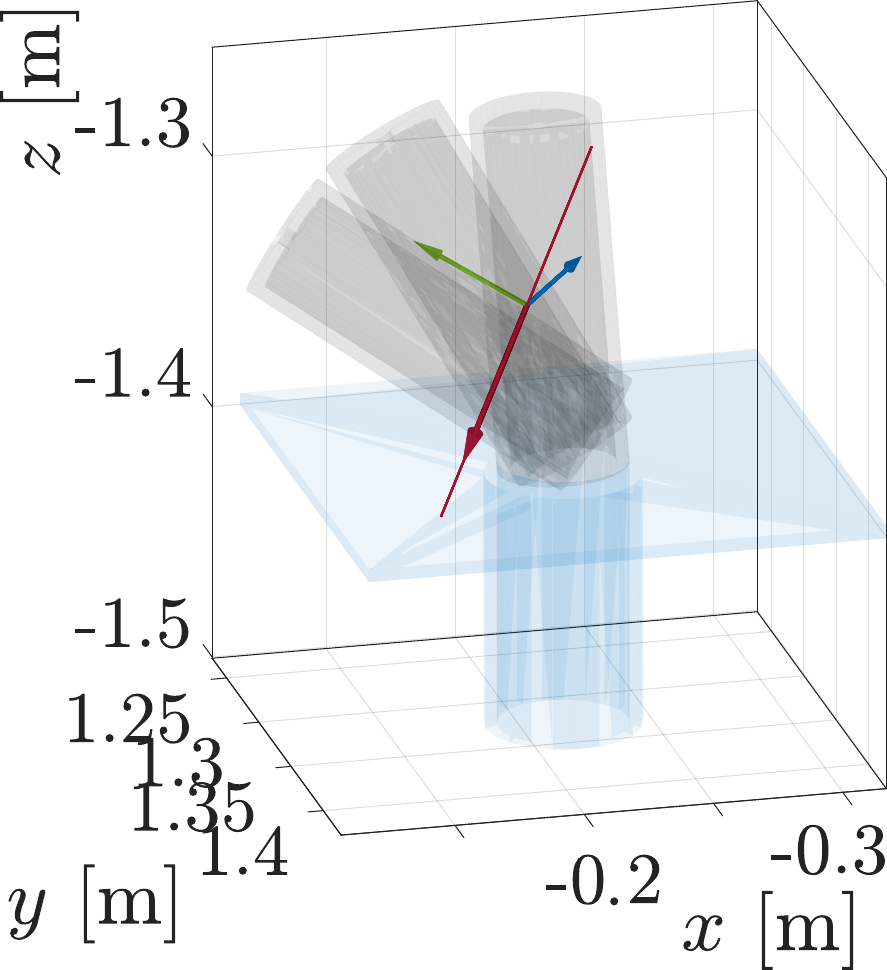}%
		\label{fig:moving_frames_peg_force_batch2}}
	\caption{Moving frames for trials in the peg-on-hole alignment task, all seen from the world frame. The ISAs are depicted using red arrows: (a) Moving frames calculated from the tool's motion for Trial 1 (batch 1), showing a stationary ISA; (b) Moving frames calculated from the wrench for Trial 1 (batch 1), showing a moving ISA; (c) Moving frames of the wrench for Trial 12 (batch 2), showing a stationary ISA.}
	\label{fig:moving_frames_peg}
\end{figure}

Figure \ref{fig:results_peg} and \ref{fig:moving_frames_peg} summarize the experimental results, focusing on screw invariants for motion and force. Additional results for the vector invariants can be generated in the provided software \cite{software}.

\textbf{Motion invariants with respect to world}: Fig. \ref{fig:results_peg_motion} shows the screw invariants of motion. The object invariant $a$ corresponds to the relative rotation of the peg with respect to the hole around the screw axis. Its value is almost constant because the re-parameterization is based on the rotation and it lies around $1 ~\text{rad/-}$. This makes sense since the integral of $a$ corresponds to the initial misalignment angle, which is about $1 ~\text{rad}$. Object invariant $b$ is small, meaning there is almost no translation along the screw axis. The moving frame invariants $\omega_\kappa$, $\omega_\tau$, $v_b$, and $v_t$ are zero or near zero, indicating that the moving frame is stationary. Hence, with the chosen values of the tolerances in the OCP, the relative motion between peg and hole can be explained by a pure rotation about a fixed axis. As a result, the small theoretical translation mentioned earlier vanishes due to the chosen tolerances. This conclusion holds for all twelve trials, indicating that the operator used the same motion strategy in all trials: a pure alignment of peg and hole achieved by a planar motion, without exciting the two other relative degrees of freedom, being relative rotations about peg or hole axes. This is confirmed by Fig. \ref{fig:moving_frames_peg_motion} where all the instantaneous screw axes and frames coincide.

\textbf{Force invariants with respect to world}: Fig. \ref{fig:results_peg_force} shows the screw invariants for the individual demonstration trials. Object invariant $a$, which corresponds to the magnitude of the force vector, is remarkably small, going from less than $4 \text{N}$ to zero during the alignment. As for the moving frame invariants, only $\omega_{\kappa}$ is significantly different from zero, at least for six of the twelve trials. This indicates that, for those six trials the instantaneous screw axis (i.e. the force vector) is rotating with respect to the world in a plane about a fixed point; for the other six trials the instantaneous screw axis remains fixed to the world, i.e. the force vector does not rotate. This in turn indicates that the operator used different wrench strategies in the trials.

\textbf{Force invariants with respect to world and tool for averaged wrench trajectories}: Based on the previous result, we hypothesized that the operator had used different wrench strategies in the two batches of demonstration trials, because they were executed at different points in time. Figure~\ref{fig:results_peg_average} therefore shows the screw invariants for the averaged wrench trajectories of batch 1 (red), batch 2 (blue), and the whole set of trials (green). We show both the invariants with respect to the world (solid lines) and to the tool (dashed lines). Only the first two rotational invariants are shown, as the other invariants are zero ($v_b$, $v_t$) or almost zero ($\omega_\tau$, $b$), similarly as in Fig. \ref{fig:results_peg_force}. Visually, there is an evident distinction between the trials of batch 1 and batch 2: while for batch 1 the average of the trials shows a force vector that rotates with respect to the world and remains fixed with respect to the peg, the opposite is true for the average of the trials of batch 2. To illustrate this further, we plot the resulting moving frames for one trial from batch 1 and one trial from batch 2 in Figures \ref{fig:moving_frames_peg_force_batch1} and \ref{fig:moving_frames_peg_force_batch2}. These figures clearly show that the ISA is rotating for the trial from batch 1 while the ISA is stationary for the trial from batch 2. A more in-depth study and statistical analysis of these wrench strategies go beyond the scope of this paper, but we conclude that different strategies can lead to successful alignments. This is a valuable result for designing robot force control strategies. It is also interesting to see that it could be obtained in an experiment with only small contact forces.

\section{Discussion and Conclusion}
\label{sec:conclusion}

The aim of this work was to introduce local invariant descriptors for force and moment trajectories. Inspired by the invariant descriptors for motion, we exploited the duality between motion and force. We extended existing concepts for motion trajectories to general vector and screw trajectories, and then applied them to force and moment trajectories.

Two types of descriptors were introduced: vector and screw invariants. Vector invariants describe 3D vector trajectories (translational velocities, rotational velocities, forces, and moments) while screw invariants describe 6D screw trajectories (twists and wrenches). Both types of descriptors are invariant for the choice of world reference frame, but the screw invariants are also independent of the reference point on the moving body that is chosen to represent the translational velocity or the moment. Hence, for obtaining the screw invariants there is no need to choose a reference point on the moving body. This is particularly useful in cases where there is no natural choice for such reference point, the reference point can change between executions, or the calibration of the reference point is time consuming. The newly proposed descriptors can be applied to both motion and force trajectories, as opposed to existing invariant descriptors, which were mainly intended for point curves and/or motion trajectories \cite{Calabi1998,piao2006,WuLi2009,shao2015integral,rao2002view,guo2018dsrf,guo2017rrv,DeSchutter2010,lee2017bidirectional,vochten2015}.

Invariant properties allow to recognize and model tasks that were demonstrated in different environments, without the need for accurate calibration. For trajectory reconstruction and generation however, calibration efforts cannot be avoided since for these applications we need initial values as pointed out in the related paragraphs in Sections~\ref{sec:invars_vec} and \ref{sec:invars_screw}. More in detail, we need initial values for the orientation $\matmf{R}_1$ or location $\matmf{T}_1$ of the moving frame, and, if we reconstruct the object trajectory up to position and orientation, we also need respective initial values $\vect{p}_1$ and $\mat{R}_1$. In practical applications, these initial values are to be derived from environment models or from sensor measurements, either directly or using model-based state estimators.

Following a similar line of reasoning, the invariants themselves cannot be used to describe or check relations \emph{between} motion and force, such as reciprocity between a screw twist and a screw wrench, unless the transformation between the respective moving frames in which the motion and force invariants are represented is known. In general, this requires calibration. For example, to check the \textit{reciprocity condition}, i.e., the condition that no power is dissipated in the contact:
$\vect{f} \cdot \vect{v} + \vect{m} \cdot \vect{\omega} = 0,$
it is required that both twist and wrench, $\screw{t}$ and $\screw{w}$, are expressed in the same reference frame and have the same reference point. Therefore, we cannot just plug in the $\vect{a}$ and $\vect{b}$ object invariants of the respective screw invariants, because they are defined relative to their respective moving frames, and, in turn, to the frames in which the twist and wrench trajectories were measured. Hence, to study relations between motion and force, we need a calibration between the respective measurement frames.

The vector and screw invariants can be made independent of time, and hence independent of the applied time-based motion profile, by expressing them as a function of a suitable geometric progress variable.  Again, this additional invariance is useful for the recognition of a task demonstrated in different situations and for the generalization of task models to new situations, as shown in \cite{vochten2019generalizing} for the case of motion in free space. However, the way we chose the progress variable in the 3D contour following application introduced an important limitation. We did not follow the approach to choose an invariant progress variable that is independent of the reference point for translation, as suggested in \cite{DeSchutter2010} and briefly discussed in Section \ref{subsec:progress}. Instead, we chose the progress variable as the integral of the translational velocity of the origin of $\refframe{tcp}$. So, based on our understanding of the contact geometry, we handpicked this point as a reference point to define arc length as a representation of task progress, and used it to represent all invariants. The benefit of this approach is that we obtained vector and screw invariants for both motion and force that are interpretable and can be compared with each other and with the reference values of the limit case (Figures \ref{fig:motion_results}-\ref{fig:force_results_world_meas}). Definitely, this progress value is a very good choice and probably even the best one for this application, but it would be better to derive a suitable progress value from the measured trajectories in an invariant data-driven way. This is subject of our future work.

Besides analytical procedures to derive the invariant descriptors, an approach was presented based on optimal control, where the main idea is to find the invariant descriptor that reconstructs the given window of measurements. This provides a way to deal with singularities and measurement noise and hence enables a much more robust calculation of the descriptors. Compared to our previous work on the calculation of motion descriptors \cite{vochten2018}, we improved the stability and repeatability of the calculation by reformulating the OCP. In the reformulated problem, trajectory reconstruction errors became inequality constraints that were bounded by a desired tolerance value. This resulted in more intuitive tuning, as we showed that these tolerance values could be directly related to the accuracy or noise characteristics of the used motion and force sensor. The reformulated OCP also resulted in a more stable behavior of the moving frame. For example, when the values of the noisy moments are below the specified tolerance, they can be described by a stationary moving frame (see Fig. \ref{fig:force_results_momenttoolpoint} and Fig. \ref{fig:moving_frames_ISA_force}). Finally, the reformulation resulted in a smaller number of tuning parameters, further reducing the tuning effort. Overall, we obtained a good repeatability of the invariants, as can be seen in Figures \ref{fig:motion_results}-\ref{fig:force_results_world_meas}, even for noisy data.

Continuity of the moving frames is ensured by minimizing the values of the moving frame invariants in the OCPs. This also provides some degree of continuity for the reconstructed trajectories. To enforce a more specific continuity on the reconstructed trajectory, it is advised to include the time profile as variables in the OCPs in order to enforce acceleration and/or jerk limit constraints. These constraints can be either on the spatial trajectory itself or on the robot joint trajectories in case a kinematic model of the robot is available, similar to \cite{vochten2019generalizing}.

The paper contains a practical experiment involving a human-demonstrated 3D contour following task which showcases the use, verifies the invariant properties, and validates the robust calculation of screw and vector invariant descriptors for both motion and force. As expected, this experiment yielded highly repeatable motion descriptors due to the highly constrained motion (in a 1-dof instantaneous twist space), and less repeatable force descriptors due to human variation (in a 5-dof instantaneous wrench space). Evidently, apart from human variation, other disturbances were present in this experiment such as sensor inaccuracy and measurement noise. 
A second experiment, involving human demonstration of a peg-on-hole alignment task, showed how motion and force invariants can be used to detect and characterize different human strategies in contact tasks.

The concepts and algorithms presented in this paper are intended as an initial toolbox \cite{software} to support future work on invariant representations of motion and force trajectories and their use in physics-informed machine learning of force-controlled manipulation skills.

\appendix[Relation between the kinematics of the FS frame and the Frenet-Serret equation]
\label{appen:Frenet-Serret}

In differential geometry, the Frenet-Serret equation is usually written as follows:
\begin{equation}
	\label{eq:frenet-serret}
	\begin{bmatrix}
		\vect{T}' \\ \vect{N}' \\ \vect{B}'
	\end{bmatrix}
	=
	\begin{bmatrix}
		0                & \omega_{\kappa} & 0             \\
		-\omega_{\kappa} & 0               & \omega_{\tau} \\
		0                & -\omega_{\tau}  & 0
	\end{bmatrix}
	\begin{bmatrix}
		\vect{T} \\ \vect{N} \\ \vect{B}
	\end{bmatrix},
\end{equation}
with $\omega_\kappa$ the curvature, $\omega_{\tau}$ the torsion, and $\vect{T}$, $\vect{N}$, $\vect{B}$ the unit vectors of the FS frame.
This equation can be shown to be equal to the kinematics of the FS frame as defined in \eqref{eq:kin_R}:
\begin{equation}
	\tag{r.\ref{eq:kin_R}}
	\matmf{R}' = \matmf{R}
	\begin{bmatrix}
		0               & -\omega_{\kappa} & 0              \\
		\omega_{\kappa} & 0                & -\omega_{\tau} \\
		0               & \omega_{\tau}    & 0
	\end{bmatrix}.
\end{equation}
In the previous expression, first expand $\matmf{R}$ into its individual columns using the unit vectors defined in \eqref{eq:tnb}:
\begin{equation}
	\begin{bmatrix}
		\vect{e}'_t & \vect{e}'_n & \vect{e}'_b
	\end{bmatrix} = 	\begin{bmatrix}
		\vect{e}_t & \vect{e}_n & \vect{e}_b
	\end{bmatrix}
	\begin{bmatrix}
		0               & -\omega_{\kappa} & 0              \\
		\omega_{\kappa} & 0                & -\omega_{\tau} \\
		0               & \omega_{\tau}    & 0
	\end{bmatrix}.
	\label{eq:fss}
\end{equation}
Applying the transpose to both sides of \eqref{eq:fss} results in:
\begin{equation}
	\begin{bmatrix}
		\vect{e}_t'^T \\ \vect{e}_n'^T \\ \vect{e}_b'^T
	\end{bmatrix}
	=
	\begin{bmatrix}
		0               & -\omega_{\kappa} & 0              \\
		\omega_{\kappa} & 0                & -\omega_{\tau} \\
		0               & \omega_{\tau}    & 0
	\end{bmatrix}^T
	\begin{bmatrix}
		\vect{e}_t^T \\ \vect{e}_n^T \\ \vect{e}_b^T
	\end{bmatrix}.
\label{eq:resultFS}
\end{equation}
Since the transpose of a skew-symmetric matrix is equal to the negative of the matrix, \eqref{eq:resultFS} corresponds to the Frenet-Serret equation in \eqref{eq:frenet-serret}.

\bibliographystyle{IEEEtran}
\bibliography{bibfile}

\begin{IEEEbiography}[{\includegraphics[width=1in,height=1.25in,clip,keepaspectratio]{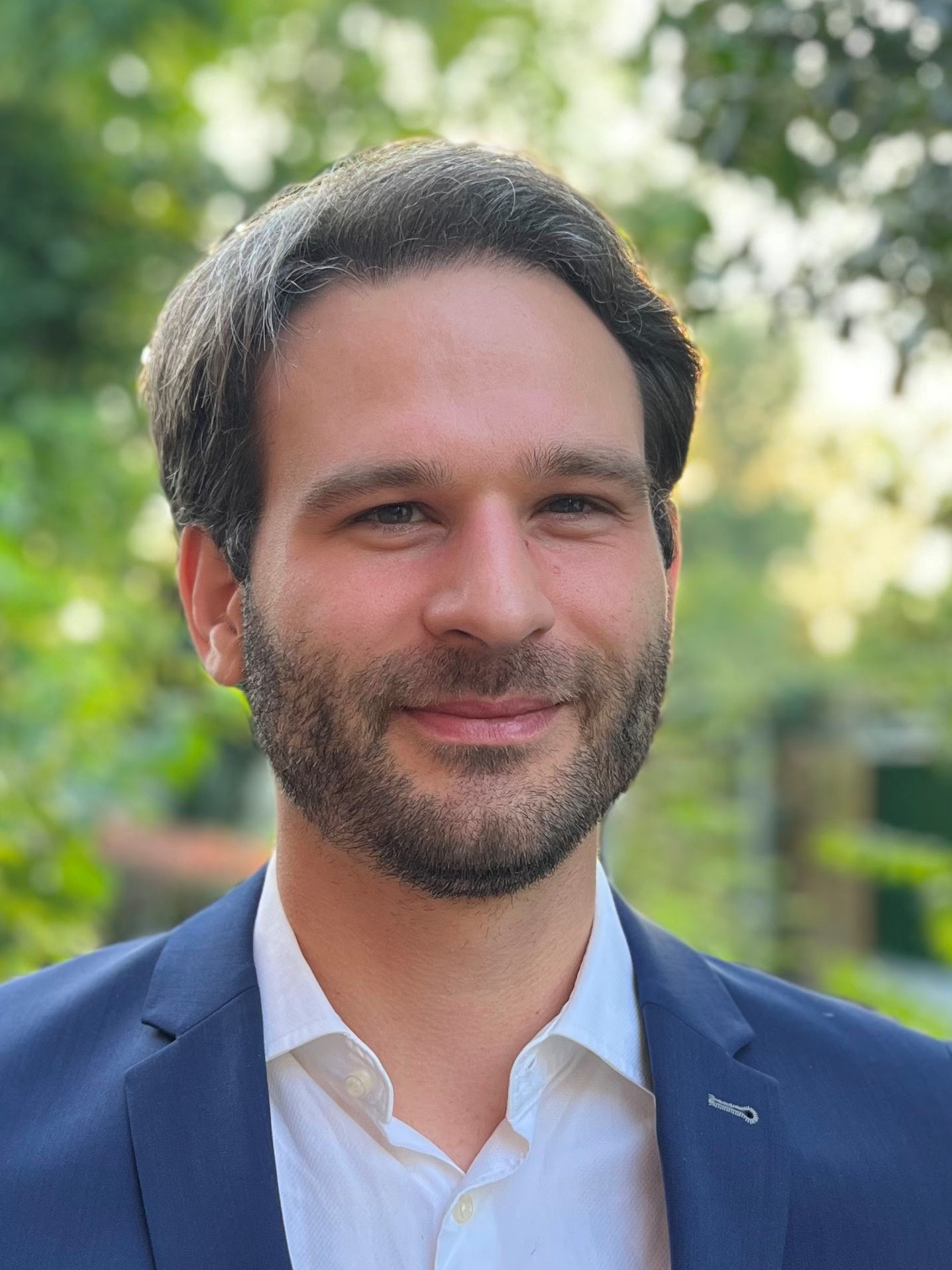}}]{Maxim Vochten}
	is a Postdoctoral Researcher within the Robotics Research Group at KU Leuven, Belgium. He earned his MSc and PhD degrees in Mechanical Engineering from KU Leuven, in 2013 and 2018, respectively. His doctoral work focused on data-efficient trajectory learning with application to motion recognition and generalization, using concepts from differential geometry, screw theory, and optimal control. His current research interests include human intent estimation, robot trajectory generation, and reactive control for human-robot collaboration.
\end{IEEEbiography} %
\vspace{-1.0cm}
\begin{IEEEbiography}[{\includegraphics[width=1in,height=1.25in,clip,keepaspectratio]{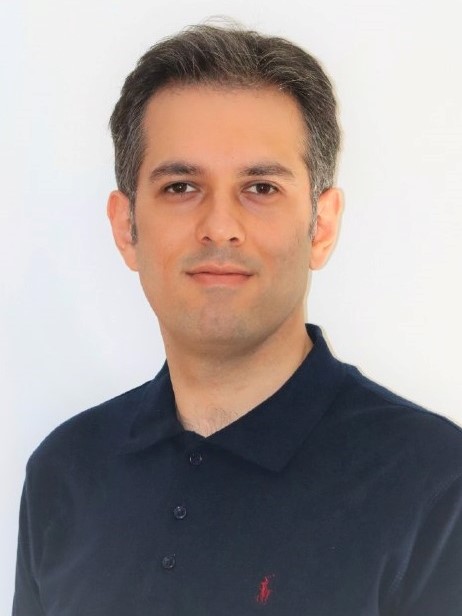}}]{Ali Mousavi Mohammadi}
	earned his B.Sc. and M.Sc. degrees in Mechanical Engineering from Ferdowsi University of Mashhad, Iran, in 2013 and 2016, respectively. In the two years following his graduation, he worked as a part-time researcher at FUM Robotics Lab. In 2019, he joined the Robotics Research Group at KU Leuven's Mechanical Engineering Department. Currently, he is a PhD candidate carrying out his research on force-based manipulation tasks learning. His research interests include robot programming by demonstration and physical human-robot interaction.
\end{IEEEbiography}%
\vspace{-1.0cm}
\begin{IEEEbiography}[{\includegraphics[width=1in,height=1.25in,clip,keepaspectratio]{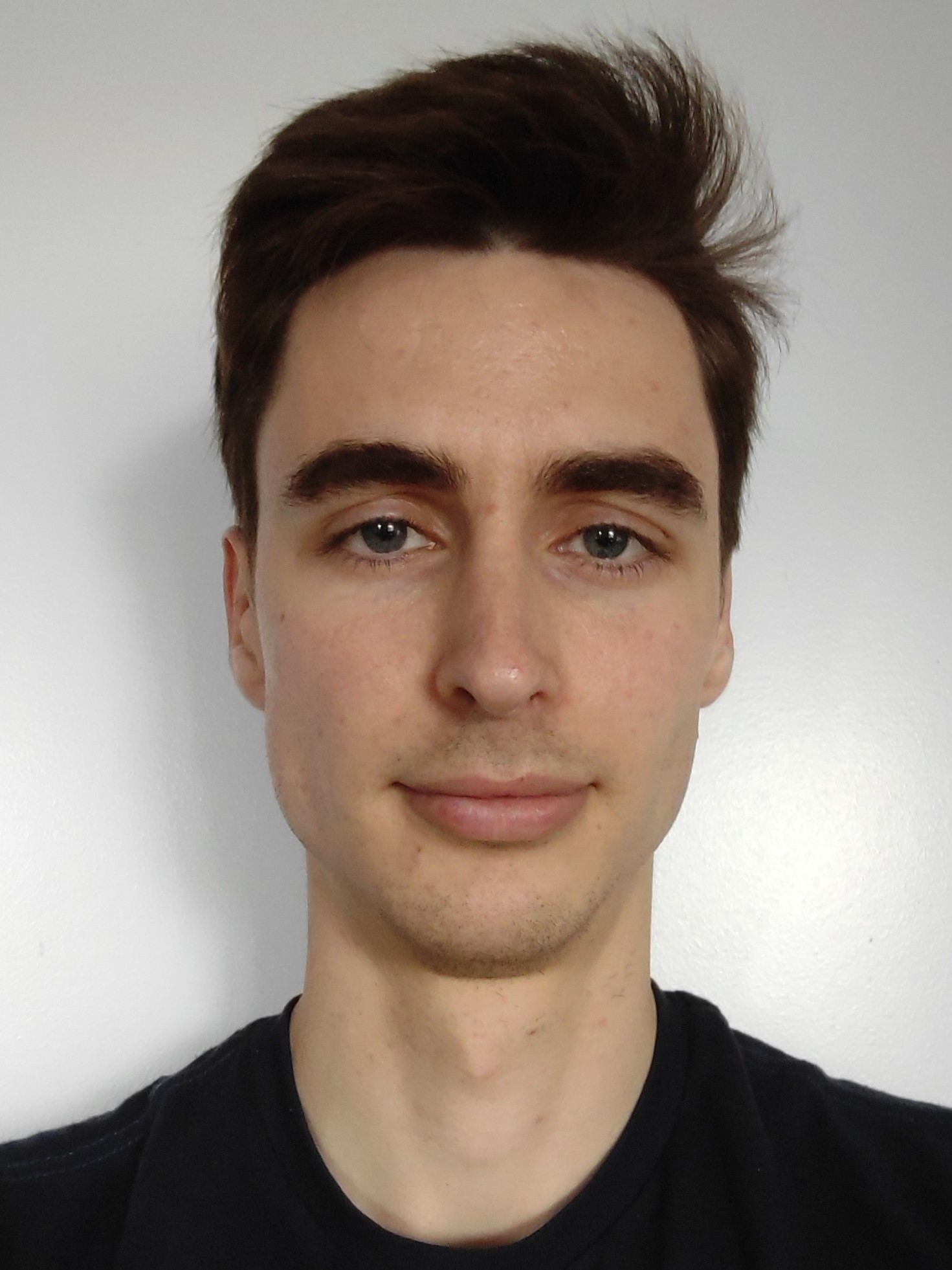}}]{Arno Verduyn} 
	received a masters degree in Mechanical Engineering at KU Leuven in 2020. He continued his research as a PhD candidate as part of the Robotics Research Group. He is currently working on invariant descriptors of rigid-body trajectories and force trajectories. His research is supervised by Professor Joris De Schutter. 
\end{IEEEbiography}%
\vspace{-1.0cm}
\begin{IEEEbiography}[{\includegraphics[width=1in,height=1.25in,clip,keepaspectratio]{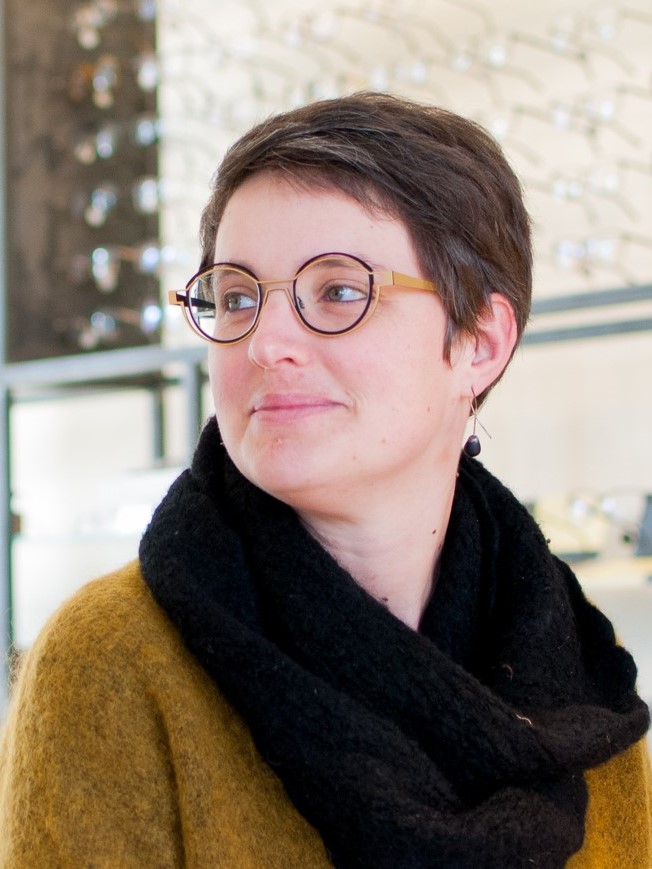}}]{Tinne De Laet}
	is associate professor at the Faculty of Engineering Science, KU Leuven. She obtained a doctoral degree in Mechanical Engineering in 2010 at KU Leuven, Belgium supported by a scholarship of the Research Foundation - Flanders (FWO). She was a post-doctoral researcher of the Research Foundation - Flanders (FWO) at KU Leuven from 2010-2013. Since 2013 she is the Head of the Tutorial Services of Engineering Science where her research focuses on using learning analytics, conceptual learning and metacognition in mechanics. 
\end{IEEEbiography}%
\vspace{-1.0cm}
\begin{IEEEbiography}[{\includegraphics[width=1in,height=1.25in,clip,keepaspectratio]{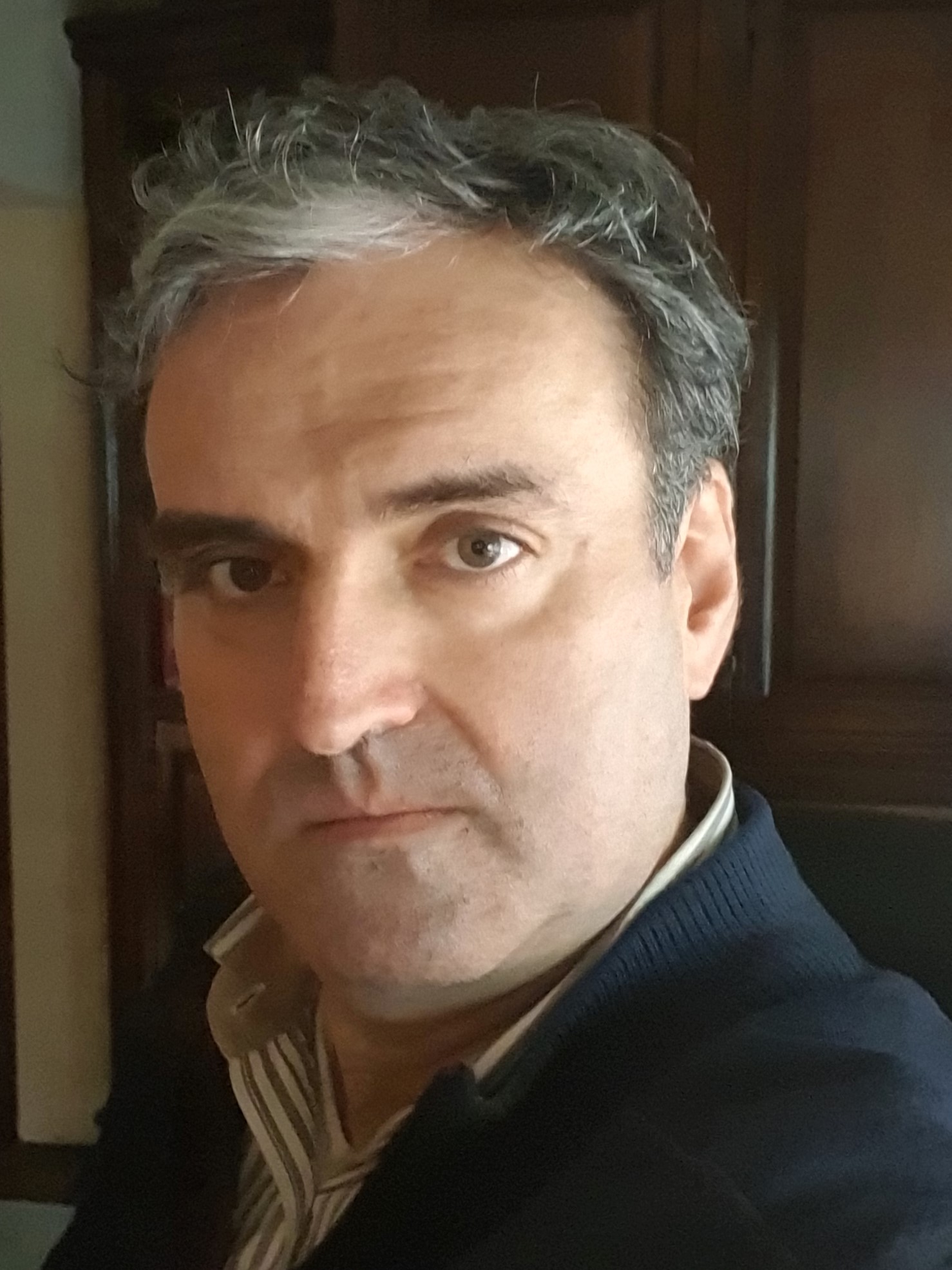}}]{Erwin Aertbeli\"en}
	was born in Antwerp, Belgium, in 1972. He earned his M.Sc. and
	Ph.D. degrees in mechanical engineering from the University of Leuven in 1995
	and 2008, respectively.  He specializes in robot task specification and
	control.  He is the developer and originator of the constraint-based task
	specification formalism known as eTaSL.  His research interests encompass
	all forms of human-robot collaboration, including lower-limb and upper-limb
	exoskeletons, co-manipulation with robot manipulators, teleoperation using
	brain interfaces, and the capturing and modeling of craftsmanship skills.
	Additionally, he has also a research activity in quantitative analysis of
	spasticity and related phenomena.
\end{IEEEbiography}%
\vspace{-1.0cm}
\begin{IEEEbiography}[{\includegraphics[width=1in,height=1.25in,clip,keepaspectratio]{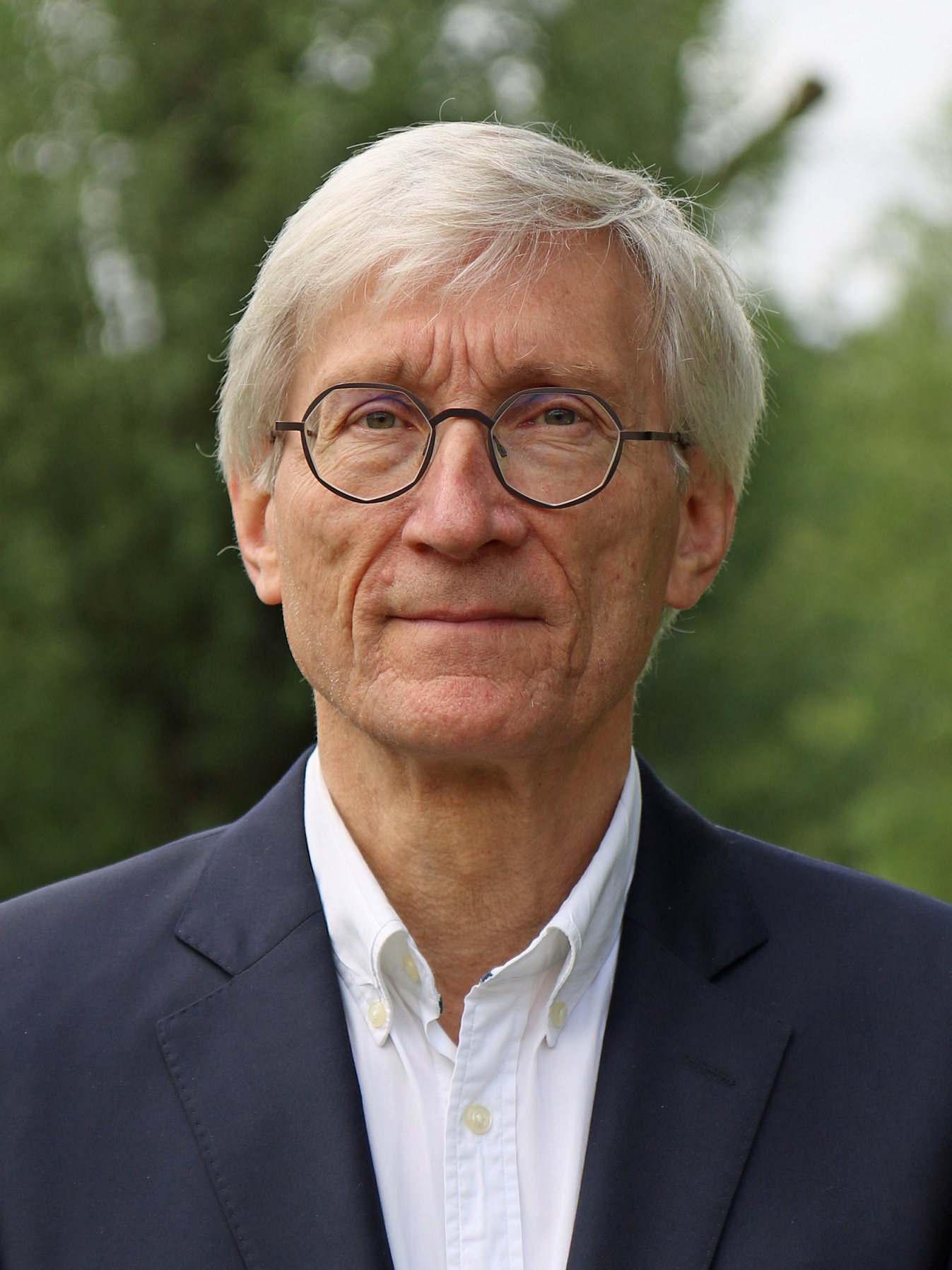}}]{Joris De Schutter}
	is full professor at KU Leuven. He received the MSc degree in mechanical engineering from KU Leuven, Belgium (1980), the MSc degree from MIT (1981), and the PhD degree in mechanical engineering, also from KU Leuven (1986). He pioneered in robot force control and sensor-based control using optimization, contributing to the iTaSC and eTaSL software frameworks. His current research interests include: robot control and programming based on models, sensors and constrained optimization; human-robot interaction; programming by human demonstration; and generalizing human-demonstrated robot skills to new situations. In 2018 he received an ERC Advanced Grant (ROBOTGENSKILL) on this last topic.
\end{IEEEbiography}
\newpage

% Can be used to pull up biographies so that the bottom of the last one
% is flush with the other column.
\enlargethispage{-10.0cm}

\end{document}